\documentclass{article}
\PassOptionsToPackage{numbers, compress}{natbib}

\usepackage[preprint]{neurips_2026}
\usepackage[utf8]{inputenc} 
\usepackage[T1]{fontenc}    
\usepackage{xcolor}         
\definecolor{cvprblue}{rgb}{0.21,0.49,0.74}

\usepackage[pagebackref,breaklinks,colorlinks,citecolor=cvprblue]{hyperref}
\usepackage{hyperref}       
\usepackage{url}            
\usepackage{booktabs}       
\usepackage{amsfonts}       
\usepackage{nicefrac}       
\usepackage{microtype}      
\usepackage[export]{adjustbox} 
\usepackage{wrapfig}
 \usepackage{amsmath} 
  \usepackage{amssymb} 

\renewcommand{\paragraph}[1]{\vspace{1.25mm}\noindent\textbf{#1}}

\usepackage{url}
\usepackage[T1]{fontenc}
\usepackage{graphicx}
\usepackage{latexsym}
\usepackage{xspace}
\usepackage{caption}
\usepackage{subcaption}
\usepackage{booktabs} 
\usepackage{multirow}
\usepackage{textcomp}
\usepackage{color, colortbl}

\usepackage{makecell}
\usepackage{enumitem}
\usepackage{overpic}
\usepackage[most]{tcolorbox}
\usepackage{tikz}
\usepackage{multicol}
\title{Instruction Anchor: Dissecting the Mechanistic Dynamics of Modality Arbitration}

\author{
\textbf{Yu Zhang}$^{1,2}$
\quad {\textbf{Mufan Xu}}$^{3}$
\quad {\textbf{Xuefeng Bai}}$^{1}$ 
\quad {\textbf{Kehai Chen}}$^{1,2}$
\quad {\textbf{Pengfei Zhang}}$^{2}$ \\
\quad {\textbf{Yang Xiang}}$^{2}$
\quad {\textbf{Min Zhang}}$^{1,2}$ \\
\textsuperscript{1}Harbin Institute of Technology, Shenzhen, China, 
\textsuperscript{2}Peng Cheng Laboratory, Shenzhen, China \\
\textsuperscript{3}Harbin Institute of Technology, Harbin, China \\
    \texttt{yuzhang2717@gmail.com}, 
    \texttt{\{baixuefeng,chenkehai\}@hit.edu.cn}\\
}

\begin{document}

\maketitle

\begin{abstract}

Modality following is the ability to selectively leverage multimodal contexts based on user instructions. 
It is fundamental to the safety and reliability of multimodal large language models (MLLMs) in real-world deployments. 
However, the internal mechanisms governing this decision-making process remain largely under-explored.
In this work, we investigate the mechanism underlying  modality following through an information flow perspective.
Our findings reveal that instruction tokens serve as structural anchor for modality arbitration:
Shallow attention layers perform undifferentiated information transfer, aggregating multimodal cues to instruction tokens as a latent buffer; 
in contrast, deep attention layers selectively strengthen the instruction-compliant subspace and resolve modality arbitration according to the instruction-specified intent, with a sparse subset of attention heads driving this process.
Targeted attention-head interventions further validate the functional specificity of these heads:
blocking only $5\%$ of the identified heads substantially degrades modality following while preserving general visual and language capabilities, whereas targeted amplification can restore failed modality-following samples by up to approximately $60\%$.
Together, this work provides a mechanistic account of modality following and informs future efforts to improve how MLLMs integrate and utilize multimodal evidence under user instructions.
\footnote{All data and code will be released upon acceptance.}
\end{abstract}

\section{Introduction}
\label{intro}

Multimodal instruction following (MIF)~\cite{bitton2023visit,Ding_2025_ICCV,multiinstruct} has become a foundational capability for multimodal large language models (MLLMs)~\cite{gpt4,qwen2.5vl,chen2024internvl}, 
enabling them to integrate information across different modalities to execute complex user directives.
%
MIF is pivotal for real-world deployments, such as multi-turn dialogues~\cite{dialogue}, graphical user interface navigation~\cite{gui}, and embodied robotic control~\cite{robot}, where models are required to reliably follow user intent across heterogeneous modalities.
Compared to conventional instruction following in large language models, which primarily focuses on complying with output-format constraints~\cite{bitton2023visit,Ding_2025_ICCV,qian2024mia}, MIF introduces an additional challenge: MLLMs are required to strictly use the modality evidence specified by the user instruction~\cite{guo2025aligned,leng2024curse}, especially when different modalities provide conflicting information.
Despite its significance, the internal decision-making process underlying this selective utilization remains a ``black box'', forming a major obstacle to diagnosing model failures and ensuring behavioral reliability.

In this work, we address this gap by dissecting the internal decision-making mechanisms of MIF through an information flow perspective.
\textbf{\begin{figure}[t]
    \centering
    \includegraphics[page=1, scale=0.7]{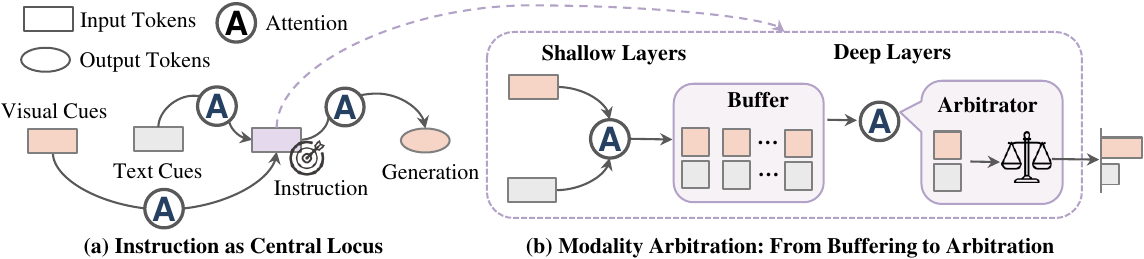}
    \caption{Information flow dissection for modality following. (a) Multimodal cues are preferentially aggregated at instruction tokens, which function as structural anchors. 
    (b) Shallow attention layers route cues to instruction tokens to form a ``latent buffer'' without enforcing selection. Deep attention layers act as the ``definitive arbitrator'', resolving modality arbitration based on instruction intent.
    }
    \label{fig:motivation}
\end{figure}}
We begin by localizing the key attention pathways that transmit modality cues to the final prediction using Attention Knockout Analysis.
To rigorously quantify the importance of each pathway for modality following while suppressing noise from task-irrelevant output variations, we propose Normalized Signed Structural Divergence ($\mathcal{I}_\text{NSSD}$), which measures probability shifts within the decision subspace spanned by the instruction-compliant and competing modalities.
As illustrated in Fig.~\ref{fig:motivation} (a), attention pathways that route modality cues into instruction tokens are far more prominent than those directly connecting cues to generated tokens, indicating that instruction tokens serve as the primary aggregation site for modality-relevant cues.
Moreover, latent-decision tracking shows that the modality decision decoded at the instruction tokens aligns with the final prediction in over $95\%$ of cases, suggesting that modality arbitration is largely resolved at instruction tokens.

Building on the observation that instruction tokens act as the primary aggregation site for multimodal cues, we further examine how modality arbitration unfolds inside the instruction tokens.
Using a logit-difference attribution analysis, we find that, compared with MLP layers, attention exhibits a more stable positive shift in the arbitration margin toward the instruction-specified intent across both successful and failed modality-following samples.
This suggests that attention is more closely associated with instruction-aligned modulation of the arbitration margin.
Motivated by this observation, we further dissect attention across layers. As shown in Fig.~\ref{fig:motivation} (b), we reveal that shallow attention layers transfer multimodal cues to instruction tokens to form a ``latent buffer'' without enforcing selection. In contrast, deep attention layers act as a ``definitive arbiter'', resolving modality arbitration based on instruction intent, with a remarkably sparse attention heads driving this process.
We perform targeted attention-head interventions to examine the functional specificity of the identified heads and their potential for effective modality following.
As a result, blocking only $5\%$ of the identified heads substantially degrades modality following while preserving general visual and language capabilities, supporting their functional specificity to modality-following behavior.
Conversely, selectively amplifying these heads restores the modality-following ratio on failed cases by up to approximately $60\%$, demonstrating the role of these heads for targeted enhancement of modality following.

Our main contributions are summarized as follows:

\begin{itemize}[leftmargin=20pt]
    \setlength{\itemsep}{0pt}
    \item  In this work, we investigate the under-explored internal mechanism of modality following from an information flow perspective.
    \item We identify instruction tokens as critical structural anchors for modality arbitration.
    \item We characterize a layer-wise transition in the attention mechanism, moving from buffering to arbitration, with sparse attention heads driving.
    \item This work proposes a new perspective for understanding modality arbitration and informs future efforts to improve multimodal evidence utilization in MLLMs.
\end{itemize}

\section{Related Work}
\noindent\textbf{Multimodal Instruction Following (MIF).}
Multimodal language models (MLLMs) have achieved remarkable success across a wide range of domains~\cite{zhu-etal-2025-benchmarking,qwen2.5vl,liang2026vanim,zhang2025cross,wei2025mm,liang2026render,yan2026knowledgeinferencescalinglaws}, demonstrating exceptional capabilities in integrating and reasoning over heterogeneous data~\cite{chen2024internvl,weifirst,Zhang_2025_ICCV,li2025miv,li2026make,zhang2026mitigating,zheng2025locot2v,zhao2026clauseagenticneurosymbolicknowledge}.
MIF is MLLMs' capacity for the precise execution of instructions, requiring the selective integration of multimodal contexts~\cite{chen2025some,leng2024curse} and adherence to predefined output formats~\cite{Ding_2025_ICCV,he2026empowering}. 
Research on MIF is fundamentally categorized into two dimensions: 
\textit{1) Instruction-Driven Format Compliance} constrains the model’s output format and has evolved from open-ended evaluation paradigms~\cite{bitton2023visit, qian2024mia} to rigorous benchmarks targeting complex, vision-dependent constraints~\cite{Ding_2025_ICCV, he2026empowering}.
Correspondingly, enhancement efforts focus on scaling high-quality instruction-following data~\cite{chen2024allava,chen2023sharegpt4vimprovinglargemultimodal} and applying preference-alignment strategies such as SFT and DPO to ensure strict adherence to structural output requirements~\cite{Ding_2025_ICCV, he2026empowering};
\textit{2) Precise Context Utilization} focuses on the accurate synthesis of evidence from heterogeneous modalities guided by instructional intent~\cite{chen2025some,guo2025aligned, leng2024curse}. 
Existing work has evolved from alignment-driven fine-tuning and rigorous evaluation~\cite{chen2025some,guo2025aligned} to enhancing behavioral fidelity, which reveals challenges such as hallucinations in cross-modal understanding~\cite{leng2024curse}. 
However, the internal mechanisms governing this decision-making process remain largely under-explored.

\noindent\textbf{Interpretability in MLLMs.}
Existing literature on the mechanistic interpretability of MLLMs~\cite{basu2024understanding,ben2024lvlm,survey1,huang2024miner} is predominantly anchored in a \textit{perception-centric perspective}, focusing on the encoding, storage, and retrieval of visual information within the Transformer architecture~\cite{gpt4,vaswani2017attention}. 
One trajectory focuses on pinpointing the specific neural topography responsible for multimodal processing, isolating modality-specific neurons~\cite{huang2024miner,pan2024finding} or task-contingent sub-circuits~\cite{nikankin2025same} that disentangle cross-modal mechanisms.
Another trajectory~\cite{basu2024understanding,ben2024lvlm,zhang2025cross} interrogates the dynamic propagation of signals, employing causal interventions and attribution methods to trace the underlying information pathways. 
Concurrently, a burgeoning line of work seeks to decode semantic content by projecting activations onto human-understandable concepts through tools such as Sparse Autoencoders (SAEs)~\cite{lou2025sae} or the Logit Lens~\cite{neo2024towards,yu2024understanding}.
While perception-centric studies have advanced our understanding of MLLMs, the mechanisms governing cross-modal arbitration remain largely overlooked. This work elucidates the dynamics of modality-following by identifying instruction tokens as the critical structural locus for decision crystallization, providing a novel lens into multimodal information utilization.

\section{Instruction Serves as Structural Anchor}
\label{sec:instruction_anchors}

We begin by investigating the pathways through which modality cues are aggregated and integrated to form decisions during modality following.
Our findings reveal that instruction tokens serve as structural anchors: they aggregate cross-modal cues (\S\ref{sec:routing}) and act as the definitive locus where modality arbitration is finalized before information is propagated to the generated tokens (\S\ref{sec:decoding}).


\subsection{Diagnostic Setup}
\label{sec:dataset}

For mechanistic readout, we construct a controlled diagnostic setting based on~\cite{zhang2025evaluating}, where visual and textual contexts support different answers. Each instance is abstracted as
$S = \langle C_p, C_c, I, A_p, A_c, \mathcal{E}_p, \mathcal{E}_c \rangle$,
where $I$ denotes the active instruction specifying the target modality, $C_p$ and $C_c$ are the instruction-compliant and competing contexts, and $A_p$, $A_c$ are the corresponding answers. The answer entity dictionaries $\mathcal{E}_p$ and $\mathcal{E}_c$ aggregate up to ten semantically equivalent surface forms to robustly track modality-specific signals.\footnote{Unless otherwise specified, our main analysis traces the internal routing and arbitration of successful modality-following instances, while failure cases are analyzed to localize the internal stage at which modality-following fails in \S\ref{sec:tripartite}.} 
The model produces an output $Y$ conditioned on $C_p$, $C_c$ and $I$, expected to align with $C_p$ based on the intent $I$. 
We study decoder-only MLLMs built on the Transformer architecture. Given an input token sequence $X = [x_1, \dots, x_N]$, we partition tokens into visual tokens ($X_\text{vision}$), textual context tokens ($X_\text{ctx}$), and instruction tokens ($X_\text{inst}$).
Each token $x_i$ passes through $L$ Transformer residual layers:$\mathbf{h}_i^l = \mathbf{h}_i^{l-1} + \mathbf{A}_i^l + \mathbf{F}_i^l$,
where $\mathbf{A}_i^l$ and $\mathbf{F}_i^l$ are the attention and MLP outputs, respectively.
Leveraging $\mathcal{E}_p$ and $\mathcal{E}_c$, we define two subspaces, $Y_p$ and $Y_c$, corresponding to the instruction-compliant and competing answers, respectively. 
Then we compute the maximum logit across all candidate entities for each subspace to obtain a robust measure of model belief.\footnote{Attention intervention experiments Fig.~\ref{fig:header_ver_ablation} in Apdx.~\ref{supp:method_ablation} reveal that the maximum logit provides a more representative measure of subspace activation than the average logit~\cite{dissecting} for modality following.}
Formally, we probe the layer-wise subspace states via Logit Lens~\cite{geva2022transformer}:
\begin{equation}
\label{eq:max_logit}
    \mathrm{Logit}(Y_m \mid \mathbf{h}_i^l) = \max_{y \in \mathcal{E}_m} (\mathbf{E}\cdot (\mathbf{h}_i^l))_y, 
    \quad m\in \{p,c\}
\end{equation}
where $\mathbf{E}$ is the unembedding matrix. 
When candidate answers consist of multiple tokens, we take the first token as the representative. 
A granular diagnostic setup is provided in Apdx.~\ref{supp:dataset_construction}.

\subsection{Information Routing Reveals Instruction Anchors}
\label{sec:routing}

\subsubsection{Method: Attention Knockout Analysis}
\label{sec:casual_attention_knockout}
Attention serves as the primary conduit for cross-modal information flow~\cite{lu2023multi,zhang2025cross}. 
The attention output $\mathbf{A}_i^l$ for $i$-th token at layer $l$ is computed via multi-head self-attention:
\begin{equation}
\label{eq:header_compute}
\mathbf{A}_i^l = \sum_{j=1}^{J} \mathrm{Head}_i^{l,j}, \quad
\mathrm{Head}_i^{l,j} = \mathrm{Softmax}\Big(\frac{\mathbf{q}_i^{l,j} (\mathbf{K}^{l,j})^\top}{\sqrt{d_k}} + \mathbf{M}^l_i) \Big) \mathbf{V}^{l,j} \cdot\mathbf{W}_O^{l,j},
\end{equation}
Here, $\mathbf{q}_i^{l,j}$ is the query vector of token $i$, $\mathbf{K}^{l,j}$ and $\mathbf{V}^{l,j}$ are the key and value projections from the previous layer, $\mathbf{W}_O^{l,j}$ is the output projection, and $\mathbf{M}^l \in \mathbb{R}^{N \times N}$ is the causal mask ($M_{s,t}^l = 0$ for $t \le s$, $-\infty$ otherwise).  
We employ Attention Knockout Analysis~\cite{dissecting} to selectively remove edges in the attention graph and probe critical attention pathways. A \textit{Target Pathway} $\mathcal{P}_{\text{src} \rightarrow \text{dst}}$ comprises edges from source token set $X_\text{src}$ to destination set $X_\text{dst}$.  
Following~\cite{zhang2025cross}, we block the target pathway by modifying the causal mask across all heads within a window of $k$ layers centered at layer $l$. we set $k=3$ unless otherwise noted. Sensitivity analyses in Apdx.~\ref{supp:app_window} confirm robustness across window sizes. The modified mask is:
\begin{equation}
\tilde{M}_{s,t}^{l} =
\begin{cases}
-\infty, & (s,t) \in \mathcal{P}_{\text{src} \rightarrow \text{dst}},\\
M_{s,t}^{l}, & \text{otherwise}.
\end{cases}
\end{equation}

\subsubsection{Metric: Normalized Signed Structural Divergence}

We quantify the effect of attention knockout on modality following by measuring the signed distributional displacement within the modality-arbitration subspace.
Let $P$ and $\widetilde{P}$ denote the output distributions before and after intervention, respectively.
For each instance, let $\mathcal{U}$ be the decision subspace covering the instruction-compliant subspace $Y_{p}$ and the competing subspace $Y_{c}$.
We denote by $P_{\mathcal{U}}$ and $\widetilde{P}_{\mathcal{U}}$ the corresponding subspace-normalized distributions.
This restriction isolates the prediction components directly involved in modality following, allowing the metric to focus on intervention-induced changes in the arbitration-relevant decision structure.
We define the Normalized Signed Structural Divergence as:
\begin{equation}
\mathcal{I}_\text{NSSD}:
=
\operatorname{sign}
\!\left[
\widetilde{P}_{\mathcal{U}}(Y_{p})
-
P_{\mathcal{U}}(Y_{p})
\right]
D_{\mathrm{KL}}
\!\left(
P_{\mathcal{U}}
\,\middle\|\,
\widetilde{P}_{\mathcal{U}}
\right),
\end{equation}
where the sign term specifies the direction of the knockout effect on the instruction-compliant interpretation.
The KL divergence term measures how strongly the intervention reshapes the prediction distribution within the modality-arbitration subspace.
Intuitively, a negative sign means that removing the pathway shifts probability mass away from the instruction-compliant region, indicating that the blocked pathway originally supported modality following.

\begin{figure*}[t] 
    \centering
    \begin{subfigure}[b]{0.495\textwidth}
        \centering
        \includegraphics[page=1, width=\textwidth]{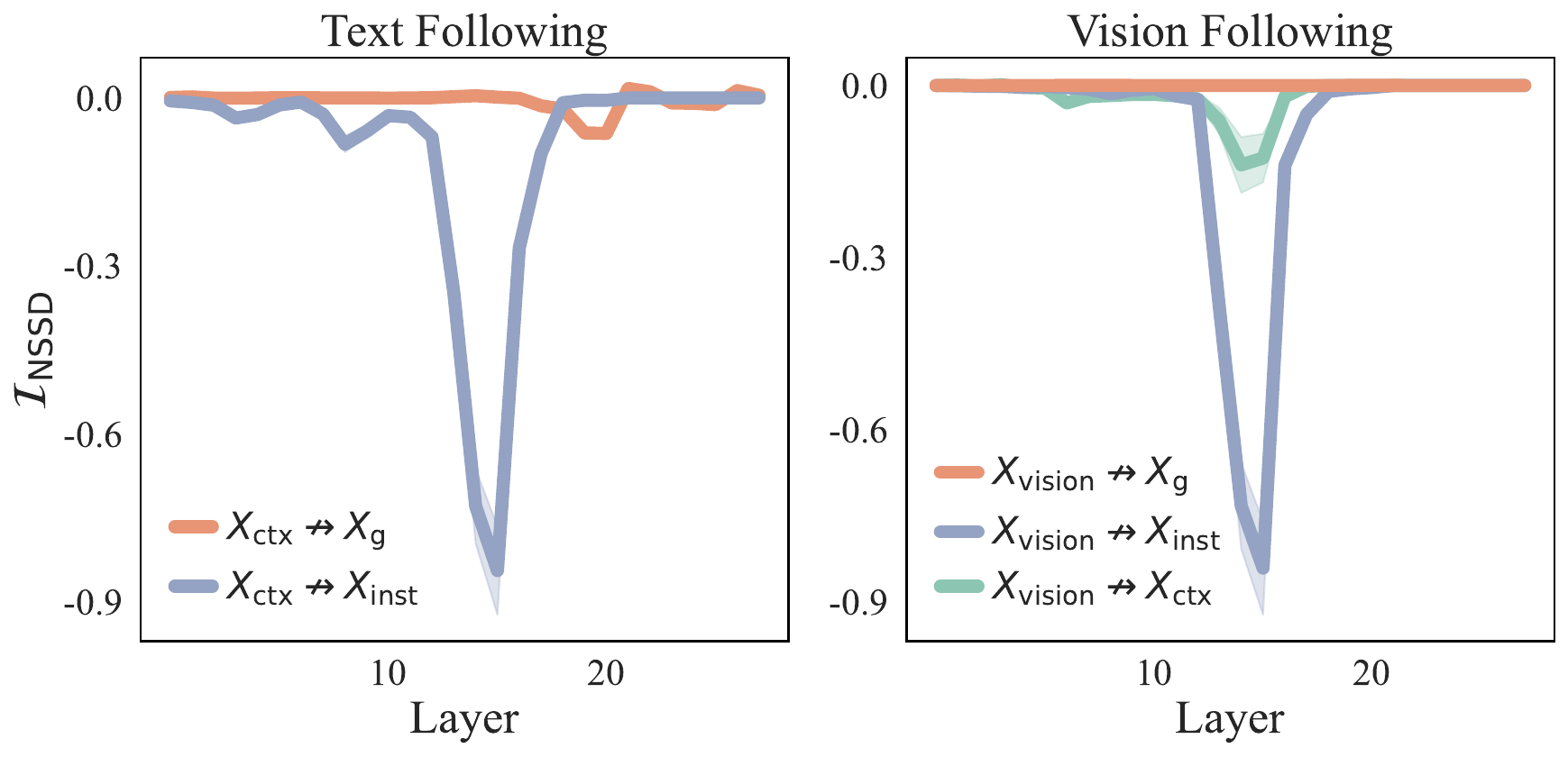}
        \caption{Qwen2.5-VL-7B}
        \label{fig:sub_qwenvl}
    \end{subfigure}
    \hfill %
    \begin{subfigure}[b]{0.495\textwidth}
        \centering
        \includegraphics[page=1, width=\textwidth]{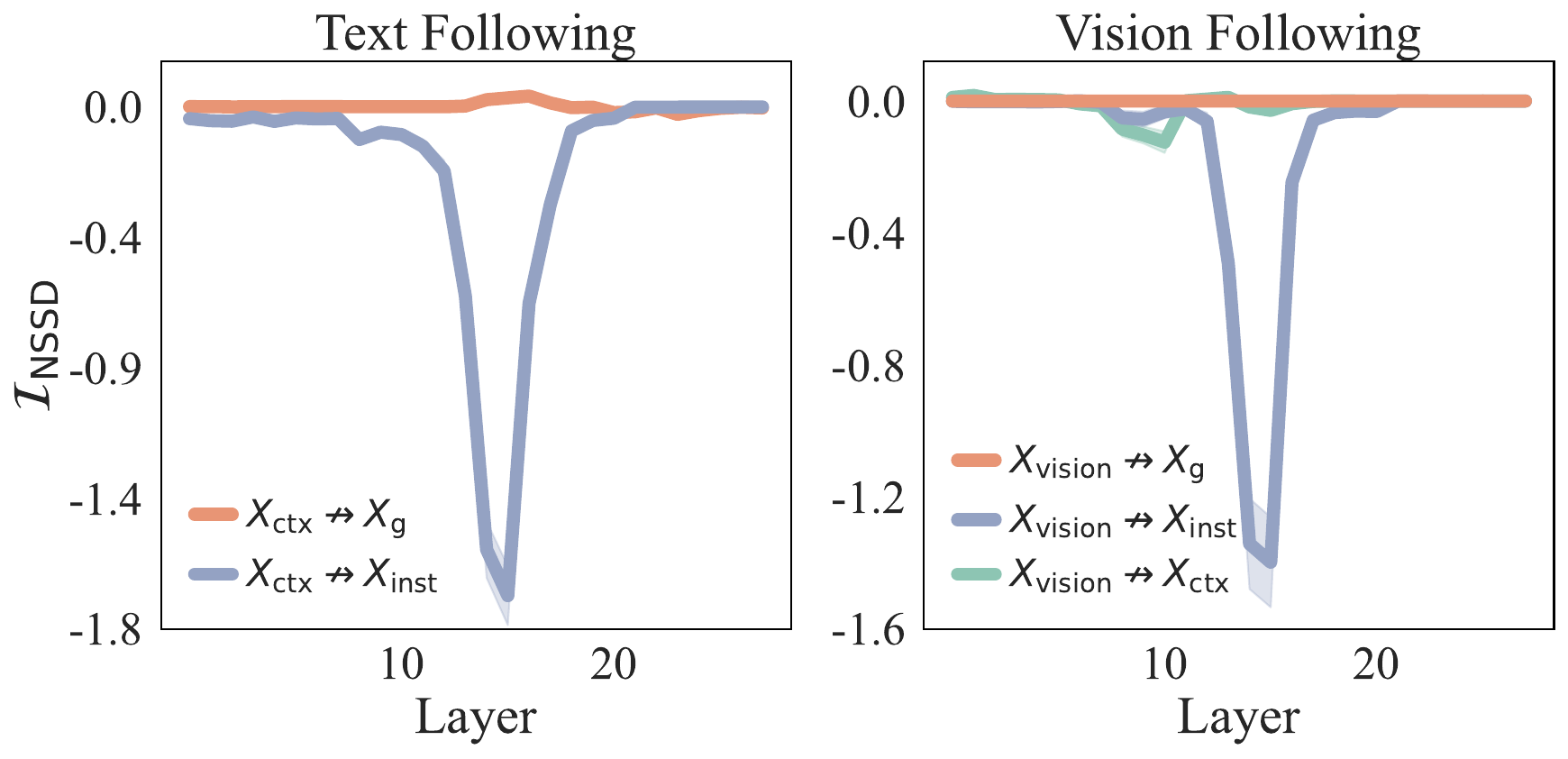}
        \caption{InternVL3-8B}
        \label{fig:sub_internvl}
    \end{subfigure}
    \caption{
\textbf{Layer-wise $\mathcal{I}_\text{NSSD}$ under attention knockout.} Each curve corresponds to removing a specific attention pathway, where $X_\text{vision}$, $X_\text{ctx}$, $X_\text{inst}$, and $X_\text{g}$ denote visual tokens, text context tokens, instruction tokens, and generated tokens. More negative values of $\mathcal{I}_\text{NSSD}$ indicate greater importance of the pathway for modality following. Pathways targeting $X_\text{inst}$ exhibit larger negative shifts than those to $X_\text{g}$, highlighting instruction tokens as the central convergence site for modality cues.}
    \label{fig:vision_inst_block_comparison}
\end{figure*}
\subsubsection{Results}
\label{sec:attention_pattern_main}

In our experiments, we examine the information flow among visual tokens ($X_\text{vision}$), text context tokens ($X_\text{ctx}$), instruction tokens ($X_\text{inst}$), and generated tokens ($X_\text{g}$) to characterize how modal signals are routed during modality following. 
For each layer, we compute the mean and standard deviation of $\mathcal{I}_\text{NSSD}$ across all evaluation samples. Several key findings emerge:

\noindent\textbf{Instruction Tokens as a Central Locus.}
Fig.~\ref{fig:vision_inst_block_comparison} reports the layer-wise $\mathcal{I}_\text{NSSD}$ profiles for Qwen2.5-VL-7B and InternVL3-8B.
Several observations can be drawn from these results.
\textbf{First}, cutting pathways from the instruction-compliant modal context to the generated tokens produces only marginal changes in $\mathcal{I}_\text{NSSD}$. 
This indicates that the generated tokens do not appear to rely primarily on direct attention to the modal cues when producing modality-following responses.
\textbf{Second}, cutting the pathways from $X_\text{vision}$ or $X_\text{ctx}$ to the instruction tokens $X_\text{inst}$ results in a pronounced negative shift in $\mathcal{I}_\text{NSSD}$. 
This suggests that the pathway from modal cues to instruction is the most sensitive routing component for maintaining the correct instruction following.
\textbf{Third}, in vision-following cases, cutting the attention pathway 
$X_\text{vision} \rightarrow X_\text{ctx}$ produces a markedly smaller effect than  
$X_\text{vision} \rightarrow X_\text{inst}$. 
This contrast indicates that the main structural sensitivity in vision-following is not distributed uniformly over textual tokens. 
Instead, it is more concentrated around pathways involving instruction tokens.
Taken together, these results reveal a cross-modal relay pattern: instruction tokens form a prominent convergence site that aggregates multimodal cues during modality following.

\noindent\textbf{Instruction Convergence Is Not a Positional Artifact.}  
To test whether convergence on instruction tokens is due to their proximity to generated tokens, we analyze two types of instruction tokens: modality-specifying semantic tokens ($X_s$) and output-format constraint tokens ($X_o$), e.g., ``Answer the question with a single word,'' both located near the generated tokens.  
We selectively cut the attention pathway from modality cues to $X_s$ or $X_o$ in Fig.~\ref{fig:xxxx} (a). Cutting the pathway to $X_o$ has minimal effect, while $X_s$ substantially reduces modality following, indicating that convergence reflects semantic, not positional, factors.  
Additional experiments—including swapping $X_s$ and $X_o$ positions and inserting unrelated instructions (Apdx.~\ref{supp:semantic_constraint})—further support this conclusion.

\noindent\textbf{Generalization and Robustness of Attention Routing.} 
In Apdx.~\ref{supp:generalization_causal}, we verify the generalization and robustness of attention routing including generalization to larger dense MLLM and MoE-based MLLM, generalization to open-ended, multi-token generation scenarios, such as image captioning, and the robustness across different analysis subsets and instruction styles.

\noindent\textbf{Takeaways.} Instruction tokens serve as the primary aggregation site for modality-relevant cues.

\subsection{Decoding Modality Arbitration in Instruction Tokens}
\label{sec:decoding}


The information routing analysis in \S\ref{sec:routing} has identified instruction tokens as a structural locus where cross-modal cues converge. A natural question arises: \textbf{Whether the modality arbitration is finalized within instruction tokens?}

\subsubsection{Method: Internal Belief Tracking}
\label{sec:crucial_method}
To answer this, we track layer-wise evolution of the internal belief using Logit Lens~\cite{geva2022transformer} for instruction-compliant subspace $Y_{p}$ and competing subspace $Y_c$.
Formally, we define a readout function $S_m$ to quantify the intensity of a specific subspace within instruction tokens:
\begin{equation}
\label{eq:s_l_m}
S_m (\mathbf{H}_\text{inst}^l) := \frac{1}{K} \sum_{i \in \mathcal{T}_m^l} \operatorname{Logit}(Y_m \mid \mathbf{h}_i^l), 
\quad m \in \{p,c\},
\end{equation}
where $\mathbf{H}_\text{inst}^l$ denotes the hidden states of the instruction tokens at layer $l$, $\mathcal{T}_m^l$ is the set of indices of the top-$K$ instruction tokens with the highest logit activations in subspace $m$. 
This follows the \textit{semantic sparsity} hypothesis, focusing on the peak activations that represent the crystallized intent~\cite{liu2023deja}. 
In practice, we set $K=1$ and robustness analyses for varying $K$ are provided in Apdx.~\ref{supp:method_ablation}.

\begin{figure*}[t] 
    \centering
    \begin{subfigure}[b]{0.5\textwidth}
        \centering
        \includegraphics[page=1, width=\textwidth]{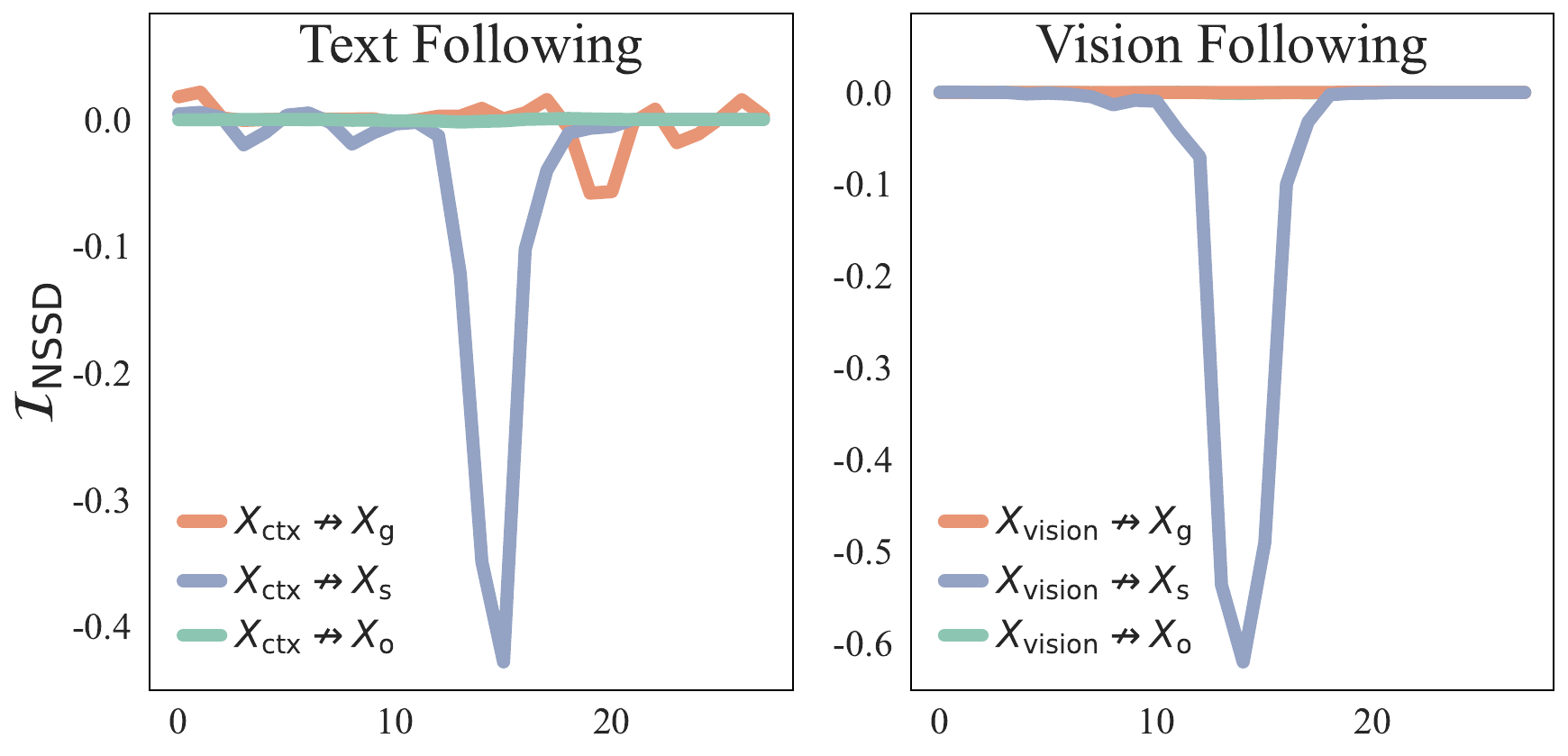}
        \caption{Comparison with semantic and format instruction}
    \end{subfigure}
    \hfill %
    \begin{subfigure}[b]{0.24\textwidth}
        \centering
        \includegraphics[page=1, width=\textwidth]{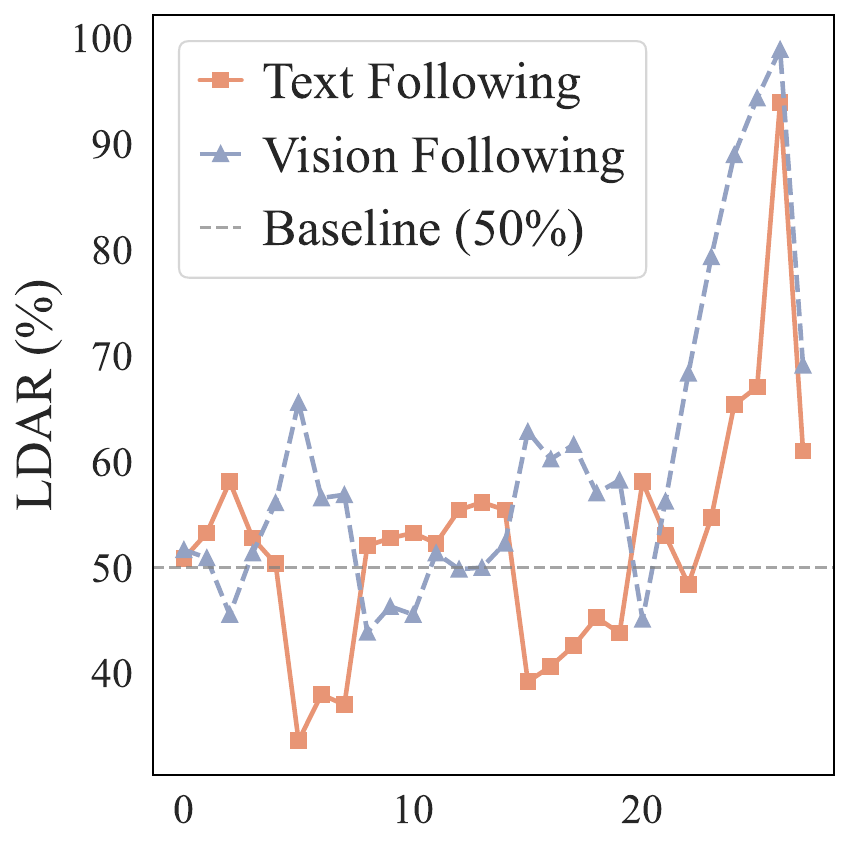}
        \caption{Layer-wise LDAR}
    \end{subfigure}
    \hfill %
    \begin{subfigure}[b]{0.24\textwidth}
        \centering
        \includegraphics[page=1, width=\textwidth]
        {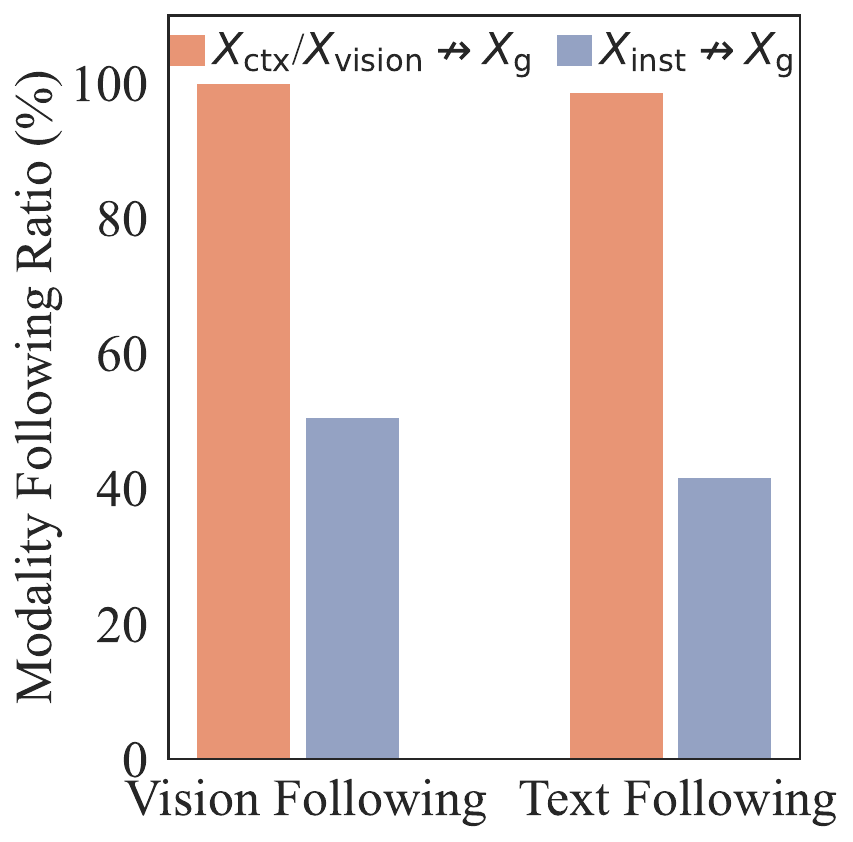}
        \caption{Effect of Pathway}
        \label{figure/Neurips/modality_following_ratio_ver.pdf}
    \end{subfigure}
    \caption{
    \textbf{(a)} Layer-wise $\mathcal{I}_\text{NSSD}$ under attention knockout, where $X_\text{vision}$, $X_\text{ctx}$, $X_\text{g}$, $X_\text{s}$ and $X_\text{o}$ denote visual, text context, generation, semantic instruction  for modality following and output-format instruction tokens.
    \textbf{(b)} Layer-wise LDAR of instruction tokens. Higher LDAR indicates greater alignment between latent state of instruction tokens and the final  output, with LDAR = 1.0 signaling that modality arbitration is resolved within instruction tokens.
    \textbf{(c)} Modality following ratio after cutting attention pathways from instruction $X_\text{inst}$ or instruction-compliant modal context ($X_\text{vision}/X_\text{ctx}$) to generated tokens ($X_\text{g}$). The lower the value, the more important the pathway.}
    \label{fig:xxxx}
\end{figure*}
\subsubsection{Metric: Latent Decision Alignment Rate}

To quantify the crystallization of modality arbitration within struction tokens, we propose Latent Decision Alignment Rate (LDAR), which measures the layer-wise synchronization between the internal state of instruction tokens and final behavioral output. 
Given a sample, we consider instruction tokens at layer $l$ to be aligned with final output if the signal strength of the instruction-compliant subspace dominates its competitor. 
Formally, for a diagnostic dataset $\mathcal{D}$, the LDAR at layer $l$ is defined as:
\begin{equation}
\operatorname{LDAR}(l) := \frac{1}{|\mathcal{D}|} \sum \mathbb{I}\left[ S_{p}(\mathbf{H}_\text{inst}^l) > S_{c}(\mathbf{H}_\text{inst}^l) \right],
\label{eq:ldar}
\end{equation}
where the summation is taken over all samples in $\mathcal{D}$ and $\mathbb{I}[\cdot]$ is the indicator function. $\text{LDAR}=1.0$ means instruction tokens’ internal states can correctly predict the final decision for all samples, while $0.5$ indicates chance-level alignment.

\subsubsection{Result}
\label{sec:ldar_analysis}
As shown in Fig.~\ref{fig:xxxx} (b), LDAR remains near chance (around 0.5) in shallow layers, indicating the instruction-compliant subspace has not yet dominated. In deeper layers, LDAR rises sharply to over 95\%, showing that modality arbitration is effectively crystallized within the instruction tokens. We further validate this with two additional analyses:

\noindent\textbf{Critical Role of Instruction-Mediated Information Flow.}  
We assess the necessity of the pathway from instruction tokens ($X_\text{inst}$) to generated tokens ($X_\text{g}$) by selectively cutting attention from either modality cues or from $X_\text{inst}$ to $X_\text{g}$ across deep layers, which are identified in Fig.~\ref{fig:xxxx} (b) as critical for decision formation at the instruction tokens.
As shown in Fig.~\ref{fig:xxxx} (c), severing the attention pathway from $X_\text{inst}$ to $X_\text{g}$ leads to a dramatic drop in modality-following performance, whereas cutting the attention pathway from the modality cues to $X_\text{g}$ has a negligible effect. These results confirm that the final modality decision is primarily mediated through instruction tokens.

\noindent\textbf{Decision Synchronization between Instruction and Generated Tokens.} 
Besides, we measure the sample-wise agreement between the latent modality decision decoded at instruction tokens ($X_\text{inst}$) and at generated tokens ($X_\text{g}$). Across critical layers in Fig.~\ref{fig:xxxx} (b), these two positions exhibit over 90\% alignment in modality arbitration, regardless of whether the instruction is correctly followed. 
Notably, this high-fidelity correspondence persists even in layers where the absolute LDAR is below 70\%, indicating that instruction anchors and generated tokens converge on a shared modality decision.

\noindent\textbf{Takeaways.} Collectively, these findings demonstrate that modality arbitration is effectively crystallized at the instruction tokens before being propagated to the generated tokens.

\section{Mechanistic Dissection of Modality Arbitration}
We have identified the structural role of instruction tokens in mediating modality arbitration (\S \ref{sec:instruction_anchors}). In this section, we investigate how instruction tokens orchestrate the arbitration process. Our analysis reveals that shallow attention layers act as a latent buffer for undifferentiated information transfer, whereas deep attention layers serve as the primary arbiters that resolve modality selection.
At a finer granularity, we find the arbitration process is largely driven by a sparse subset of specialized attention heads (\S \ref{sec:tripartite}). 
Finally, the attention intervention experiments confirm that the identified attention heads are functionally specialized and can effectively modulate modality following (\S \ref{sec:attention_head_verification}).

\subsection{Component Analysis for Modality Arbitration}
\label{sec:tripartite}
We quantify the behavior of attention and MLP for modality arbitration based on the readout function $S_m(\mathbf{H}_\text{inst}^l)$ in Eq.~(\ref{eq:s_l_m}), which measures the layer-wise logit intensity of subspace $m \in \{p,c\}$ at instruction tokens. 
Following the residual update, $\mathbf{H}_\text{inst}^l = \mathbf{H}_\text{inst}^{l-1} + \mathbf{A}_\text{inst}^l + \mathbf{F}_\text{inst}^l$
we also apply this readout to the intermediate states as $S_m(\mathbf{H}^{l-1}_\text{inst} + \mathbf{A}^l_\text{inst})$. 
Besides, we define the layer-wise modality arbitration margin as $\Delta S = S_p - S_c$, which provides a direct measure of the residual-stream signals relevant for modality arbitration within instruction tokens. 
For convenience, let $\mathcal{M} \in \{S_p, S_c, \Delta S\}$ denote a generic readout. 
To this end, we track how attention and MLP modulate the logit intensity of \textit{each subspace} or the \textit{modality arbitration margin} with attribution of logit difference~\cite{lv2024interpreting,ortu2024competition}:
\begin{align}
\delta_A^l(\mathcal{M}) &:= \mathcal{M}(\mathbf{H}^{l-1}_\text{inst} + \mathbf{A}^l_\text{inst}) 
- \mathcal{M}(\mathbf{H}^{l-1}_\text{inst}), \\
\delta_F^l(\mathcal{M}) &:= \mathcal{M}(\mathbf{H}^l_\text{inst}) - \mathcal{M}(\mathbf{H}^{l-1}_\text{inst} + \mathbf{A}^l_\text{inst}).
\end{align}
For robustness, we average these metrics across all samples. In the following, we focus on text-following results, as vision-following shows similar trends (see Apdx.~\ref{supp:mechanistic_dis_qwen}).
\begin{figure*}[t] %
    \centering
    \begin{subfigure}[b]{0.245\textwidth}
        \centering
        \includegraphics[page=1, width=\textwidth]{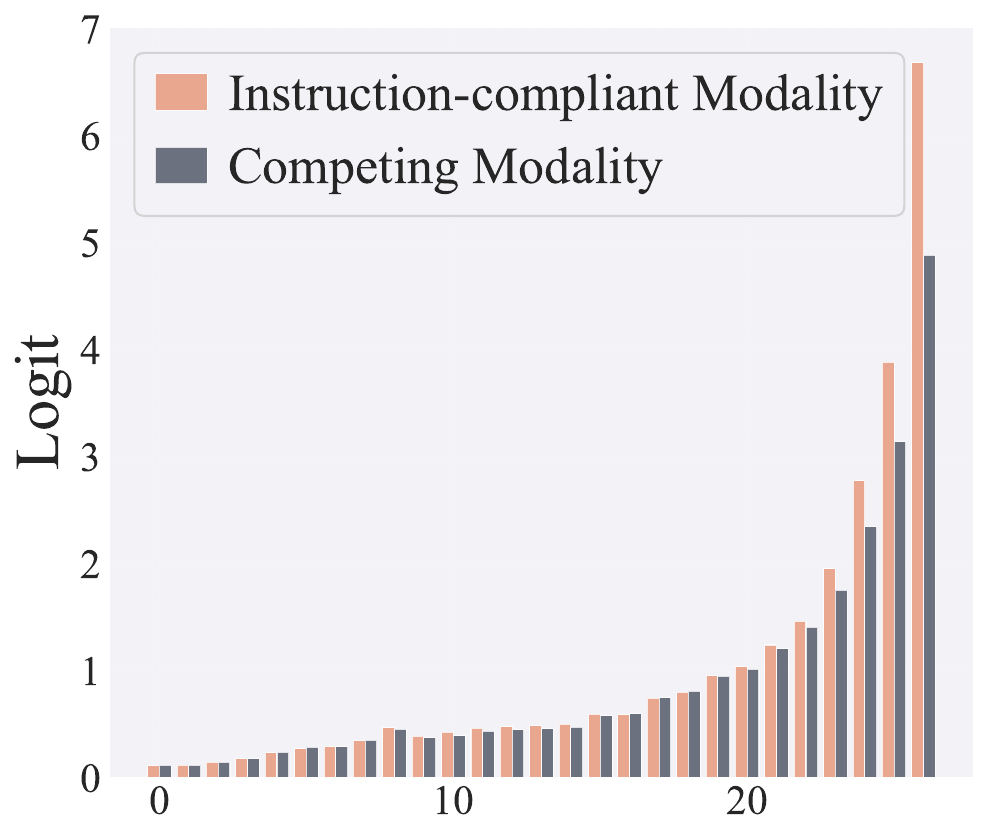}
        \vspace{-6mm}
        \caption{}
    \end{subfigure}
    \hfill 
    \begin{subfigure}[b]{0.245\textwidth}
        \centering
        \includegraphics[page=1, width=\textwidth]
        {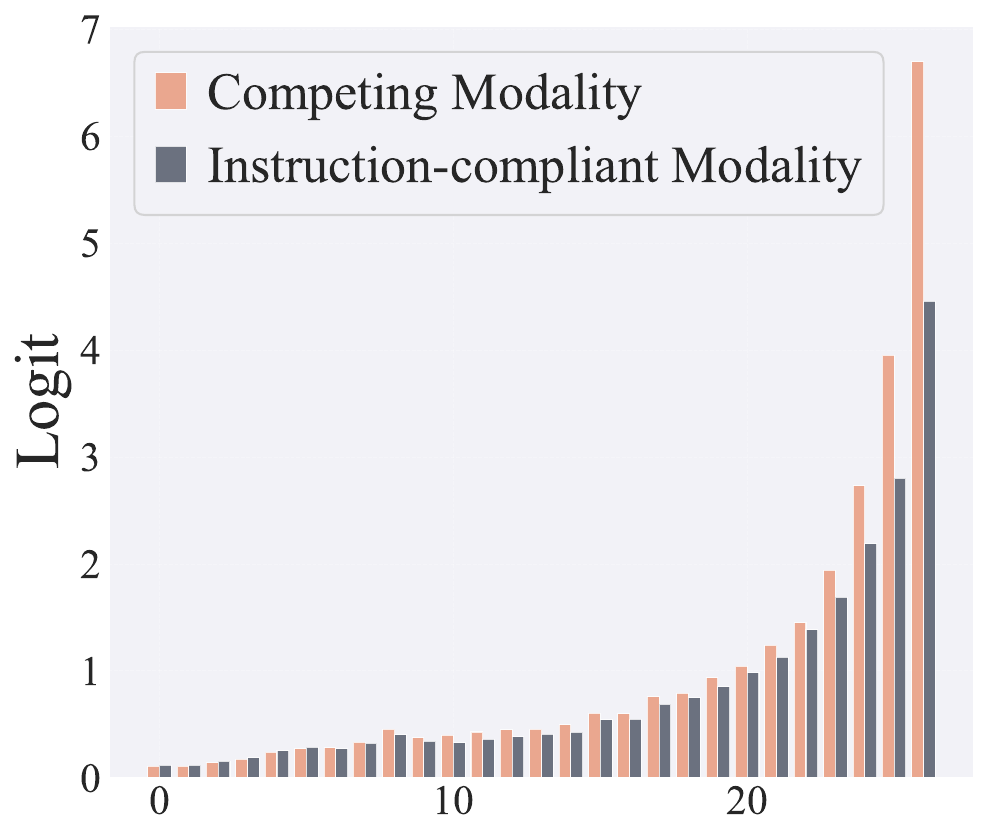}
        \vspace{-6mm}
        \caption{}
    \end{subfigure}
    \hfill 
    \begin{subfigure}[b]{0.245\textwidth}
        \centering
        \includegraphics[page=1, width=\textwidth]
        {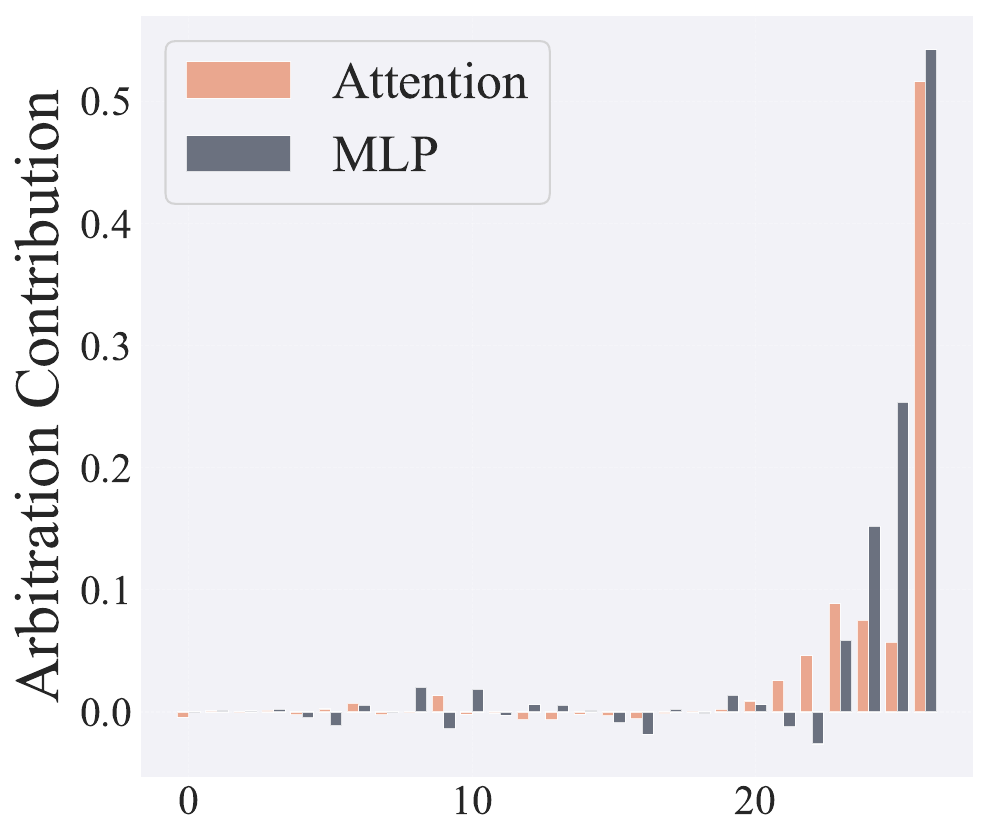}
         \vspace{-6mm}
        \caption{}
       
    \end{subfigure}
    \hfill 
    \begin{subfigure}[b]{0.245\textwidth}
        \centering
        \includegraphics[page=1, width=\textwidth]{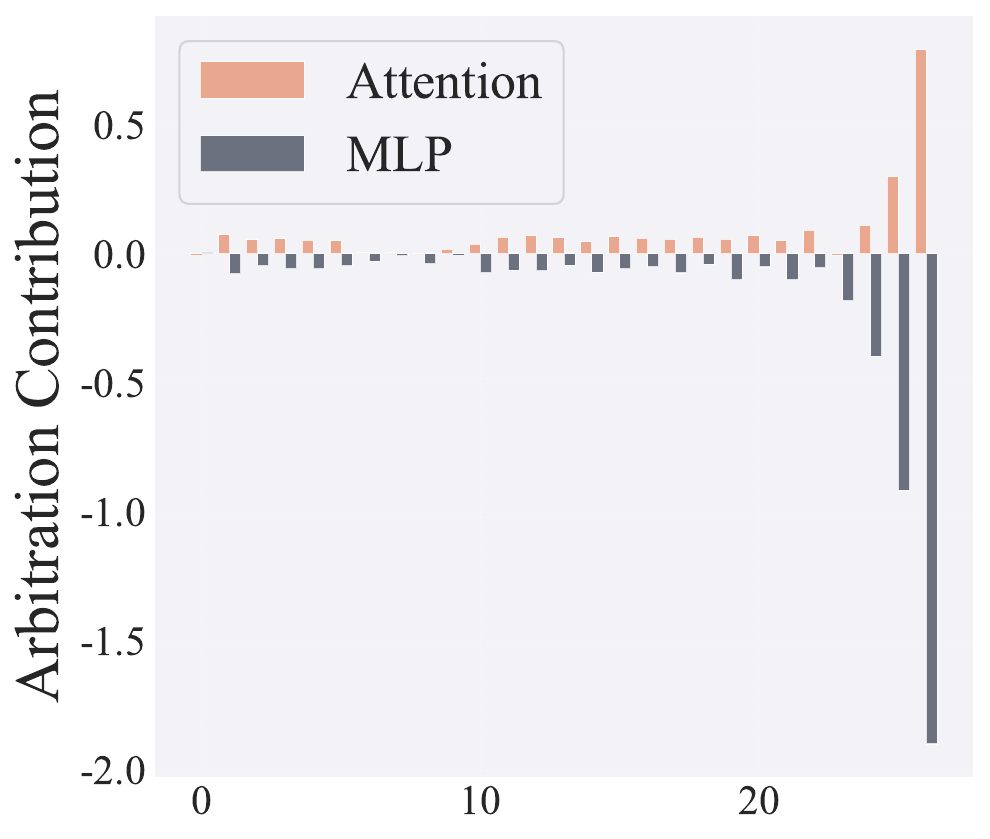}
        \vspace{-6mm}
        \caption{}
    \end{subfigure}
\caption{\textbf{Layer-wise component analysis.}  
(a,b) Layer-wise evolution of the instruction-compliant and competing subspace readouts for successful and failed modality-following samples, respectively.  
(c,d) Layer-wise contributions of attention and MLP to the modality arbitration margin for successful and failed samples, respectively. 
A positive contribution indicates that the corresponding component shifts the arbitration margin toward the instruction-specified intent.
}
    \label{fig:text_following_qwenvl}
\end{figure*}

\noindent\textbf{Emergence of Modality Decisions in Deep Layers.}   
We first track the layer-wise subspace evolution using $S_m(\mathbf{H}^l)$ in Fig.~\ref{fig:text_following_qwenvl} (a) and (b).
In shallow layers, both subspace signals are nearly indistinguishable, indicating that modality arbitration has not yet been resolved for either successful or failed samples. 
This matches the chance-level LDAR in the shallow layers, reported in \S\ref{sec:ldar_analysis}. 
In deeper layers, divergence emerges: for successful samples, the instruction-compliant subspace gradually dominates, whereas for failed samples, competing modality signals prevail. This suggests that the decision for modality arbitration is crystallized in deep layers, also consistent with the LDAR analysis.

\noindent\textbf{Attention Modulates Modality Arbitration Toward Instruction Intent.}  
Then we examine how attention and MLP modulate the modality arbitration margin using $\delta_A^l(\Delta S)$ and $\delta_F^l(\Delta S)$. As shown in Fig.~\ref{fig:text_following_qwenvl} (c,d), attention consistently exhibits positive modulation of the modality arbitration margin, regardless of successful or failed modality-following samples, while MLP does not show such consistent behavior. 
This demonstrates that attention tends to help resolve modality arbitration based on instruction intent.
These observations align with prior findings that attention in LLMs primarily orchestrates external context~\cite{ferrando-etal-2022-measuring}, whereas MLPs mainly encode internal parametric knowledge, which can occasionally counteract external signals~\cite{lv2024interpreting,zhao2025understanding}.

\begin{figure*}[t] %
    \centering
    \begin{subfigure}[b]{0.255\textwidth}
        \centering
        \includegraphics[page=1, width=\textwidth]{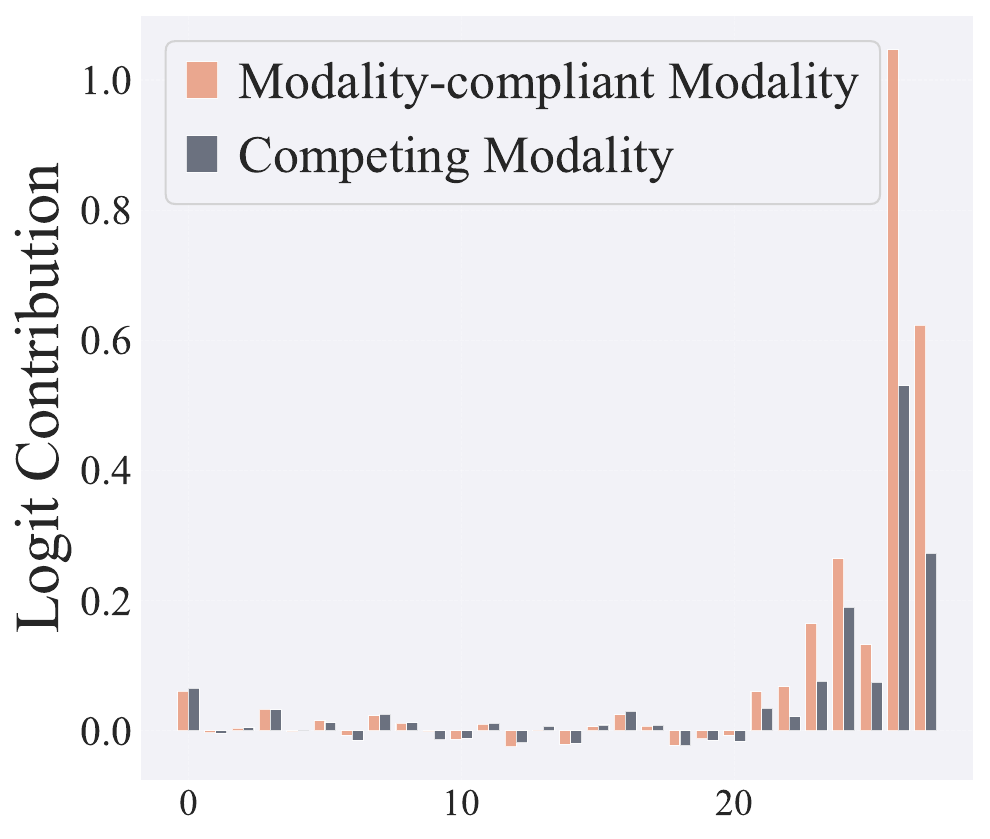}
        \vspace{-6mm}
        \caption{}
    \end{subfigure}
    \hfill 
    \begin{subfigure}[b]{0.255\textwidth}
        \centering
        \includegraphics[page=1, width=\textwidth]{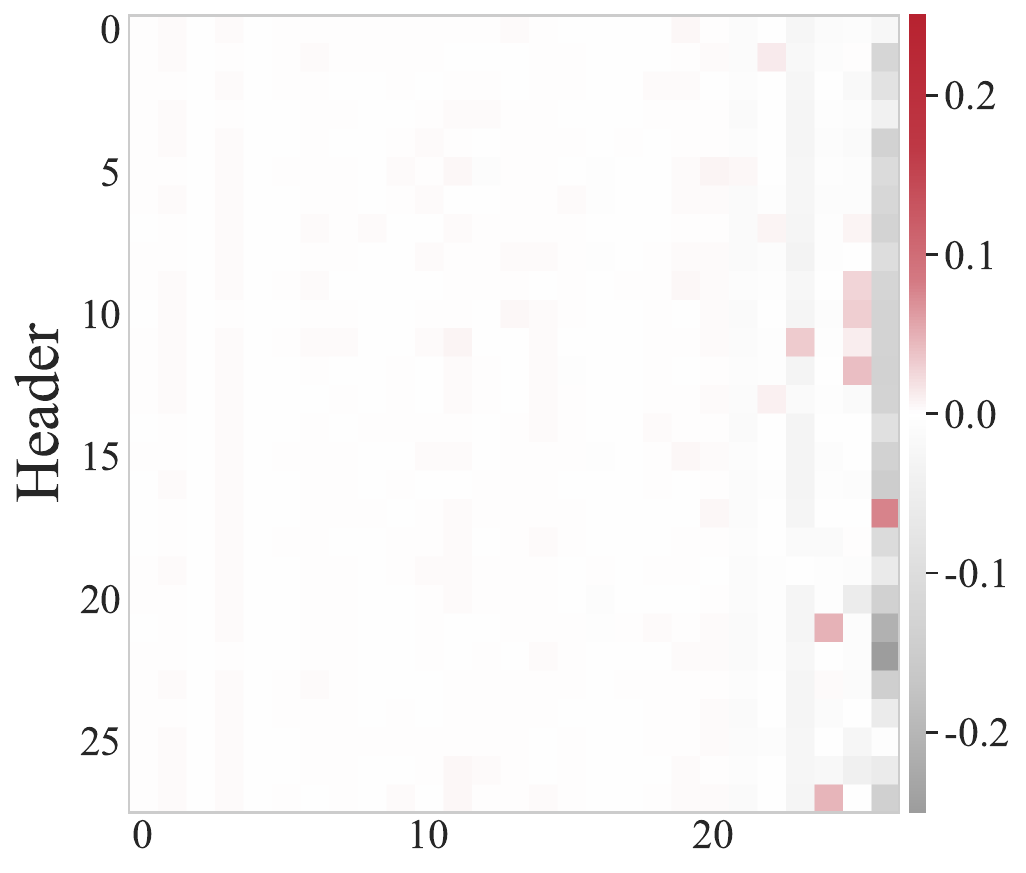}
        \vspace{-6mm}
        \caption{}
    \end{subfigure}
    \begin{subfigure}[b]{0.47\textwidth}
        \centering
        \includegraphics[page=1, width=\textwidth]{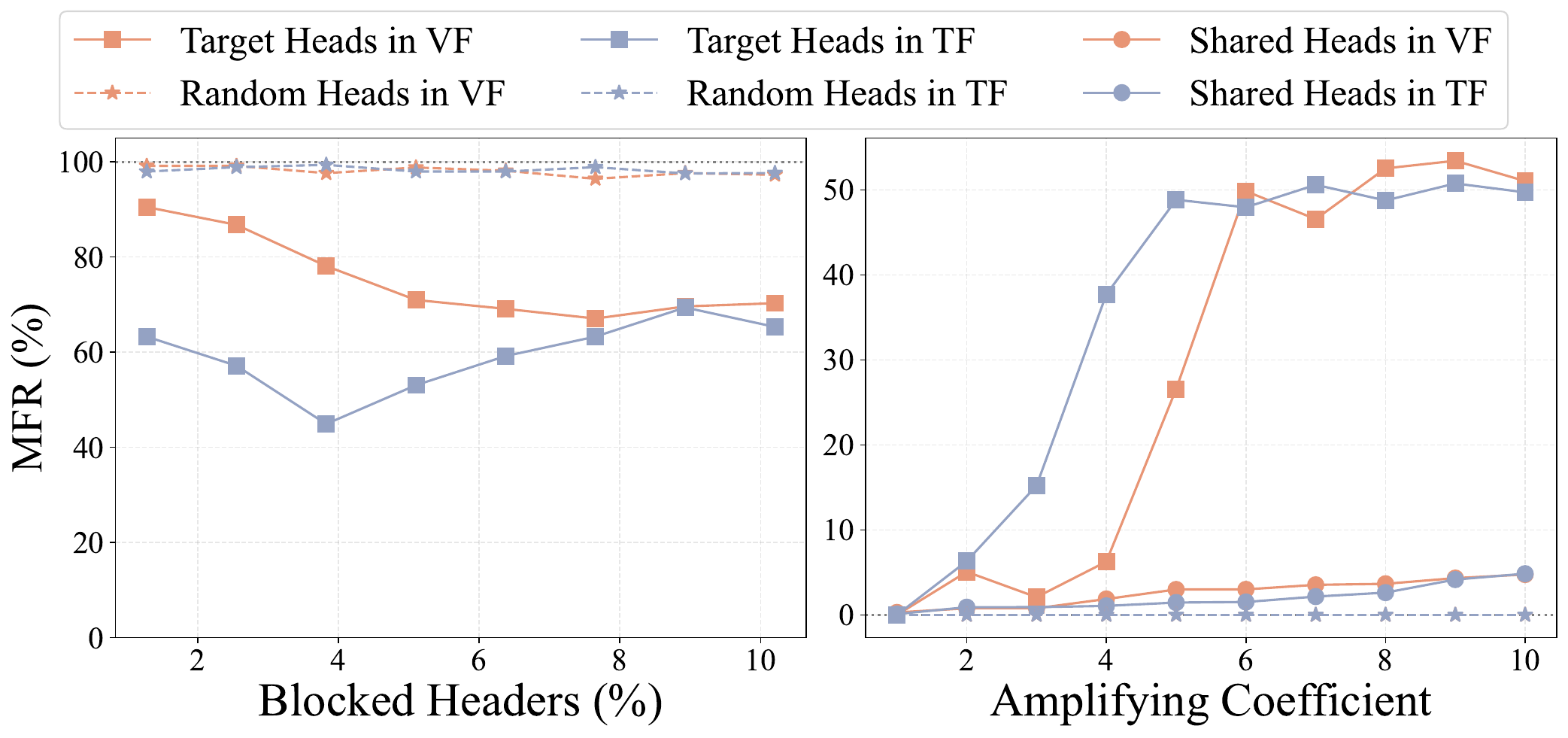}
        \vspace{-6mm}
        \caption{}
    \end{subfigure}
\caption{\textbf{Attention analysis and validation of functional specialization.}
(a) Layer-wise attention roles to the instruction-compliant and competing subspaces for successful text-following samples.
(b) Layer-wise roles of individual attention heads to the modality arbitration margin for successful text-following samples.
(c) Attention intervention analysis through attention head blocking (left) and amplification (right).
 We compare the Modality Following Ratio (MFR) under targeted, random, and shared-head interventions for Text Following (TF) and Vision Following (VF).
}
  \label{fig:header_qwenvl_analysis}
\end{figure*}

\noindent\textbf{Attention: From Buffering to Arbitration.}
Given the role of attention in resolving modality arbitration toward instruction intent, we further track its modulation of the individual instruction-compliant and competing subspaces using $\delta_A^l(\mathcal{S}_p)$ and $\delta_A^l(\mathcal{S}_c)$ for successful modality-following samples in Fig.~\ref{fig:header_qwenvl_analysis} (a).  
We find that in shallow layers, attention modulates both modalities to a similar extent. 
Correspondingly, its effect on the modality arbitration margin is near zero, indicating that shallow attention primarily performs undifferentiated information transfer, acting as a latent buffer.
In deep layers, attention selectively amplifies the instruction-compliant modality relative to the competing modality, driving the resolution of modality arbitration, consistent with Fig.~\ref{fig:text_following_qwenvl} (c). 

\noindent\textbf{Sparse Attention Head Arbitrators.}  
For finer analysis, we further decompose $\delta_A^l(\Delta S)$ into individual attention heads $\mathrm{Head}^{l,j}$ in Eq.~(\ref{eq:header_compute}):
\begin{equation}
    \delta_A^{l,j} = \Delta \mathcal{S}(\mathbf{H}^{l-1}_\text{inst} + \mathbf{Head}^{l,j}_\text{inst}) - \Delta \mathcal{S}(\mathbf{H}^{l-1}_\text{inst}).
\end{equation}
As shown in Fig.~\ref{fig:header_qwenvl_analysis} (b), only a small subset of deep-layer heads exhibit strong modulation of the arbitration margin. While vision-following and text-following tasks activate largely distinct heads, a few top-ranking heads overlap, indicating a set of modality-shared arbitrators.

\noindent\textbf{Takeaways.}  
Our analyses characterize how modality arbitration is internally resolved: shallow attention layers buffer multimodal cues, deep attention layers perform selective arbitration, with a sparse subset of attention heads driving this. This provides a detailed understanding of cross-modality evidence utilization and informs the design of efficient interventions to enhance modality-following behavior (\S\ref{sec:attention_head_verification}). We further verify that these can generalize to other MLLMs in Apdx.~\ref{supp:Dissection_generalization}.

\subsection{Validation of Functional Specialization of Attention Heads}
\label{sec:attention_head_verification}

Having identified a sparse set of attention heads as potential modality arbitrators, we evaluate their functional specialization to confirm the reliability of our analysis framework and demonstrate their effect on modality-following behavior.

\noindent\textbf{Experimental Setup.}  
For each attention head $j$ at layer $l$, we manipulate its output at instruction tokens ($\text{Head}_\text{inst}^{l,j}$) in two ways.  
1) \textit{Attention Blocking}: The head’s output is zeroed ($\text{Head}_\text{inst}^{l,j} = 0$) to test its necessity for modality-following behavior, while monitoring its impact on general language and vision capabilities.  
2) \textit{Attention Amplifying}: The head’s output is scaled by a factor $\alpha > 1$ to intensify its signal, demonstrating the potential to effectively modulate modality-following behavior.  

The Top-$G$ heads are selected based on their metric scores for text following in Fig.~\ref{fig:header_qwenvl_analysis} (b) and vision following in Fig.~\ref{fig:vision_following_qwenvl_further} (b) in Apdx.~\ref{supp:mechanistic_dis_qwen}.
We use the Modality Following Ratio (MFR) to quantify the effects of these interventions. We compare four configurations as follows.
1) Original: Standard inference without intervention.  
2) Targeted Heads: Interventions applied to the Top-$G$ heads identified by our framework.  
3) Random Heads: Interventions applied to an equal number of randomly selected heads.  
4) Shared Heads: Interventions applied only to modality-shared heads.  
For blocking experiments, we evaluate on samples where the model originally follows the instruction correctly (Original MFR = 100\%) to test the necessity of identified heads. For amplifying experiments, we evaluate on failure cases (Original MFR = 0\%) to test whether enhancing these heads can restore correct modality following.
General capabilities are assessed on both visual and text understanding benchmarks. 
For vision understanding tasks, the performance is measured on VQA task, TextVQA~\cite{textvqa} measured by Accuracy and image captioning task, Flickr30k~\cite{young2014image} measured by METEOR, CIDEr, SPICE, and PPL.  
For language capabilities: we evaluate pure text understanding task, MMLU~\cite{hendrycks2020measuring} measured by Accuracy and instruction following task, IFEval~\cite{zhou2023instruction} measured by Accuracy.

\noindent\textbf{Blocking Targeted Heads Suppresses Modality Following with Minimal Impact on General Capabilities.} 
As shown in the left panel of Fig.~\ref{fig:header_qwenvl_analysis} (c), progressively blocking targeted heads leads to a clear decline in MFR, whereas blocking the same number of randomly selected heads has minimal effect. 
Notably, blocking only the Top-40 heads (around 5\% of total heads) causes a dramatic MFR drop (around 60\% for text following), supporting the view that these heads play an important role in modulating modality-following behavior. Gradually increasing the number of blocked heads further highlights their coordinated effect. Blocking only the shared heads produces negligible changes, indicating that effective modality-following requires the modality-specific heads.
Since around 30 blocked heads are sufficient to substantially reduce modality-following performance, we restrict interventions to this number. Table~\ref{tab:block_general_capability} shows that blocking these heads has a negligible impact on general vision or language tasks, including the instruction following task, IFEval. This demonstrates the functional specificity of the identified attention heads for modality following.

\noindent\textbf{Targeted Amplification Enhances Modality Following.}  
As shown in the right panel of Fig.~\ref{fig:header_qwenvl_analysis} (c), amplifying the top 30 targeted heads substantially improves MFR, yielding an absolute gain of around 60\%. By contrast, applying the same amplification to random or shared heads leads to smaller changes.
Besides, MFR rises progressively with the amplification coefficient before reaching a plateau. The experiments in Table~\ref{tab:collapse} in Apdx.~\ref{supp:amplifying_cofficient}, indicate that excessive amplification (e.g., $\alpha = 18$) can impair modality-following by disrupting general language capabilities.

\noindent\textbf{Takeaways.} 
Taken together, these experiments confirm that the identified attention heads are functionally specialized and can effectively modulate modality-following behavior. This provides empirical validation for our mechanistic analysis of modality arbitration. 
We further verify that these patterns can generalize to other open-ended scenarios, such as modality interference tasks in Apdx.~\ref{supp:Dissection_generalization}. 
Finally, we also perform additional ablation studies on key design components $S_m$ in Eq.~(\ref{eq:s_l_m}) to assess the robustness of our diagnostic framework in Apdx.~\ref{supp:readout_function}.

\begin{table}[t]
\centering
\small
\caption{Impact of blocking identified attention heads on general capabilities of Qwen2.5-VL-7B for vision understanding with TextVQA (Acc) and Flickr30k (METEOR/ CIDEr/ SPICE/ PPL); language understanding with MMLU (Acc) and IFEval (Acc). Higher is better except for PPL.}
\label{tab:block_general_capability}
\begin{tabular}{lcccc}
\toprule
Method & TextVQA & Flickr30k & MMLU & IFEval \\
\midrule
Qwen2.5-VL-7B & 63.10  & 21.4/ 28.4/ 14.9/ 19.6  & 68.3 & 66.4\\
Block-\#10 & 62.95   & 21.5/ 28.0/ 14.9/ 19.1  & 68.5& 66.0\\
Block-\#20 & 63.26   & 21.4/ 26.9/ 14.9/ 18.4 & 67.8& 65.7\\
Block-\#30 & 63.32   & 21.4/ 28.5/ 14.8/ 20.3 & 66.9& 65.3\\
\bottomrule
\end{tabular}
\end{table}

\section{Discussion}
In this work, we adopt a cross-modal relay perspective to study how external multimodal contexts are integrated and leveraged, in contrast to prior studies that focus solely on information changes at the generated tokens.   
Based on these findings, we highlight two potential directions:  
1) \textbf{Efficient computation}: Insights from instruction-anchor–mediated context integration could inform context-compression strategies to achieve more efficient and effective context learning.  
2) \textbf{Architectural design for omni-modal models}: Current architectures involve substantial redundant computation, particularly in attention pathways. Alternative designs that better align computation with context usage may improve capability ceilings under constrained resources.


\section{Conclusion}

This paper investigates the underlying mechanisms of modality following in MLLMs through the lens of information flow. 
We identify instruction tokens as structural anchors where modality competition is resolved. 
Our analysis reveals a functional stratification within the transformer architecture: shallow attention layers act as latent buffers, while deep attention layers resolve modality arbitration based on instruction intent, with a sparse subset of attention heads driving this process.
Targeted attention-head interventions validate the functional specificity of these heads, thereby validating the robustness of our mechanistic framework. 
This work,provides a mechanistic account of modality following and informs future efforts to improve how MLLMs integrate and utilize multimodal evidence.

\newpage
{
    \small
    \bibliographystyle{plainurl}
    \bibliography{main}

@String(CVPR= {IEEE Conf. Comput. Vis. Pattern Recog.})

@String(ICCV= {Int. Conf. Comput. Vis.})

@String(ICLR = {Int. Conf. Learn. Represent.})

@String(AAAI = {AAAI})

@String(CVPR  = {CVPR})

@String(ICCV  = {ICCV})

@String(ICLR  = {ICLR})

@article{qwen2.5vl,
  title={Qwen2.5-vl technical report},
  author={Bai, Shuai and Chen, Keqin and Liu, Xuejing and Wang, Jialin and Ge, Wenbin and Song, Sibo and Dang, Kai and Wang, Peng and Wang, Shijie and Tang, Jun and others},
  journal={arXiv preprint arXiv:2502.13923},
  year={2025}
}

@article{zhang2025cross,
  title={Cross from Left to Right Brain: Adaptive Text Dreamer for Vision-and-Language Navigation},
  author={Zhang, Pingrui and Su, Yifei and Wu, Pengyuan and An, Dong and Zhang, Li and Wang, Zhigang and Wang, Dong and Ding, Yan and Zhao, Bin and Li, Xuelong},
  journal={arXiv preprint arXiv:2505.20897},
  year={2025}
}

@inproceedings{weifirst,
  title={First SFT, Second RL, Third UPT: Continual Improving Multi-Modal LLM Reasoning via Unsupervised Post-Training},
  author={Wei, Lai and Li, Yuting and Wang, Chen and Wang, Yue and Kong, Linghe and Huang, Weiran and Sun, Lichao},
  booktitle={The Thirty-ninth Annual Conference on Neural Information Processing Systems}
}

@inproceedings{chen2024internvl,
    title={Internvl: Scaling up vision foundation models and aligning for generic visual-linguistic tasks},
    author={Zhe, Chen and
                  Jiannan, Wu and
                  Wenhai, Wang and
                  Weijie, Su and
                  Guo, Chen and
                  Sen, Xing and
                  Muyan, Zhong and
                  Qinglong, Zhang and
                  Xizhou, Zhu and
                  Lewei, Lu and
                  Bin, Li and
                  Ping, Luo and
                  Tong, Lu and
                  Yu, Qiao and
                  Jifeng, Dai},
    booktitle={Proceedings of the IEEE/CVF Conference on Computer Vision and Pattern Recognition},
    pages={24185--24198},
    year={2024}
}

@article{gpt4,
  title={Gpt-4 technical report},
  author={Achiam, Josh and Adler, Steven and Agarwal, Sandhini and Ahmad, Lama and Akkaya, Ilge and Aleman, Florencia Leoni and Almeida, Diogo and Altenschmidt, Janko and Altman, Sam and Anadkat, Shyamal and others},
  journal={arXiv preprint arXiv:2303.08774},
  year={2023}
}

@InProceedings{Ding_2025_ICCV,
    author    = {Ding, Shengyuan and Wu, Shenxi and Zhao, Xiangyu and Zang, Yuhang and Duan, Haodong and Dong, Xiaoyi and Zhang, Pan and Cao, Yuhang and Lin, Dahua and Wang, Jiaqi},
    title     = {MM-IFEngine: Towards Multimodal Instruction Following},
    booktitle = {Proceedings of the IEEE/CVF International Conference on Computer Vision (ICCV)},
    month     = {October},
    year      = {2025},
    pages     = {1099-1109}
}

@article{wu2024deepseek,
  title={Deepseek-vl2: Mixture-of-experts vision-language models for advanced multimodal understanding},
  author={Wu, Zhiyu and Chen, Xiaokang and Pan, Zizheng and Liu, Xingchao and Liu, Wen and Dai, Damai and Gao, Huazuo and Ma, Yiyang and Wu, Chengyue and Wang, Bingxuan and others},
  journal={arXiv preprint arXiv:2412.10302},
  year={2024}
}

@article{cai2025diagnosing,
  title={Diagnosing and mitigating modality interference in multimodal large language models},
  author={Cai, Rui and Li, Bangzheng and Wen, Xiaofei and Chen, Muhao and Zhao, Zhe},
  journal={arXiv preprint arXiv:2505.19616},
  year={2025}
}

@inproceedings{ortu2024competition,
  title={Competition of mechanisms: Tracing how language models handle facts and counterfactuals},
  author={Ortu, Francesco and Jin, Zhijing and Doimo, Diego and Sachan, Mrinmaya and Cazzaniga, Alberto and Sch{\"o}lkopf, Bernhard},
  booktitle={Proceedings of the 62nd Annual Meeting of the Association for Computational Linguistics (Volume 1: Long Papers)},
  pages={8420--8436},
  year={2024}
}

@article{zhou2023instruction,
  title={Instruction-following evaluation for large language models},
  author={Zhou, Jeffrey and Lu, Tianjian and Mishra, Swaroop and Brahma, Siddhartha and Basu, Sujoy and Luan, Yi and Zhou, Denny and Hou, Le},
  journal={arXiv preprint arXiv:2311.07911},
  year={2023}
}

@article{lv2024interpreting,
  title={Interpreting key mechanisms of factual recall in transformer-based language models},
  author={Lv, Ang and Chen, Yuhan and Zhang, Kaiyi and Wang, Yulong and Liu, Lifeng and Wen, Ji-Rong and Xie, Jian and Yan, Rui},
  journal={arXiv preprint arXiv:2403.19521},
  year={2024}
}

@inproceedings{ferrando-etal-2022-measuring,
    title = "Measuring the Mixing of Contextual Information in the Transformer",
    author = "Ferrando, Javier  and
      G{\'a}llego, Gerard I.  and
      Costa-juss{\`a}, Marta R.",
    editor = "Goldberg, Yoav  and
      Kozareva, Zornitsa  and
      Zhang, Yue",
    booktitle = "Proceedings of the 2022 Conference on Empirical Methods in Natural Language Processing",
    month = dec,
    year = "2022",
    address = "Abu Dhabi, United Arab Emirates",
    publisher = "Association for Computational Linguistics",
    pages = "8698--8714",
}

@inproceedings{zhao2025understanding,
  title={Understanding parametric and contextual knowledge reconciliation within large language models},
  author={Zhao, Jun and Yang, Yongzhuo and Hu, Xiang and Tong, Jingqi and Lu, Yi and Wu, Wei and Gui, Tao and Zhang, Qi and Huang, Xuanjing},
  booktitle={The Thirty-ninth Annual Conference on Neural Information Processing Systems},
  year={2025}
}

@article{young2014image,
  title={From image descriptions to visual denotations: New similarity metrics for semantic inference over event descriptions},
  author={Young, Peter and Lai, Alice and Hodosh, Micah and Hockenmaier, Julia},
  journal={Transactions of the Association for Computational Linguistics},
  volume={2},
  pages={67--78},
  year={2014},
  publisher={MIT Press}
}

@article{hendrycks2020measuring,
  title={Measuring massive multitask language understanding},
  author={Hendrycks, Dan and Burns, Collin and Basart, Steven and Zou, Andy and Mazeika, Mantas and Song, Dawn and Steinhardt, Jacob},
  journal={arXiv preprint arXiv:2009.03300},
  year={2020}
}

@inproceedings{textvqa,
  title={Towards vqa models that can read},
  author={Singh, Amanpreet and Natarajan, Vivek and Shah, Meet and Jiang, Yu and Chen, Xinlei and Batra, Dhruv and Parikh, Devi and Rohrbach, Marcus},
  booktitle={Proceedings of the IEEE/CVF conference on computer vision and pattern recognition},
  pages={8317--8326},
  year={2019}
}

@inproceedings{dissecting,
    title = "Dissecting Recall of Factual Associations in Auto-Regressive Language Models",
    author = "Geva, Mor  and
      Bastings, Jasmijn  and
      Filippova, Katja  and
      Globerson, Amir",
    booktitle = "Proceedings of the 2023 Conference on Empirical Methods in Natural Language Processing",
    month = dec,
    year = "2023",
    address = "Singapore",
    publisher = "Association for Computational Linguistics",
    pages = "12216--12235",

}

@inproceedings{multiinstruct,
    title = "{M}ulti{I}nstruct: Improving Multi-Modal Zero-Shot Learning via Instruction Tuning",
    author = "Xu, Zhiyang  and
      Shen, Ying  and
      Huang, Lifu",
    booktitle = "Proceedings of the 61st Annual Meeting of the Association for Computational Linguistics (Volume 1: Long Papers)",
    month = jul,
    year = "2023",
    address = "Toronto, Canada",
    publisher = "Association for Computational Linguistics",
    pages = "11445--11465",
}

@inproceedings{dialogue,
    title = "Multimodal Dialogue Response Generation",
    author = "Sun, Qingfeng  and
      Wang, Yujing  and
      Xu, Can  and
      Zheng, Kai  and
      Yang, Yaming  and
      Hu, Huang  and
      Xu, Fei  and
      Zhang, Jessica  and
      Geng, Xiubo  and
      Jiang, Daxin",
    editor = "Muresan, Smaranda  and
      Nakov, Preslav  and
      Villavicencio, Aline",
    booktitle = "Proceedings of the 60th Annual Meeting of the Association for Computational Linguistics (Volume 1: Long Papers)",
    month = may,
    year = "2022",
    address = "Dublin, Ireland",
    publisher = "Association for Computational Linguistics",
    pages = "2854--2866",
}

@article{zheng2025locot2v,
  title={LoCoT2V-Bench: A Benchmark for Long-Form and Complex Text-to-Video Generation},
  author={Zheng, Xiangqing and Wu, Chengyue and Chen, Kehai and Zhang, Min},
  journal={arXiv preprint arXiv:2510.26412},
  year={2025}
}

@inproceedings{gui,
  title={GUIOdyssey: A comprehensive dataset for cross-app GUI navigation on mobile devices},
  author={Lu, Quanfeng and Shao, Wenqi and Liu, Zitao and Du, Lingxiao and Meng, Fanqing and Li, Boxuan and Chen, Botong and Huang, Siyuan and Zhang, Kaipeng and Luo, Ping},
  booktitle={Proceedings of the IEEE/CVF International Conference on Computer Vision},
  pages={22404--22414},
  year={2025}
}

@InProceedings{robot,
    author    = {Zheng, Jinliang and Li, Jianxiong and Liu, Dongxiu and Zheng, Yinan and Wang, Zhihao and Ou, Zhonghong and Liu, Yu and Liu, Jingjing and Zhang, Ya-Qin and Zhan, Xianyuan},
    title     = {Universal Actions for Enhanced Embodied Foundation Models},
    booktitle = {Proceedings of the IEEE/CVF Conference on Computer Vision and Pattern Recognition (CVPR)},
    month     = {June},
    year      = {2025},
    pages     = {22508-22519}
}

@InProceedings{Zhang_2025_ICCV,
    author    = {Zhang, Pingrui and Gao, Xianqiang and Wu, Yuhan and Liu, Kehui and Wang, Dong and Wang, Zhigang and Zhao, Bin and Ding, Yan and Li, Xuelong},
    title     = {MoMa-Kitchen: A 100K+ Benchmark for Affordance-Grounded Last-Mile Navigation in Mobile Manipulation},
    booktitle = {Proceedings of the IEEE/CVF International Conference on Computer Vision (ICCV)},
    month     = {October},
    year      = {2025},
    pages     = {6315-6326}
}

@article{wei2025mm,
  title={MM-LIMA: Less Is More for Alignment in Multi-Modal Datasets},
  author={Wei, Lai and Li, Xiaozhe and Jiang, Zihao and Huang, Weiran and Sun, Lichao},
  journal={Artificial Intelligence for Engineering},
  year={2025},
  publisher={Wiley Online Library}
}

@inproceedings{bitton2023visit,
  title={{VisIT-Bench}: A benchmark for vision-language instruction following inspired by real-world use},
  author={Bitton, Yonatan and Bansal, Hritik and Hessel, Jack and Shao, Rulin and Zhu, Wanrong and Awadalla, Anas and Gardner, Josh and Taori, Rohan and Schmidt, Ludwig},
  booktitle={NeurIPS, Datasets and Benchmarks},
  year={2023}
}

@inproceedings{qian2024mia,
  title={{MIA-Bench}: Towards better instruction following evaluation of multimodal llms},
  author={Qian, Yusu and Ye, Hanrong and Fauconnier, Jean-Philippe and Grasch, Peter and Yang, Yinfei and Gan, Zhe},
  booktitle={ICLR},
  year={2025}
}

@inproceedings{zhu-etal-2025-benchmarking,
    title = "Benchmarking and Improving Large Vision-Language Models for Fundamental Visual Graph Understanding and Reasoning",
    author = "Zhu, Yingjie  and
      Bai, Xuefeng  and
      Chen, Kehai  and
      Xiang, Yang  and
      Yu, Jun  and
      Zhang, Min",
    editor = "Che, Wanxiang  and
      Nabende, Joyce  and
      Shutova, Ekaterina  and
      Pilehvar, Mohammad Taher",
    booktitle = "Proceedings of the 63rd Annual Meeting of the Association for Computational Linguistics (Volume 1: Long Papers)",
    month = jul,
    year = "2025",
    address = "Vienna, Austria",
    publisher = "Association for Computational Linguistics",
    pages = "30678--30701",
}

@article{zhang2026mitigating,
  title={Mitigating Multimodal Hallucination via Phase-wise Self-reward},
  author={Zhang, Yu and Sun, Chuyang and Chen, Kehai and Bai, Xuefeng and Xiang, Yang and Zhang, Min},
  journal={arXiv preprint arXiv:2604.17982},
  year={2026}
}

@misc{chen2023sharegpt4vimprovinglargemultimodal,
      title={ShareGPT4V: Improving Large Multi-Modal Models with Better Captions}, 
      author={Lin Chen and Jinsong Li and Xiaoyi Dong and Pan Zhang and Conghui He and Jiaqi Wang and Feng Zhao and Dahua Lin},
      year={2023},
      archivePrefix={arXiv},
      primaryClass={cs.CV},
}

@article{chen2024allava,
  title={Allava: Harnessing gpt4v-synthesized data for lite vision-language models},
  author={Chen, Guiming Hardy and Chen, Shunian and Zhang, Ruifei and Chen, Junying and Wu, Xiangbo and Zhang, Zhiyi and Chen, Zhihong and Li, Jianquan and Wan, Xiang and Wang, Benyou},
  journal={arXiv preprint arXiv:2402.11684},
  year={2024}
}

@article{he2026empowering,
  title={Empowering Reliable Visual-Centric Instruction Following in MLLMs},
  author={He, Weilei and Ju, Feng and Fan, Zhiyuan and Min, Rui and Cheng, Minhao and Fung, Yi R},
  journal={arXiv preprint arXiv:2601.03198},
  year={2026}
}

@article{liang2026render,
  title={Render-in-the-Loop: Vector Graphics Generation via Visual Self-Feedback},
  author={Liang, Guotao and Wang, Zhangcheng and Hu, Juncheng and Zhou, Haitao and Xue, Ziteng and Zhang, Jing and Xu, Dong and Yu, Qian},
  journal={arXiv preprint arXiv:2604.20730},
  year={2026}
}

@article{liang2026vanim,
  title={VAnim: Rendering-Aware Sparse State Modeling for Structure-Preserving Vector Animation},
  author={Liang, Guotao and Wang, Zhangcheng and Wang, Chuang and Hu, Juncheng and Zhou, Haitao and Liu, Junhua and Zhang, Jing and Xu, Dong and Yu, Qian},
  journal={arXiv preprint arXiv:2605.01517},
  year={2026}
}

@article{zhang2025evaluating,
  title={Evaluating and steering modality preferences in multimodal large language model},
  author={Zhang, Yu and Ma, Jinlong and Hou, Yongshuai and Bai, Xuefeng and Chen, Kehai and Xiang, Yang and Yu, Jun and Zhang, Min},
  journal={arXiv preprint arXiv:2505.20977},
  year={2025}
}

@inproceedings{geva2022transformer,
  title={Transformer feed-forward layers build predictions by promoting concepts in the vocabulary space},
  author={Geva, Mor and Caciularu, Avi and Wang, Kevin and Goldberg, Yoav},
  booktitle={Proceedings of the 2022 conference on empirical methods in natural language processing},
  pages={30--45},
  year={2022}
}

@article{lu2023multi,
  title={The multi-modal fusion in visual question answering: a review of attention mechanisms},
  author={Lu, Siyu and Liu, Mingzhe and Yin, Lirong and Yin, Zhengtong and Liu, Xuan and Zheng, Wenfeng},
  journal={PeerJ Computer Science},
  volume={9},
  pages={e1400},
  year={2023},
  publisher={PeerJ Inc.}
}

@inproceedings{liu2023deja,
  title={Deja vu: Contextual sparsity for efficient llms at inference time},
  author={Liu, Zichang and Wang, Jue and Dao, Tri and Zhou, Tianyi and Yuan, Binhang and Song, Zhao and Shrivastava, Anshumali and Zhang, Ce and Tian, Yuandong and Re, Christopher and others},
  booktitle={International Conference on Machine Learning},
  pages={22137--22176},
  year={2023},
  organization={PMLR}
}

@article{huang2024miner,
  title={Miner: Mining the underlying pattern of modality-specific neurons in multimodal large language models},
  author={Huang, Kaichen and Huo, Jiahao and Yan, Yibo and Wang, Kun and Yue, Yutao and Hu, Xuming},
  journal={arXiv preprint arXiv:2410.04819},
  year={2024}
}

@article{yu2024understanding,
  title={Understanding multimodal llms: the mechanistic interpretability of llava in visual question answering},
  author={Yu, Zeping and Ananiadou, Sophia},
  journal={arXiv preprint arXiv:2411.10950},
  year={2024}
}

@article{nikankin2025same,
  title={Same Task, Different Circuits: Disentangling Modality-Specific Mechanisms in VLMs},
  author={Nikankin, Yaniv and Arad, Dana and Gandelsman, Yossi and Belinkov, Yonatan},
  journal={arXiv preprint arXiv:2506.09047},
  year={2025}
}

@article{basu2024understanding,
  title={Understanding information storage and transfer in multi-modal large language models},
  author={Basu, Samyadeep and Grayson, Martin and Morrison, Cecily and Nushi, Besmira and Feizi, Soheil and Massiceti, Daniela},
  journal={Advances in Neural Information Processing Systems},
  volume={37},
  pages={7400--7426},
  year={2024}
}

@inproceedings{li2025miv,
title={M{\texttwosuperior}{IV}: Towards Efficient and Fine-grained Multimodal In-Context Learning via Representation Engineering},
author={Yanshu Li and Yi Cao and Hongyang He and Qisen Cheng and Xiang Fu and Xi Xiao and Tianyang Wang and Ruixiang Tang},
booktitle={Second Conference on Language Modeling},
year={2025},
}

@inproceedings{li2026make,
  title={Make LVLMs Focus: Context-Aware Attention Modulation for Better Multimodal In-Context Learning},
  author={Li, Yanshu and Yang, Jianjiang and Yang, Ziteng and Li, Bozheng and Han, Ligong and He, Hongyang and Yao, Zhengtao and Chen, Yingjie Victor and Fei, Songlin and Liu, Dongfang and others},
  booktitle={Proceedings of the AAAI Conference on Artificial Intelligence},
  volume={40},
  number={8},
  pages={6610--6618},
  year={2026}
}

@article{neo2024towards,
  title={Towards interpreting visual information processing in vision-language models},
  author={Neo, Clement and Ong, Luke and Torr, Philip and Geva, Mor and Krueger, David and Barez, Fazl},
  journal={arXiv preprint arXiv:2410.07149},
  year={2024}
}

@article{lou2025sae,
  title={Sae-v: Interpreting multimodal models for enhanced alignment},
  author={Lou, Hantao and Li, Changye and Ji, Jiaming and Yang, Yaodong},
  journal={arXiv preprint arXiv:2502.17514},
  year={2025}
}

@article{survey1,
  title={Explainable and interpretable multimodal large language models: A comprehensive survey},
  author={Dang, Yunkai and Huang, Kaichen and Huo, Jiahao and Yan, Yibo and Huang, Sirui and Liu, Dongrui and Gao, Mengxi and Zhang, Jie and Qian, Chen and Wang, Kun and others},
  journal={arXiv preprint arXiv:2412.02104},
  year={2024}
}

@inproceedings{ben2024lvlm,
  title={Lvlm-intrepret: An interpretability tool for large vision-language models},
  author={Ben Melech Stan, Gabriela and Aflalo, Estelle and Rohekar, Raanan Yehezkel and Bhiwandiwalla, Anahita and Tseng, Shao-Yen and Olson, Matthew Lyle and Gurwicz, Yaniv and Wu, Chenfei and Duan, Nan and Lal, Vasudev},
  booktitle={Proceedings of the IEEE/CVF Conference on Computer Vision and Pattern Recognition},
  pages={8182--8187},
  year={2024}
}

@inproceedings{pan2024finding,
  title={Finding and editing multi-modal neurons in pre-trained transformers},
  author={Pan, Haowen and Cao, Yixin and Wang, Xiaozhi and Yang, Xun and Wang, Meng},
  booktitle={Findings of the Association for Computational Linguistics: ACL 2024},
  pages={1012--1037},
  year={2024}
}

@article{yan2026knowledgeinferencescalinglaws,
  title={From Knowledge to Inference: Scaling Laws of Specialized Reasoning on GlobalHealthAtlas},
  author={Yan, Zhaokun and Liu, Zhaohan and Dong, Wuzheng and Feng, Lijie and Dai, Chengxiao},
  journal={arXiv preprint arXiv:2602.00491},
  year={2026}
}

@article{zhao2026clauseagenticneurosymbolicknowledge,
  title={CLAUSE: Agentic Neuro-Symbolic Knowledge Graph Reasoning via Dynamic Learnable Context Engineering},
  author={Zhao, Yang and Dai, Chengxiao and Zhuo, Wei and Xiu, Yue and Niyato, Dusit},
  journal={arXiv preprint arXiv:2509.21035},
  year={2025}
}

@article{vaswani2017attention,
  title={Attention is all you need},
  author={Vaswani, Ashish and Shazeer, Noam and Parmar, Niki and Uszkoreit, Jakob and Jones, Llion and Gomez, Aidan N and Kaiser, {\L}ukasz and Polosukhin, Illia},
  journal={Advances in neural information processing systems},
  volume={30},
  year={2017}
}

@article{chen2025some,
  title={Some Modalities are More Equal Than Others: Decoding and Architecting Multimodal Integration in MLLMs},
  author={Chen, Tianle and Chakka, Chaitanya and Akula, Arjun Reddy and Thomas, Xavier and Ghadiyaram, Deepti},
  journal={arXiv preprint arXiv:2511.22826},
  year={2025}
}

@article{leng2024curse,
  title={The curse of multi-modalities: Evaluating hallucinations of large multimodal models across language, visual, and audio},
  author={Leng, Sicong and Xing, Yun and Cheng, Zesen and Zhou, Yang and Zhang, Hang and Li, Xin and Zhao, Deli and Lu, Shijian and Miao, Chunyan and Bing, Lidong},
  journal={arXiv preprint arXiv:2410.12787},
  year={2024}
}

@article{guo2025aligned,
  title={Aligned better, listen better for audio-visual large language models},
  author={Guo, Yuxin and Ma, Shuailei and Ma, Shijie and Bao, Xiaoyi and Xie, Chen-Wei and Zheng, Kecheng and Weng, Tingyu and Sun, Siyang and Zheng, Yun and Zou, Wei},
  journal={arXiv preprint arXiv:2504.02061},
  year={2025}
}

@inproceedings{coco,
  title={Microsoft coco: Common objects in context},
  author={Lin, Tsung-Yi and Maire, Michael and Belongie, Serge and Hays, James and Perona, Pietro and Ramanan, Deva and Doll{\'a}r, Piotr and Zitnick, C Lawrence},
  booktitle={Computer vision--ECCV 2014: 13th European conference, zurich, Switzerland, September 6-12, 2014, proceedings, part v 13},
  pages={740--755},
  year={2014},
  organization={Springer}
}

@article{deepseek,
  title={Deepseek-v3 technical report},
  author={Liu, Aixin and Feng, Bei and Xue, Bing and Wang, Bingxuan and Wu, Bochao and Lu, Chengda and Zhao, Chenggang and Deng, Chengqi and Zhang, Chenyu and Ruan, Chong and others},
  journal={arXiv preprint arXiv:2412.19437},
  year={2024}
}
}

\clearpage
\newpage
\appendix
\section*{Appendices}

Our supplementary materials are summarized as follows:
\begin{itemize} 
    \item Appendix~\ref{supp:disc}: Limitations, Use of LLM, Impact Statement and License of Assets.
    \item Appendix~\ref{supp:dataset_construction}: Detailed Diagnostic Setup.
    \item Appendix~\ref{supp:more_attention_knockout}: More Results for Attention Knockout Analysis.
    \item Appendix~\ref{supp:more_results_mechanistic_dissection}: More Results for Mechanistic Dissection of Modality Arbitration.
    \item Appendix~\ref{supp:method_ablation}:
    Ablation Studies and Robustness Analysis.
\end{itemize}

\section{Discussion}
\label{supp:disc}

\subsection{Limitations}
This paper investigates the underlying mechanisms of modality following in MLLMs through the lens of information flow. 
Our analysis reveals a functional stratification within the transformer architecture: shallow attention layers act as latent buffers, while deep attention layers resolve modality arbitration based on instruction intent.
However, we recognize that a more microscopic investigation into neuron-level activation patterns could potentially uncover even more fundamental principles of cross-modal arbitration. We leave this fine-grained circuit decomposition—transitioning from functional heads to atomic neurons—for future research to further refine the theoretical boundaries of multimodal integration.

\subsection{Use of LLM}
In this work, we leveraged DeepSeek-V3~\citep{deepseek} to curate and generate an answer entity dictionary, which served as the foundation for constructing our analysis dataset. Furthermore, we conducted a mechanistic interrogation of modality-following behaviors in several state-of-the-art MLLMs, including Qwen2.5-VL-7B~\citep{qwen2.5vl}, InternVL3-8B~\citep{chen2024internvl}, InternVL3-14B~\cite{chen2024internvl} and DeepSeek-VL2-Tiny~\cite{wu2024deepseek}.
Additionally, large language models were employed to assist with grammatical refinement and linguistic polishing of the manuscript.

\subsection{Impact Statement}
This study elucidates the mechanisms of modality following in MLLMs, revealing how instruction-driven arbitration governs the resolution and prioritization of multimodal inputs. While these insights highlight potential vulnerabilities where safety filters might be bypassed, they primarily establish a structural foundation for developing more robust
and transparent AI safeguards.

\subsection{License of Assets}
All images used are publicly available from COCO~\citep{coco}.
We release our analysis under a Creative Commons Attribution 4.0 License (CC BY 4.0) to enhance global accessibility and foster innovation and collaboration in research.

\section{Detailed Diagnostic Setup}
\label{supp:dataset_construction}
We construct our analysis dataset based on MC$^2$~\cite{zhang2025evaluating}, which includes various visual and textual evidences corresponding to different answer candidates For each sample, we instantiate a modality-following sample by attaching an explicit modality following instruction that specifies which source the model should follow, e.g., ``You should follow the textual context rather than the visual content.'' Under the abstraction used in the main text, this yields a modality-following sample:
\[
S = \langle C_p, C_c, I, A_p, A_c, \mathcal{E}_p, \mathcal{E}_c \rangle,
\]
where $I$ denotes the active instruction specifying the target modality, $C_p$ and $C_c$ are the instruction-compliant and competing contexts, and $A_p$, $A_c$ are the corresponding answers. The answer entity dictionaries $\mathcal{E}_p$ and $\mathcal{E}_c$ aggregate up to ten semantically equivalent surface forms to robustly track modality-specific signals.
To support robust latent-signal tracing under surface-form variation, we construct the Answer Entity Dictionaries \(\mathcal{E}_p\) and \(\mathcal{E}_c\) from the canonical answers \(A_p\) and \(A_c\) based on lexical resources and LLMs~\cite{deepseek}. The construction process consists of two stages:

\noindent\textbf{1. Candidate Retrieval.} We first use WordNet via NLTK to retrieve semantically related candidate entities for each canonical answer.

\noindent\textbf{2. Semantic Verification.} We then filter these candidates using DeepSeek-V3~\cite{deepseek} to ensure strict semantic alignment with the canonical answer in the sample-specific context.
The semantic verification step uses the following prompt template:

\begin{tcolorbox}[colback=gray!5!white,colframe=black!75!white,title=Validate candidate answers using DeepSeek-V3]
Instruction: \\
\# You are an expert English linguist helping to judge if a candidate word can be a synonym of a label in a specific context.

Your task:
1. You are given:
   - a natural language question (the context),
   - an original label (the target meaning),
   - a single candidate word to evaluate.
2. Decide if the candidate word can express approximately the same meaning as the label in this question's context.
3. The candidate word may differ in tense or part of speech (noun/verb/adjective/etc.) but should still preserve the core meaning.
4. Answer "yes" if the candidate word is a valid synonym/near-synonym in this context, "no" otherwise.

Output format:
You MUST respond using the following tags:
\textless answer\textgreater yes \textless /answer\textgreater\ or \textless answer\textgreater no \textless /answer\textgreater \\
\textless reason\textgreater brief English explanation of your judgment \textless /reason\textgreater

Do NOT output anything other than these tags. \\
Question: \{\textbf{question}\} \\
Original label: \{\textbf{label}\} \\
Candidate word: \{\textbf{candidate}\} \\
Instructions: In the context of the Question, can the candidate word be considered a valid synonym or near-synonym of the Original label? Answer yes or no using the required tags.
\end{tcolorbox}

Preliminary experiments further showed that intermediate hidden states are often decoded into Chinese tokens. To improve coverage in latent-space readout, we therefore augment each dictionary with Chinese candidate expressions generated by DeepSeek-V3 using the following prompt template:

\begin{tcolorbox}[colback=gray!5!white,colframe=black!75!white,title=Generate Chinese candidate answers using DeepSeek-V3]
Instruction: \\
\# You are a Chinese linguist expert assisting in generating Chinese synonyms or related terms based on English vocabulary and specific contexts.

Your task:

You are given:

An English word (label),

An English question (the context).

Understand the specific meaning of the English word within the context of the question.

Generate a list of Chinese synonyms or related terms, ensuring that:

The Chinese terms accurately convey the core meaning of the English word in the given context.

They may include synonyms, near-synonyms, or related expressions.

Return only individual words or phrases; do not provide full sentences.

The quantity should be controlled between 5 and 15 terms.

Output format: You MUST respond using the following tag format:
\textless chinese words \textgreater word1, word2, word3, ... \textless /chinese words \textgreater \\
\textless explanation \textgreater A brief explanation in Chinese describing how these terms correspond to the English word's meaning \textless /explanation \textgreater

Do NOT output anything other than these tags. \\
Question: \{\textbf{question}\} \\
Original label: \{\textbf{label}\} \\
Instructions: Generate a list of corresponding Chinese synonyms or related terms based on the meaning of this English word within the context of the question. Output the results using the required tag format.
\end{tcolorbox}

To ensure data quality, we employ both LLM verification and manual verification methods:
1) Verification of LLM: We utilize ChatGPT-4o-mini~\cite{gpt4} to verify whether each candidate word is appropriate for the given question;
2) Human Verification: We engaged three students in the field of artificial intelligence to conduct cross-validation of the candidate words. Only when all three agreed that the candidate word matched the question, did we consider it as a valid candidate.

Finally, each canonical answer is ultimately associated with up to 10 bilingual synonym candidates.
After filtering out samples with insufficient answer-set coverage, the final dataset contains 2,000 instances for analysis. 
We also verify that the analysis dataset is sufficient for robust modality-following analysis, the results in Fig.~\ref{fig:vision_text_block_qwenvl_w3_dataset_scale} in Apdx.~\ref{supp:generalization_causal} demonstrate that consistent results are obtained when the analysis is conducted on smaller subsets and different instruction styles. All experiments are implemented on a single H20-80G GPU.

\section{More Results for Causal Attention Analysis}
\label{supp:more_attention_knockout}
In this section, we provide more results for causal attention analysis including 
the results for sensitivity analysis of attention knockout windows in Apdx.~\ref{supp:app_window},
additional evidence that instruction convergence is not a positional artifact in Apdx.~\ref{supp:semantic_constraint},
more results for generalization and robustness of attention routing in Apdx.\ref{supp:generalization_causal} including more results for larger dense MLLM and MoE-based MLLM, the results of generalization to open-ended, multi-token generation tasks and
the sensitivity analysis for dataset scale and instruction styles.

\begin{figure*}[h] 
    \centering
    \begin{subfigure}[b]{0.495\textwidth}
        \centering
        \includegraphics[page=1, width=\textwidth]{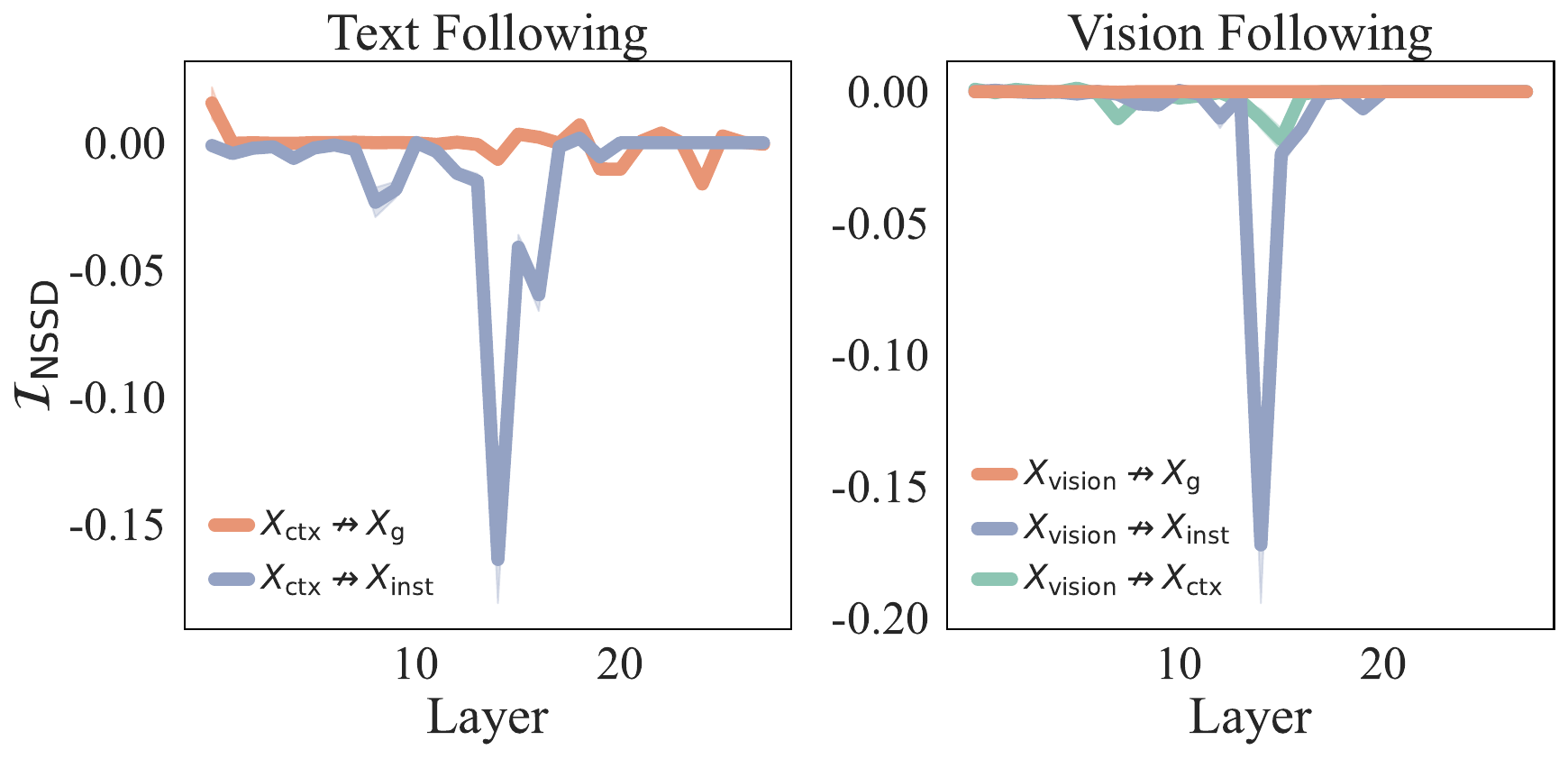}
        \caption{Window Size $1$}
    \end{subfigure}
    \hfill 
    \begin{subfigure}[b]{0.495\textwidth}
        \centering
        \includegraphics[page=1, width=\textwidth]{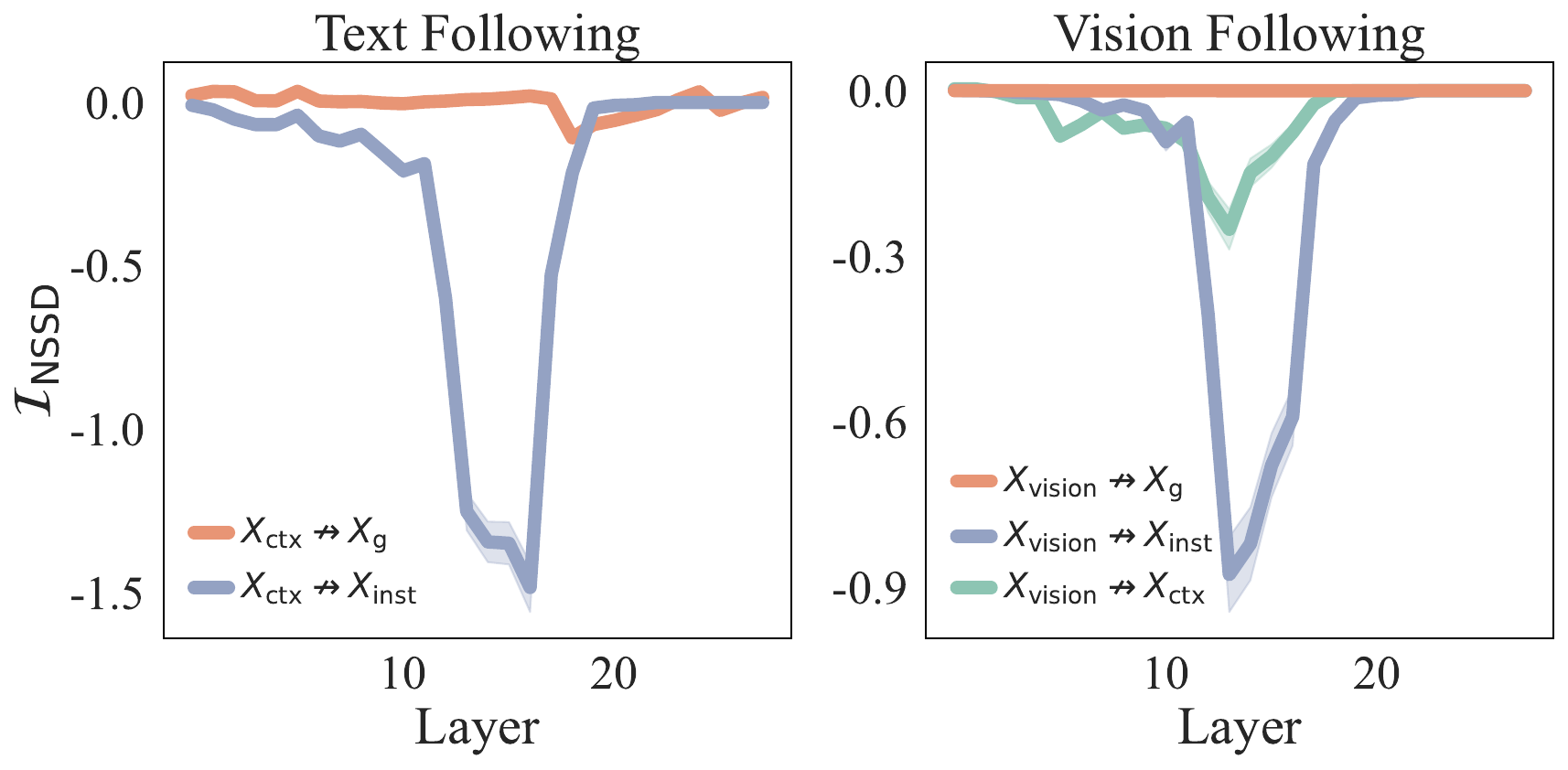}
        \caption{Window Size $5$.}
    \end{subfigure}
    \caption{
\textbf{Layer-wise $\mathcal{I}_\text{NSSD}$ under attention-knockout for different window sizes.}  
Each curve represents cutting a attention pathway, where $X_\text{vision}$, $X_\text{ctx}$, $X_\text{inst}$, and $X_\text{g}$ denote visual tokens, text context tokens, instruction tokens, and generated tokens.
Cutting pathways to $X_\text{inst}$ cause larger negative shifts than those to $X_\text{g}$, highlighting instruction tokens as the central convergence site for modality cues.}

    \label{fig:window_size}
\end{figure*}
\subsection{Sensitivity Analysis of Attention Knockout Windows}
\label{supp:app_window}
In \S\ref{sec:casual_attention_knockout}, our primary analysis utilized a default attention knockout window $3$. To ensure the robustness of our mechanistic findings and confirm that the observed trends are not artifacts of a specific window configuration, we conducted a sensitivity analysis across various window sizes. 
Specifically, maintaining the experimental setup described in Fig.~\ref{fig:vision_inst_block_comparison}, we report the $\mathcal{I}_\text{NSSD}$ results of Qwen2.5VL-7B for alternative window sizes in Fig.~\ref{fig:window_size}. We observe that while the absolute values of $\mathcal{I}_\text{NSSD}$ fluctuate slightly as the window expands, the fundamental conclusions regarding attention patterns remain consistent, further validating the stability of our method.

\subsection{Additional Evidence that Instruction Convergence Is Not a Positional Artifact}
\label{supp:semantic_constraint}
Beyond the evidence in Fig.~\ref{fig:xxxx} (a) of the main text, we present four experiments in a more compact form to validate that instruction tokens function as genuine semantic anchors rather than positional artifacts. 

\begin{figure}[h!]
    \centering
    \begin{subfigure}[b]{0.48\textwidth}
        \centering
        \includegraphics[page=1, width=\textwidth]{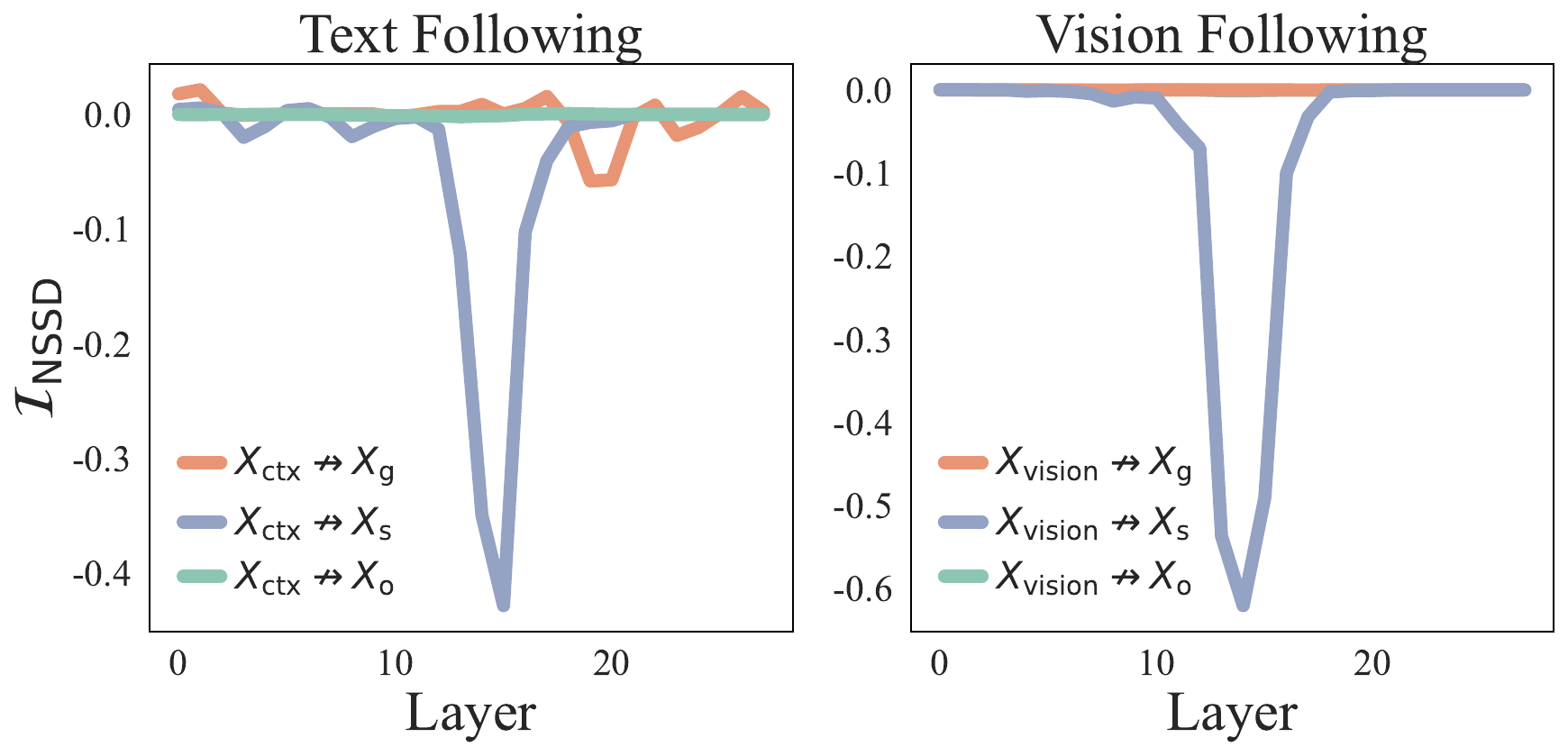}
        \caption{Semantic instruction tokens ($X_s$) are placed before the output-format constraint instruction tokens ($X_o$).}
    \end{subfigure}
    \hfill
    \begin{subfigure}[b]{0.48\textwidth}
        \centering
        \includegraphics[page=1, width=\textwidth]{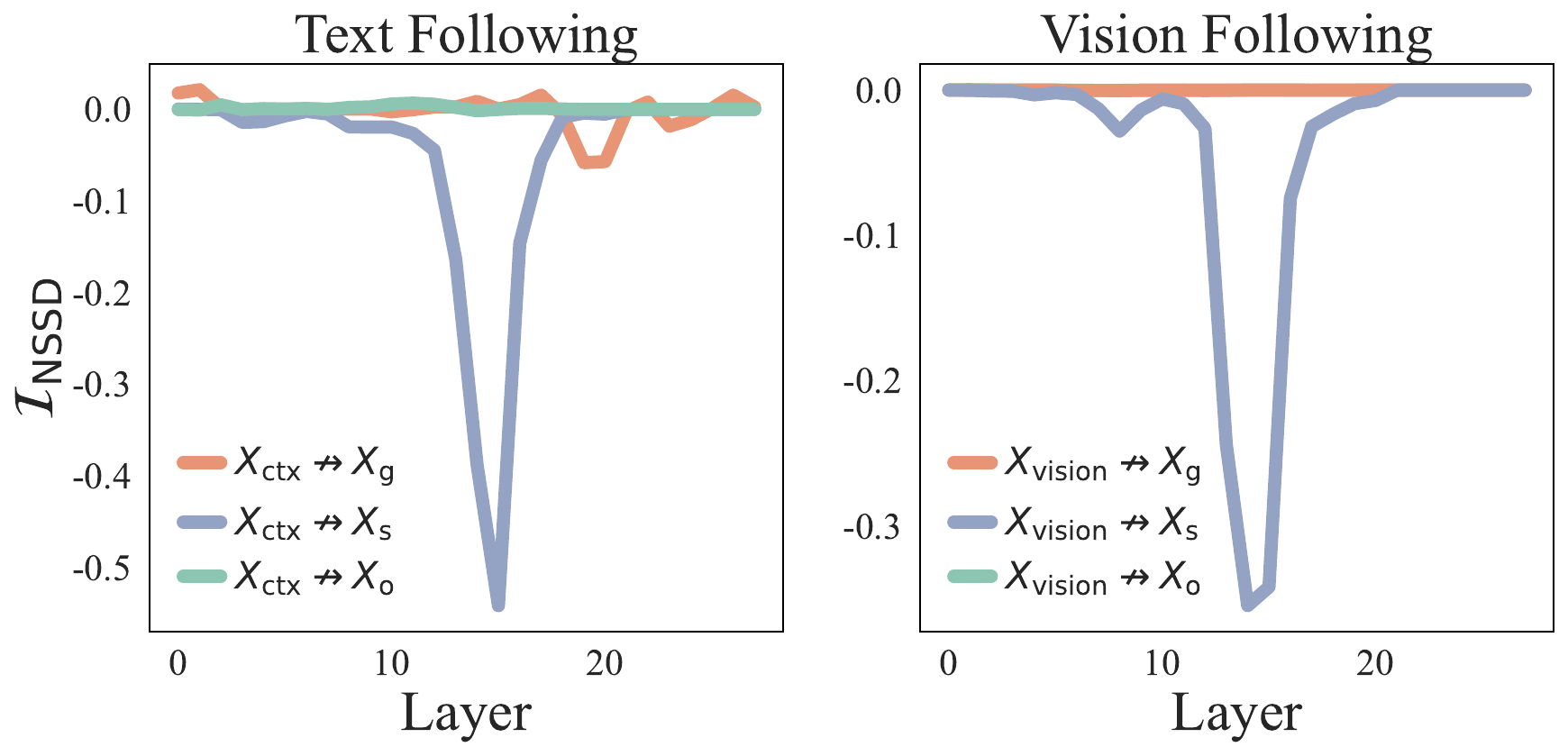}
        \caption{Semantic instruction tokens ($X_s$) are placed after the output-format constraint instruction tokens ($X_o$).}
    \end{subfigure}
    \vspace{0.3cm}

    \begin{subfigure}[b]{0.48\textwidth}
        \centering
        \includegraphics[page=1, width=\textwidth]{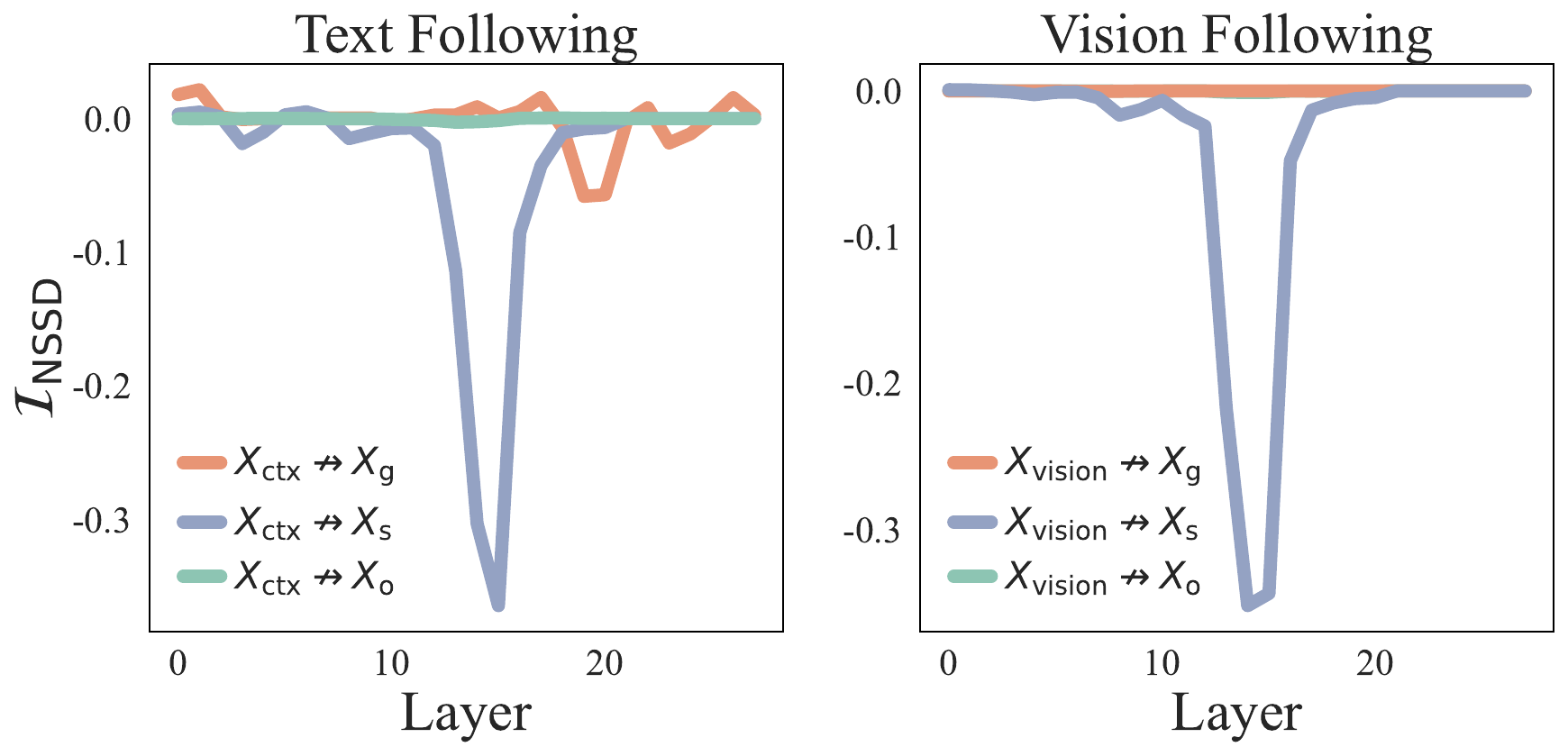}
        \caption{Adding system prompt into the instruction tokens, \textbf{which are placed before semantic tokens and previous constraint tokens}. $X_s$ comprises original semantic tokens and the inserted system prompt.}
    \end{subfigure}
    \hfill
    \begin{subfigure}[b]{0.48\textwidth}
        \centering
        \includegraphics[page=1, width=\textwidth]{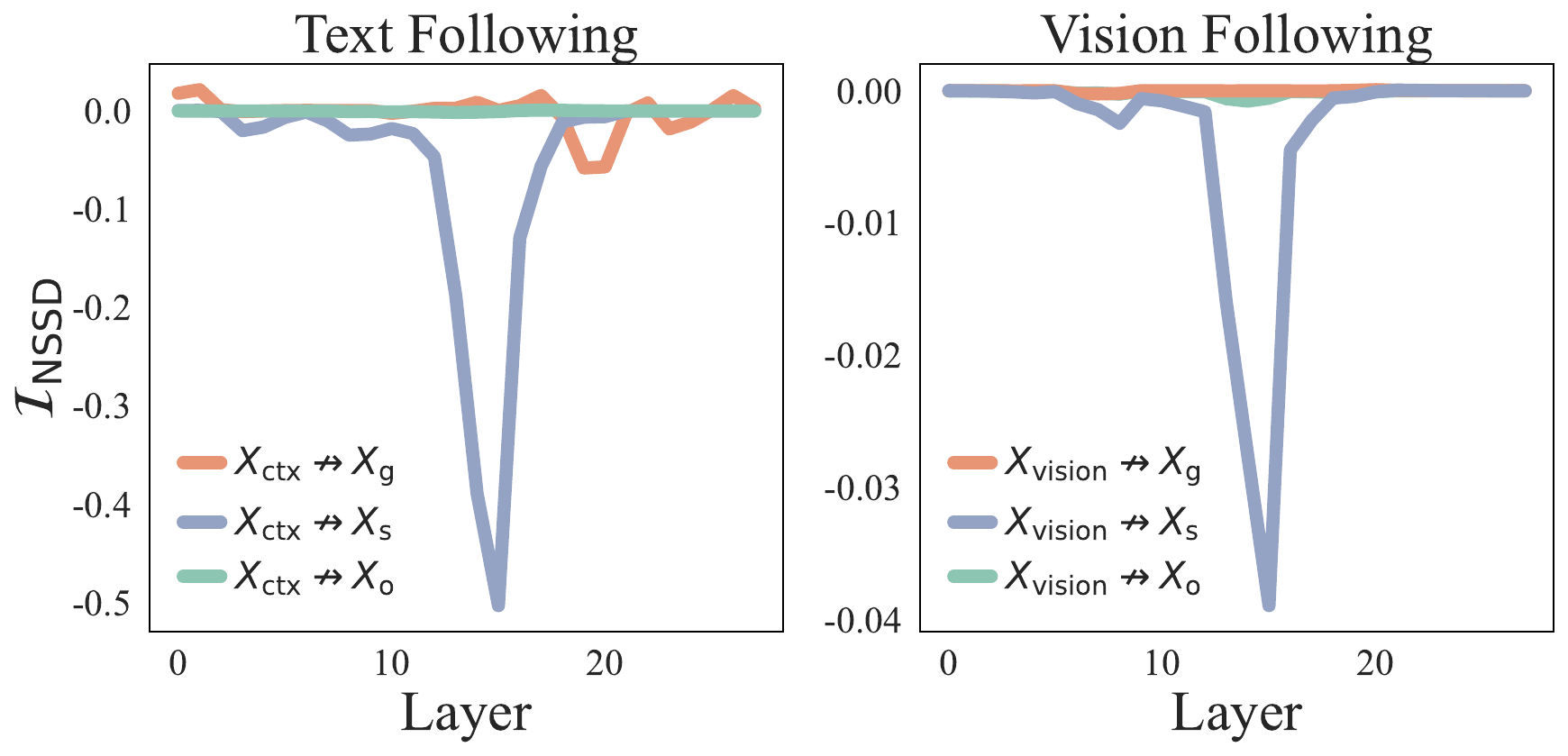}
        \caption{Adding system prompt into instruction tokens, \textbf{which are placed in the middle of semantic tokens and previous constraint tokens}. $X_s$ comprises original semantic tokens and the inserted system prompt.}
    \end{subfigure}
    \caption{Additional knockout attention analyses verifying that modality-specifying semantic tokens serve as true semantic anchors rather than positional artifacts. Here, the newly introduced system prompt and the original constraint tokens are both treated as $X_o$.}
    \label{fig:semantic_constraint_fig}
\end{figure}

First, we examined the effect of swapping the positions of modality-specifying semantic tokens ($X_s$) and output-format constraint tokens ($X_o$) within the input sequence. knockout attention analysis was performed to measure the impact of cutting attention from modality cues to $X_s$ or $X_o$. The results, shown in Fig.~\ref{fig:semantic_constraint_fig} (a,b), indicate that cutting attention to $X_s$ substantially reduces modality-following performance, whereas cutting attention to $X_o$ has only a marginal effect.

Second, to further rule out positional confounds, we introduced a longer general system prompt (``You are a helpful and honest assistant who strictly follows all user instructions and provides accurate responses ...''), placed either before or between $X_s$ and $X_o$. knockout attention analysis. Here, the newly introduced system prompt and the original constraint tokens are both treated as $X_o$. As shown in Fig.~\ref{fig:semantic_constraint_fig} (c,d), exhibits the same pattern: cutting attention from modality cues to $X_s$ significantly impairs modality following, whereas cutting attention to $X_o$ has negligible effect. Collectively,  These results confirm that the convergence on $X_s$ cannot be explained by positional proximity alone.

\begin{figure*}[h] 
    \centering
    \begin{subfigure}[b]{0.495\textwidth}
        \centering
        \includegraphics[page=1, width=\textwidth]{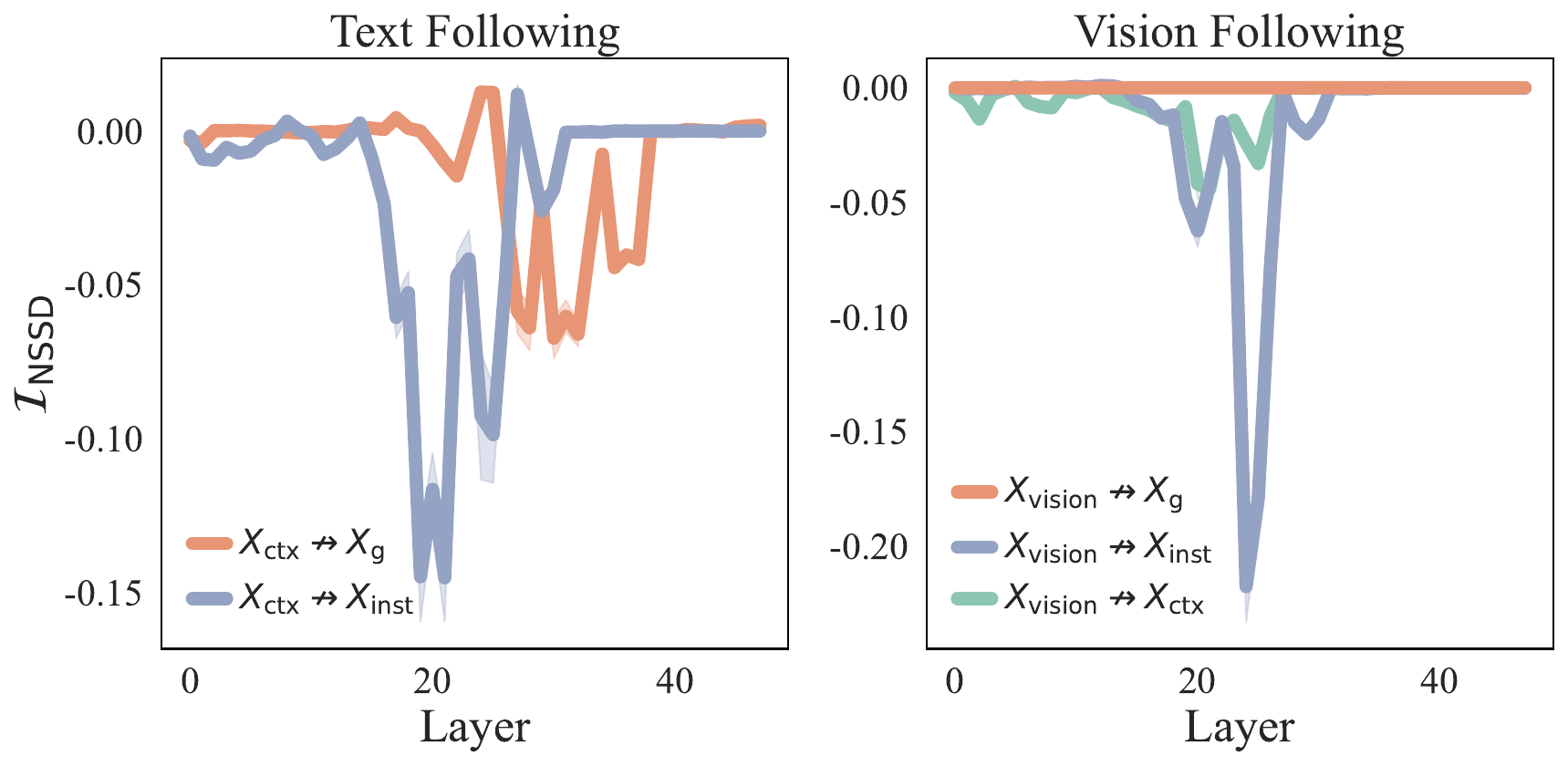}
        \caption{InternVL3-14B}
    \end{subfigure}
    \hfill 
    \begin{subfigure}[b]{0.495\textwidth}
        \centering
        \includegraphics[page=1, width=\textwidth]{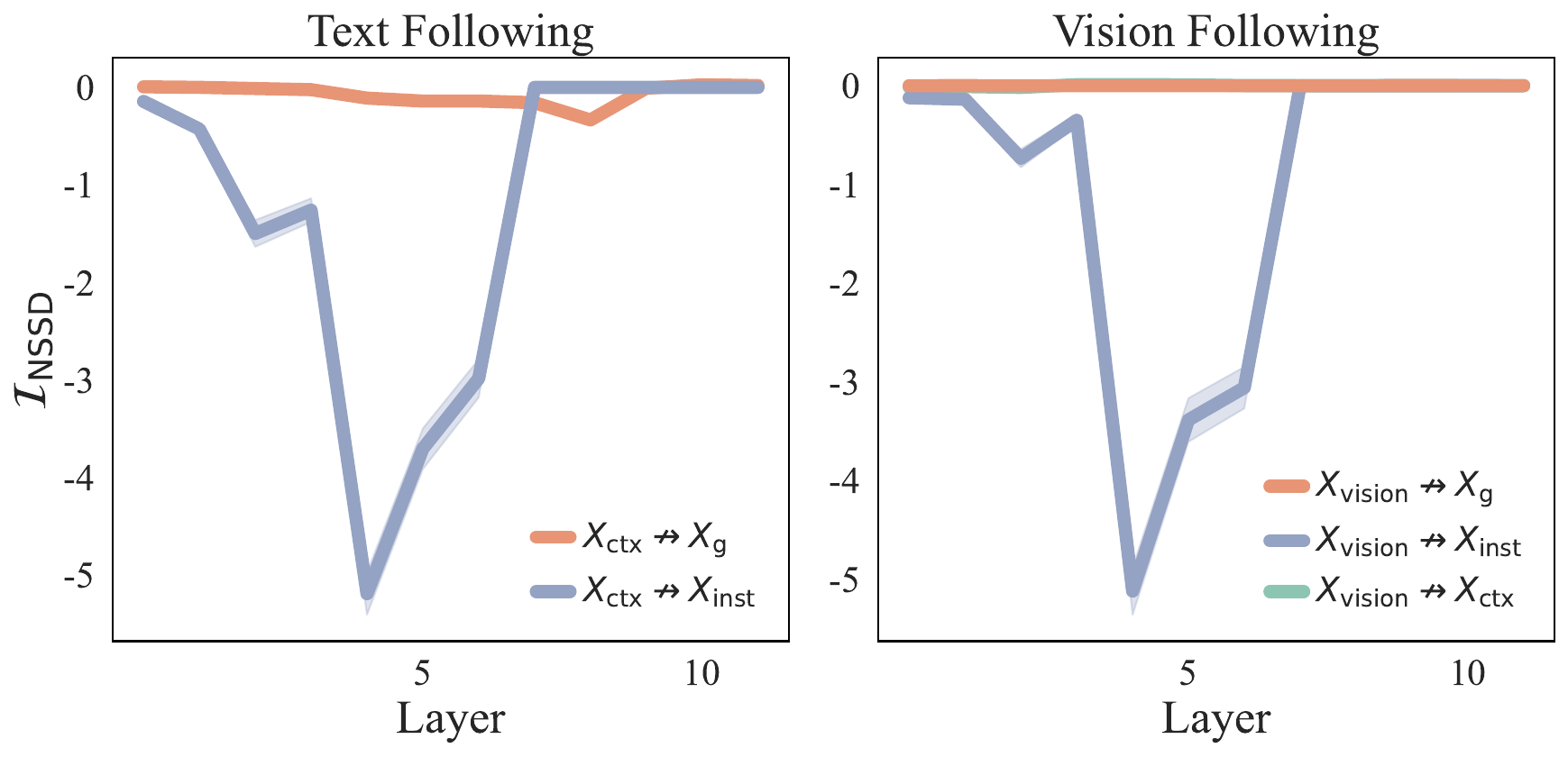}
        \caption{DeepSeek-VL2-Tiny}
    \end{subfigure}
    \caption{
    \textbf{Layer-wise $\mathcal{I}_\text{NSSD}$ profiles under different attention-knockout pathways for larger dense model and MoE-based model.} 
    Each curve denotes cutting a pathway from a source token set to a target token set, where $X_\text{vision}$, $X_\text{ctx}$, $X_\text{inst}$, and $X_\text{g}$ denote visual tokens, text context tokens, instruction tokens, and generated tokens, respectively. 
    Across both models, pathways targeting $X_\text{inst}$ produce substantially larger negative shifts than direct pathways targeting $X_\text{g}$, indicating that instruction tokens serve as a central convergence site for modality-relevant cues.
    }
    \label{fig:attention_block_larger_moe}
\end{figure*}

\subsection{Results for Generalization and Robustness of Attention Routing}
\label{supp:generalization_causal}

\paragraph{Knockout attention analysis for larger dense MLLM and MoE-based MLLM}
Consistent with Fig.~\ref{fig:vision_inst_block_comparison}, we extend our knockout attention analysis for larger dense model (InternVL3-14B~\cite{chen2024internvl}) and MoE-based model (DeepSeek-VL2-Tiny~\cite{wu2024deepseek}).
Fig.~\ref{fig:attention_block_larger_moe} shows that the fundamental mechanistic conclusions for larger dense model and MoE-based model remain highly consistent with Qwen2.5-VL-7B and InternVL3-8B in the main text.

\paragraph{Generalization to open-ended, multi-token generation scenarios}
In this section, we verify that the attention patterns observed at instruction anchors in \S\ref{sec:attention_pattern_main} generalize to open-ended, multi-token generation tasks. To this end, we investigate their role in modality interference tasks~\cite{cai2025diagnosing}. Specifically, we conduct experiments on an image captioning task paired with a misleading text context.
We selectively cut attention pathways from the vision context to the misleading text context, the instruction tokens, or the generated tokens, and measure the impact on generation quality using METEOR, CIDEr and SPICE. The interventions target layers 14–16, which were identified as critical for instruction-mediated information flow in Fig.~\ref{fig:vision_inst_block_comparison} (a) for Qwen2.5-VL-7B.

As summarized in Table~\ref{tab:causal_attention_caption}, cutting the attention pathways from the vision context to either the misleading text context or the generated tokens produces negligible changes across all metrics. In contrast, severing the pathways from the vision context to the instruction tokens consistently degrades performance across all measures.
These results demonstrate that the attention patterns centered on instruction tokens effectively generalize to open-ended, multi-token generation tasks, highlighting the structural role of instruction anchors in guiding cross-modal information integration.

\begin{table}[t]
\centering
\small
\caption{Results on Flickr30k under misleading text interference by cutting the specific attention pathway. The performance is measured by METEOR, CIDEr and SPICE (Higher is better).}
\label{tab:causal_attention_caption}
\begin{tabular}{lccc}
\toprule
Method & METEOR & CIDEr & SPICE \\
\midrule
Qwen2.5-VL-7B & 20.0 & 18.6 & 13.5 \\
Vision Context $\nrightarrow$ Text Context & 19.6 & 18.5 & 13.3 \\
Vision Context $\nrightarrow$ Generated Tokens& 20.0 & 18.4& 13.6\\
\midrule
Vision Context $\nrightarrow$ Instruction & 2.3 & 0.8 & 0.6 \\
\bottomrule
\end{tabular}
\end{table}

\begin{figure}[h!]
    \centering
    \begin{subfigure}[b]{0.48\textwidth}
        \centering
        \includegraphics[page=1, width=\textwidth]{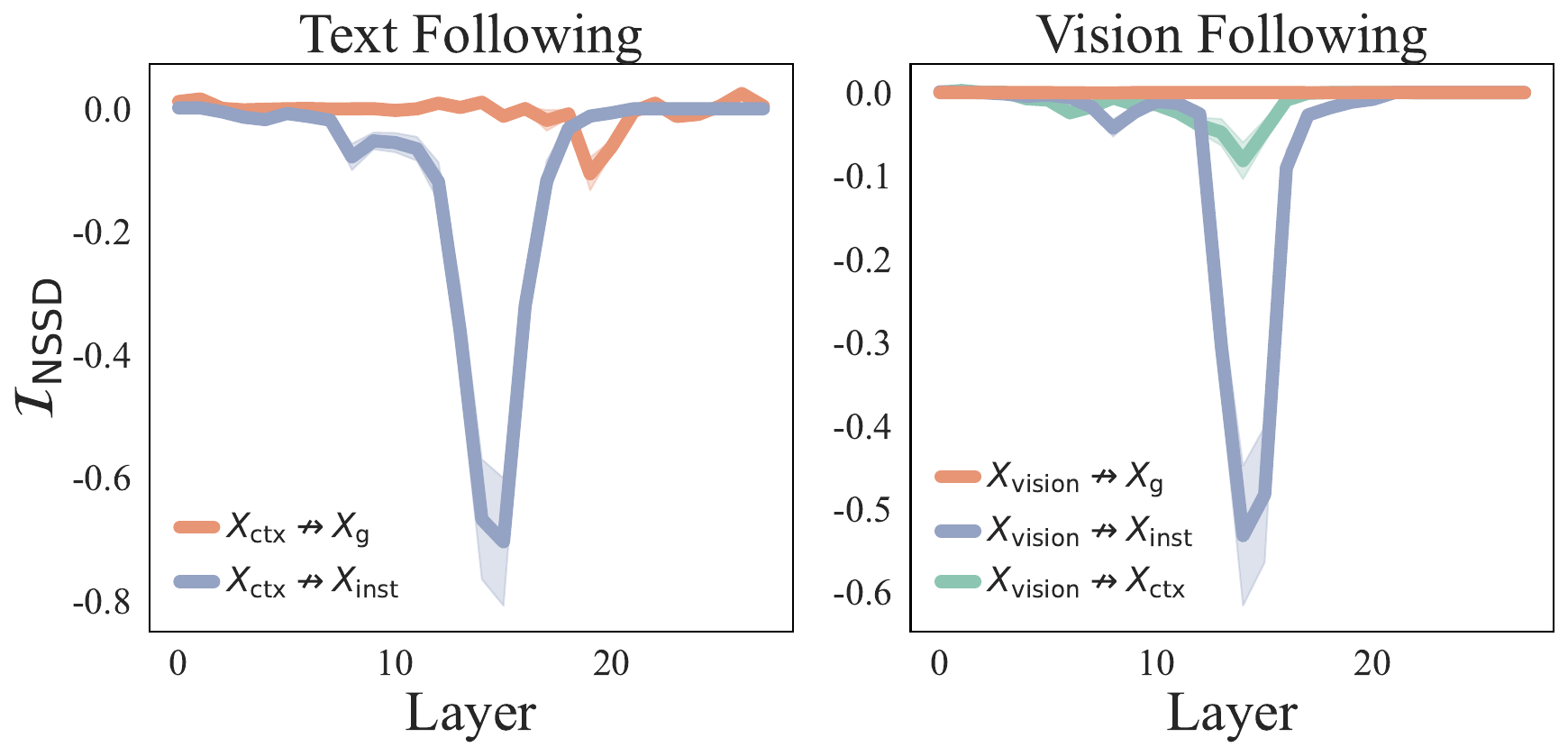}
        \caption{20\% of the evaluation data.}
    \end{subfigure}
    \hfill
    \begin{subfigure}[b]{0.48\textwidth}
        \centering
        \includegraphics[page=1, width=\textwidth]{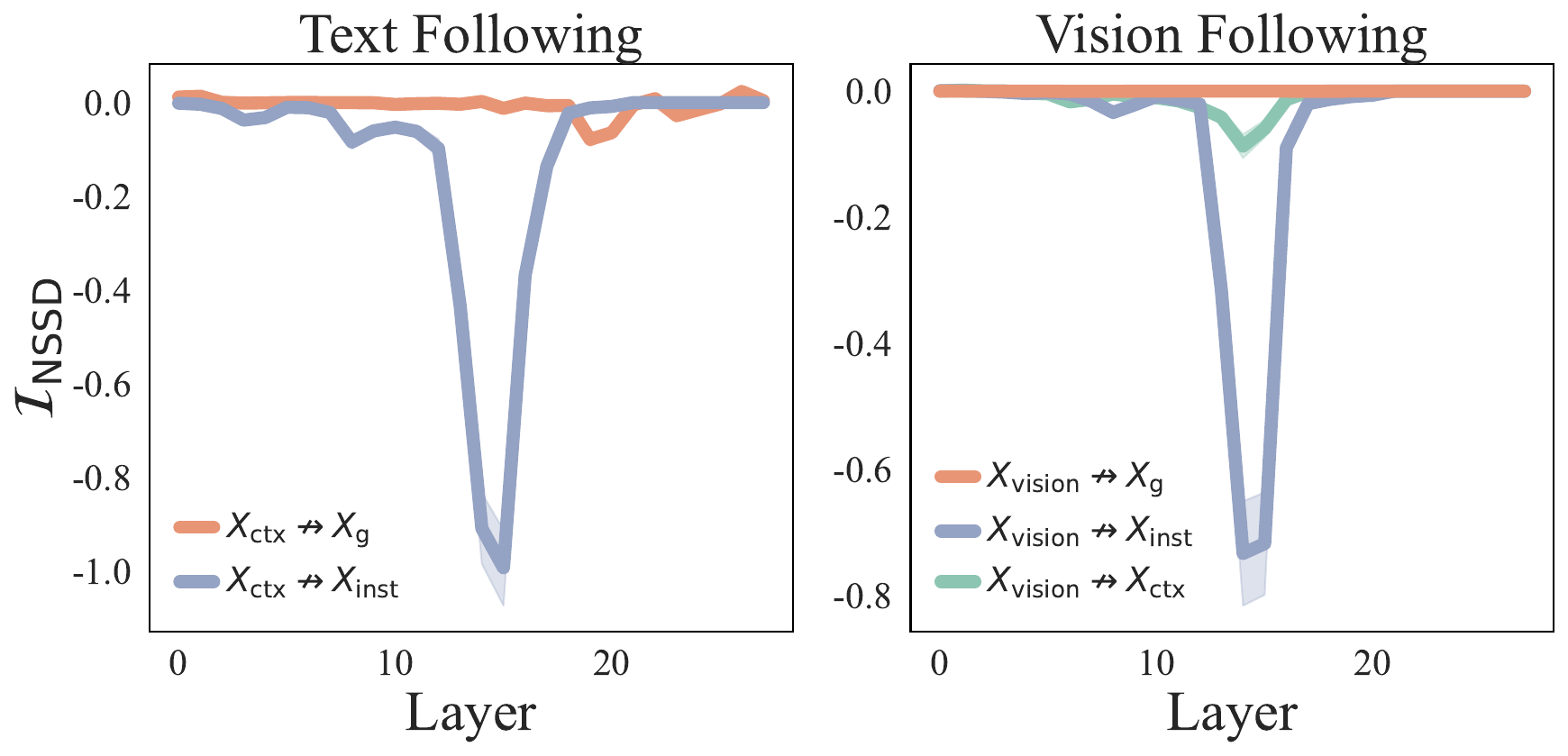}
        \caption{40\% of the evaluation data.}
    \end{subfigure}
    \vspace{0.3cm}

    \begin{subfigure}[b]{0.48\textwidth}
        \centering
        \includegraphics[page=1, width=\textwidth]{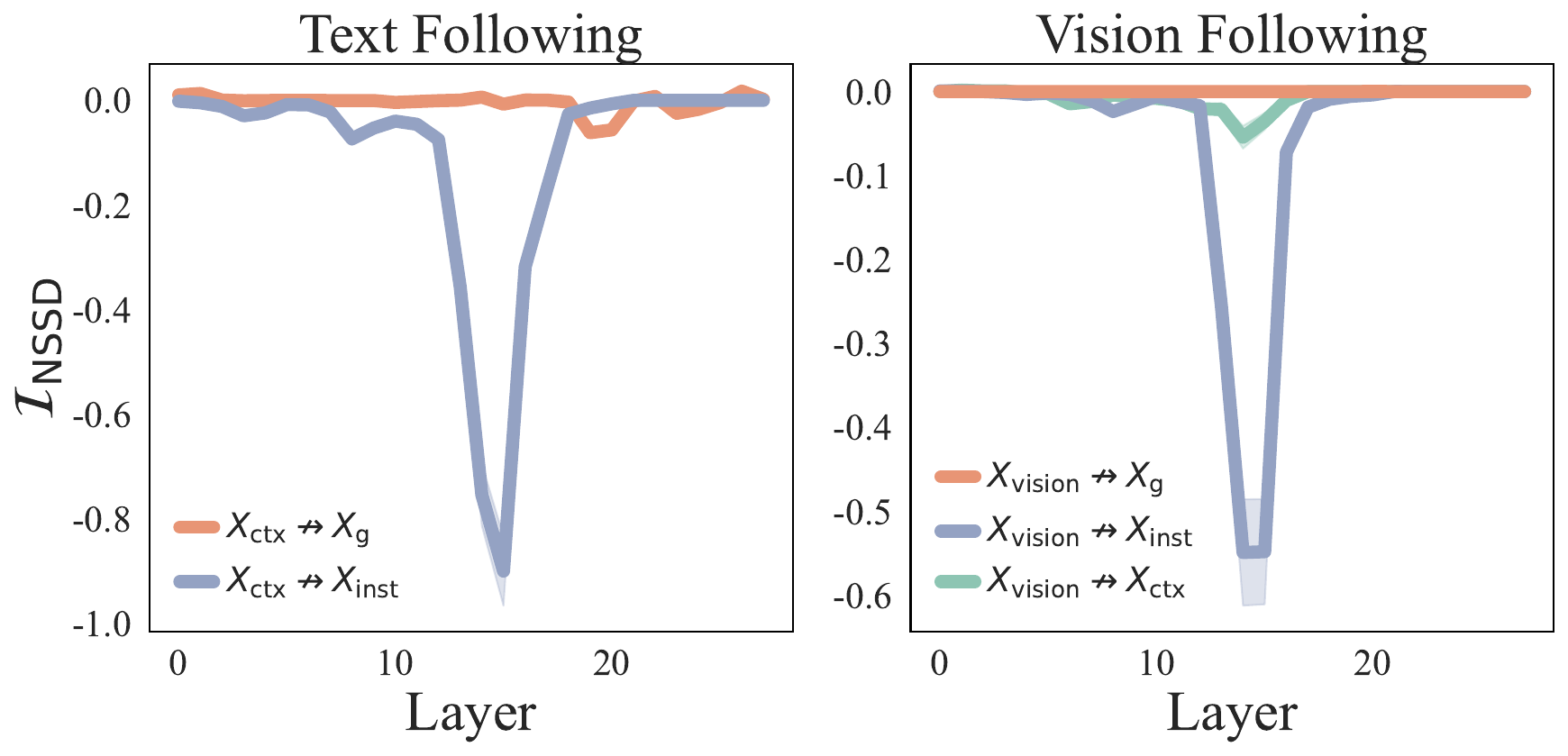}
        \caption{60\% of the evaluation data.}
    \end{subfigure}
    \hfill
    \begin{subfigure}[b]{0.48\textwidth}
        \centering
        \includegraphics[page=1, width=\textwidth]{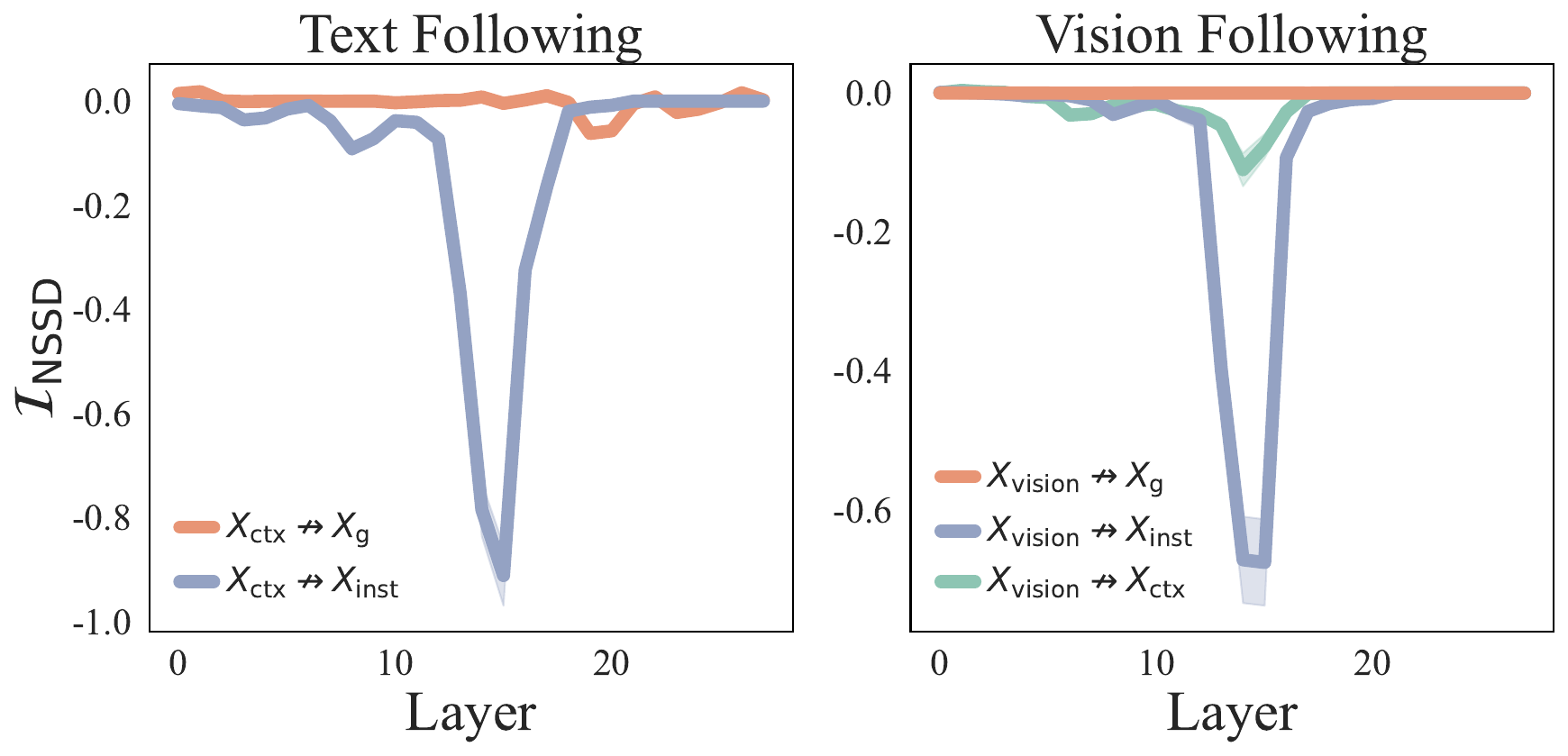}
        \caption{80\% of the evaluation data.}
    \end{subfigure}
    
    \caption{Sensitivity analysis for dataset scale. We compare the results for layer-wise causal attention analysis under different ratios: $20\%$, $40\%$, $60\%$ and $80\%$ of the evaluation dataset. The overall attention patterns remain consistent across different scales, while larger sample sizes produce smoother curves with lower variance, indicating improved stability of the estimates.}
    \vspace{+5mm}
    \label{fig:vision_text_block_qwenvl_w3_dataset_scale}
    
\end{figure}

\paragraph{Sensitivity analysis for dataset scale.}
To verify that the current dataset scale is sufficient to robustly support our causal attention analysis, we repeat the main experiment in Fig.~\ref{fig:vision_inst_block_comparison} using only $20\%$, $40\%$, $60\%$ and $80\%$ of the evaluation data, and compare the results with those obtained on the full dataset.

As shown in Fig.~\ref{fig:vision_text_block_qwenvl_w3_dataset_scale}, the core attention patterns remain highly consistent across different data ratios. In particular, the overall layer-wise patterns are preserved even when only a fraction of the data is used, indicating that our findings do not depend on a specific sample scale. At the same time, as the amount of evaluation data increases, the curves exhibit lower variance, suggesting improved statistical reliability. These results provide empirical evidence that the current dataset size is sufficiently robust to capture the attention patterns.

\begin{figure*}[h] 
    \centering
    \begin{subfigure}[b]{0.495\textwidth}
        \centering
        \includegraphics[page=1, width=\textwidth]{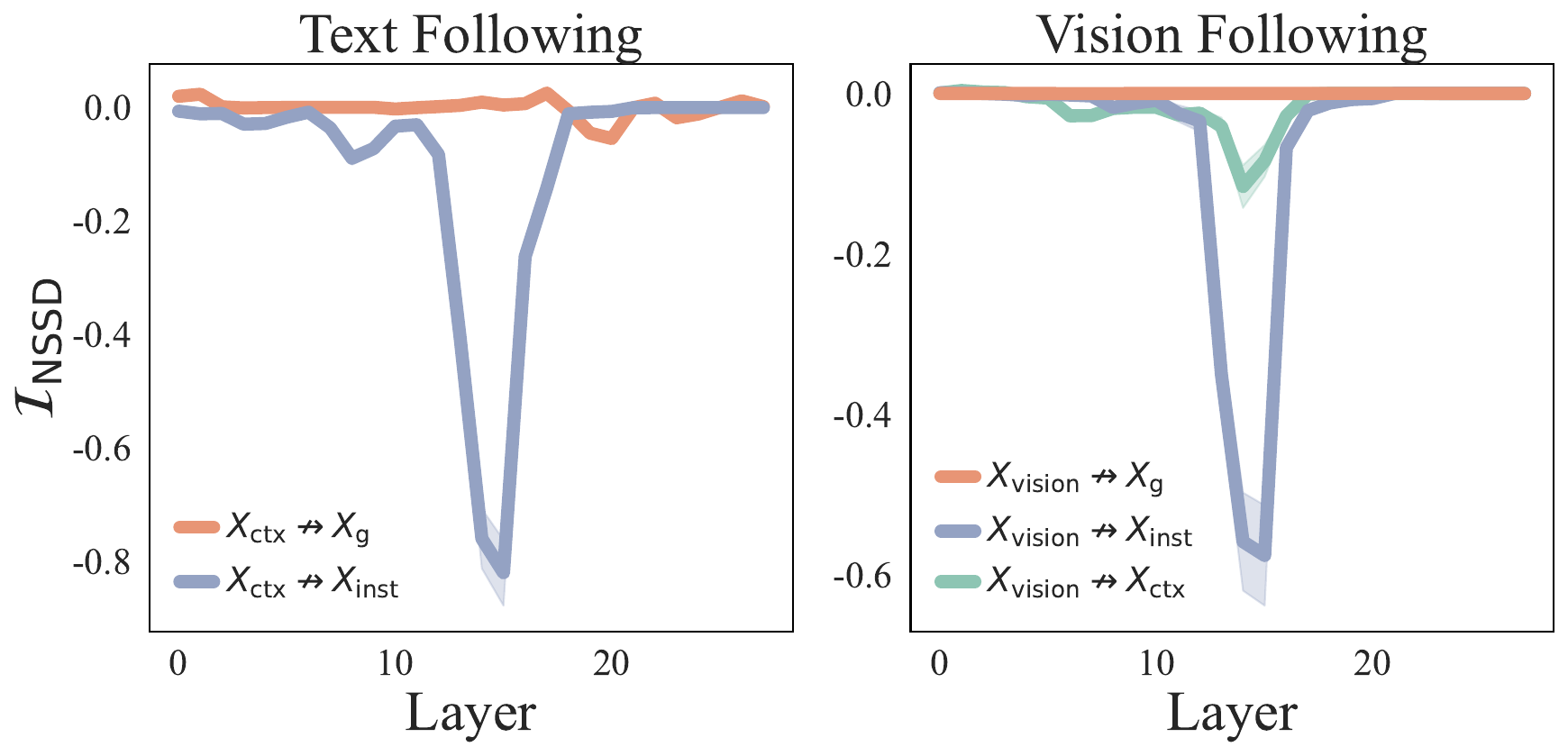}
        \caption{New Instruction Style $1$.}
    \end{subfigure}
    \hfill 
    \begin{subfigure}[b]{0.495\textwidth}
        \centering
        \includegraphics[page=1, width=\textwidth]{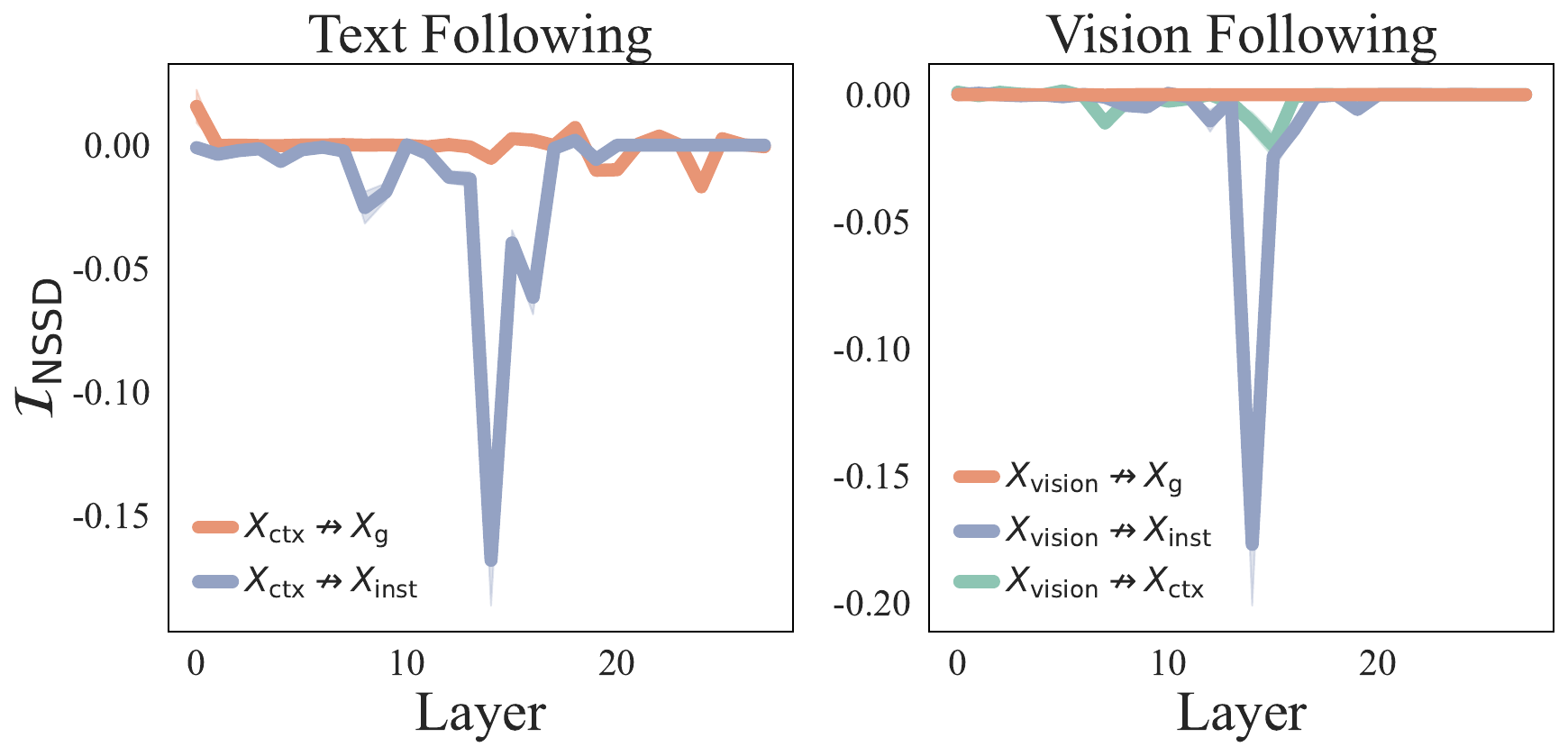}
        \caption{New Instruction Style $2$.}
    \end{subfigure}
    \caption{
\textbf{Layer-wise $\mathcal{I}_\text{NSSD}$ under attention-knockout for different instruction styles.}  
Each curve represents cutting a attention pathway, where $X_\text{vision}$, $X_\text{ctx}$, $X_\text{inst}$, and $X_\text{g}$ denote visual tokens, text context tokens, instruction tokens, and generated tokens.
The overall attention patterns remain consistent across different instruction styles.}

    \label{fig:instruction_style}
    \vspace{0.2cm}
\end{figure*}
\paragraph{Sensitivity analysis for instruction styles.}
To verify that the our causal attention analysis is robust for different instruction styles, we repeat the main experiment in Fig.~\ref{fig:vision_inst_block_comparison} using different instruction styles: ``Prioritize the information in the text over any visual elements'' and ``Focus on the textual context; ignore visual content.''.
As shown in Fig.~\ref{fig:instruction_style}, the core mechanistic conclusions remain highly consistent across different instruction styles, which confirms that our causal attention analysis is robust for different instruction styles.

\section{More Results for Mechanistic Dissection of Modality Arbitration}
\label{supp:more_results_mechanistic_dissection}
In this section, we include more results for mechanistic dissection of modality arbitration: the results for successful vision following in Apdx.~\ref{supp:Dissection_vision_following} and the results for generalization across other MLLM and open-ended, multi-token generation task in Apdx.~\ref{supp:Dissection_generalization}. 

\begin{figure*}[h] %
    \centering
    \begin{subfigure}[b]{0.245\textwidth}
        \centering
        \includegraphics[page=1, width=\textwidth]{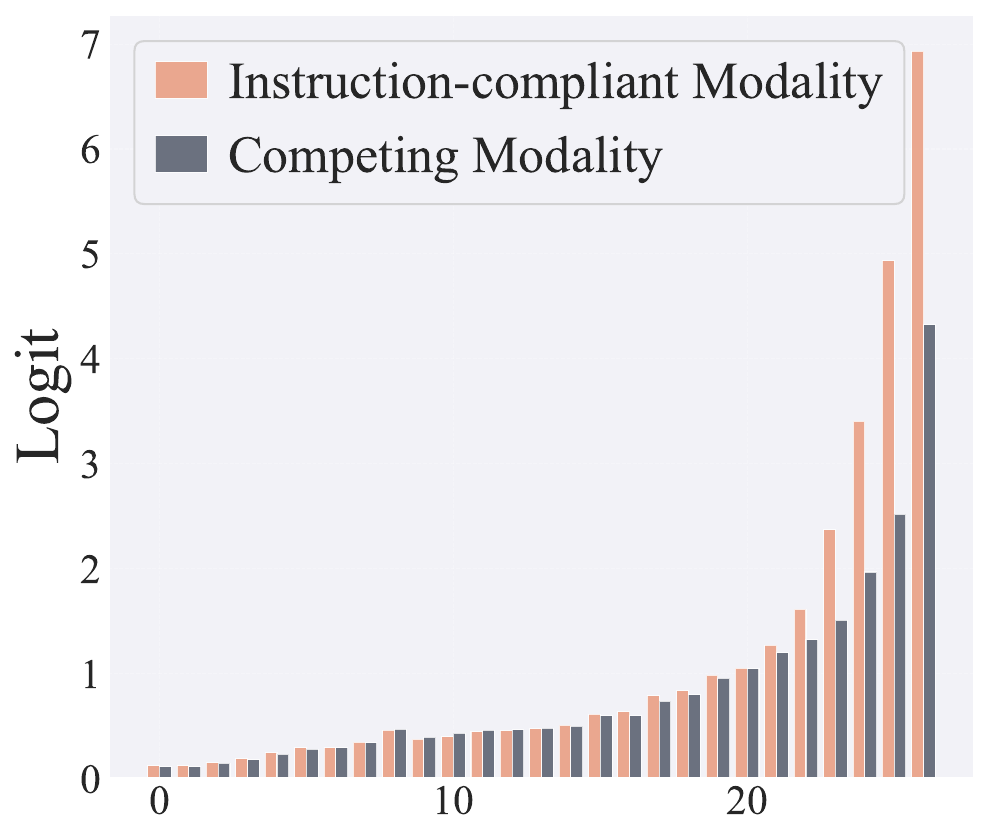}
        \vspace{-0.6cm}
        \caption{}
    \end{subfigure}
    \hfill 
    \begin{subfigure}[b]{0.245\textwidth}
        \centering
        \includegraphics[page=1, width=\textwidth]{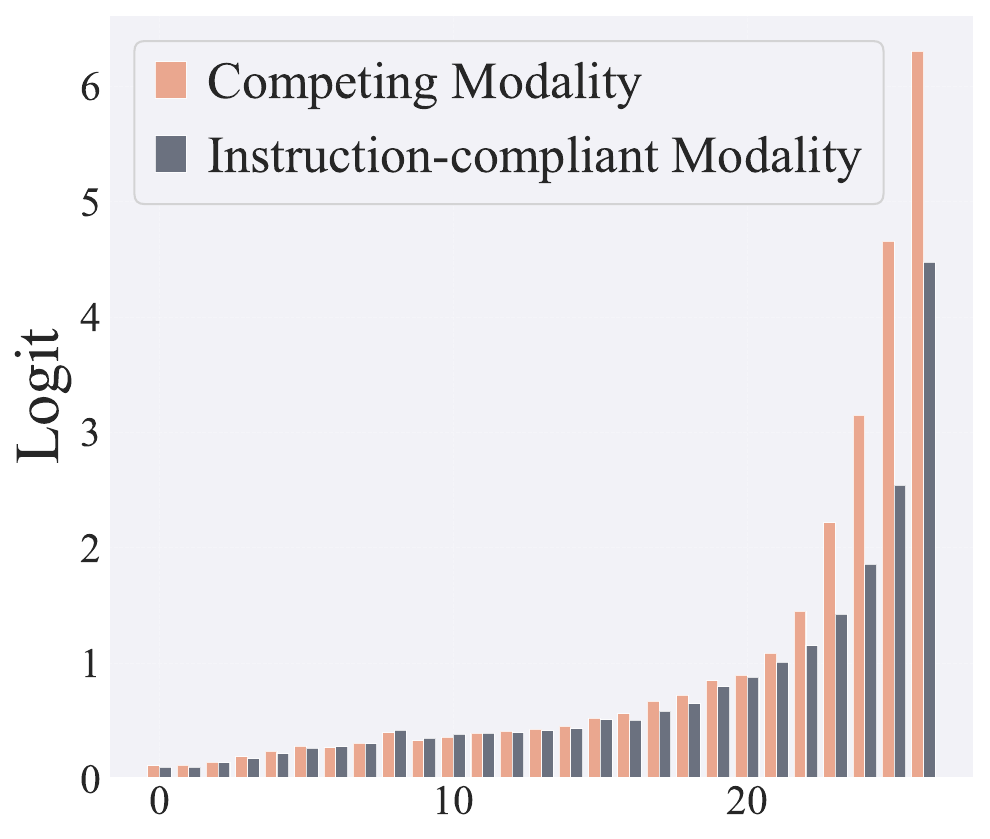}
        \vspace{-0.6cm}
        \caption{}  
    \end{subfigure}
    \hfill 
    \begin{subfigure}[b]{0.245\textwidth}
        \centering
        \includegraphics[page=1, width=\textwidth]{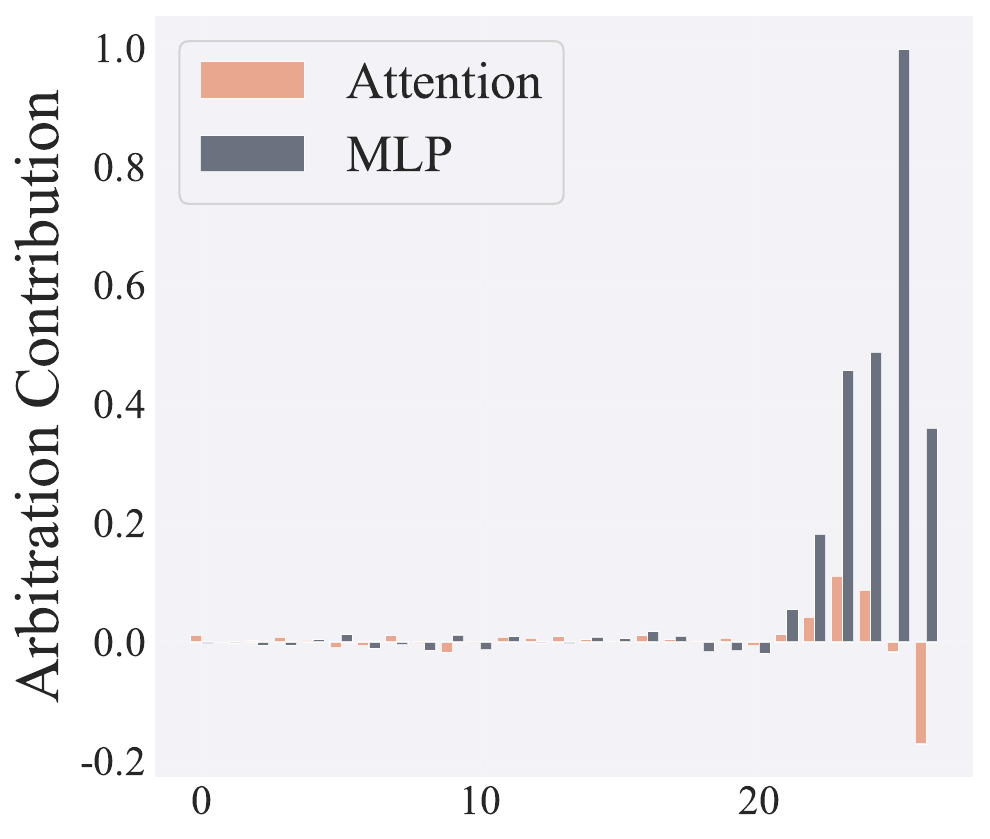}
        \vspace{-0.6cm}
        \caption{}  
    \end{subfigure}
    \hfill 
    \begin{subfigure}[b]{0.245\textwidth}
        \centering
        \includegraphics[page=1, width=\textwidth]{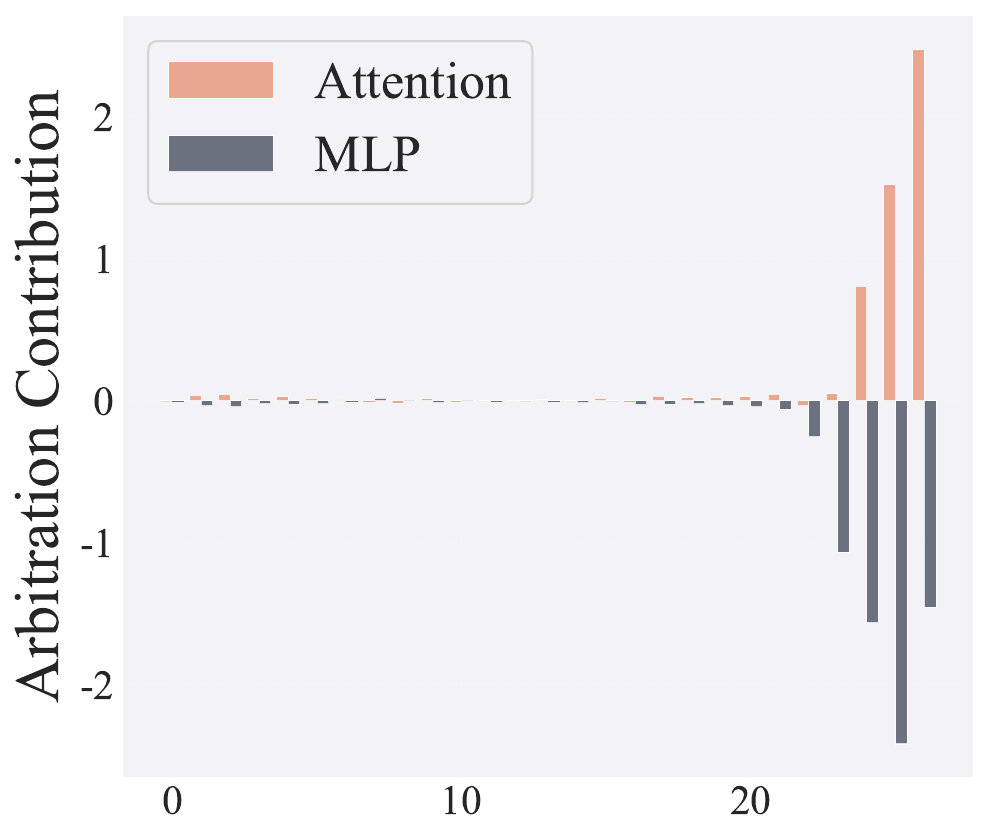}
        \vspace{-0.6cm}
        \caption{} 
    \end{subfigure}
\caption{\textbf{Layer-wise component analysis.}  
(a,b) Layer-wise evolution of the instruction-compliant and competing subspace readouts for successful and failed vision-following samples, respectively.  
(c,d) Layer-wise contributions of attention and MLP to the modality arbitration margin for successful and failed vision-following samples, respectively. 
A positive contribution indicates that the corresponding component shifts the arbitration margin toward the instruction-specified intent.
}
    \label{fig:vision_following_qwenvl}
    \vspace{-5mm}
\end{figure*}

\subsection{Detailed Results for Mechanistic Dissection of Modality Arbitration for Qwen2.5VL-7B}
\label{supp:mechanistic_dis_qwen}
\subsubsection{Results for Successful Vision Following}
\label{supp:Dissection_vision_following}
Due to the similar trends, we provide the results of mechanistic dissection of modality arbitration for text following in Fig.~\ref{fig:text_following_qwenvl} and Fig.~\ref{fig:header_qwenvl_analysis} (a,b) in \S\ref{sec:tripartite}.
In this section, we report the results for vision following.
As shown in Fig.~\ref{fig:vision_following_qwenvl}, these results show similar trend with Fig.~\ref{fig:text_following_qwenvl}.
\begin{wrapfigure}{r}{.6\textwidth}
\centering
\vspace{-3mm}
    \begin{subfigure}[b]{0.28\textwidth}
        \centering
        \includegraphics[page=1, width=\textwidth]{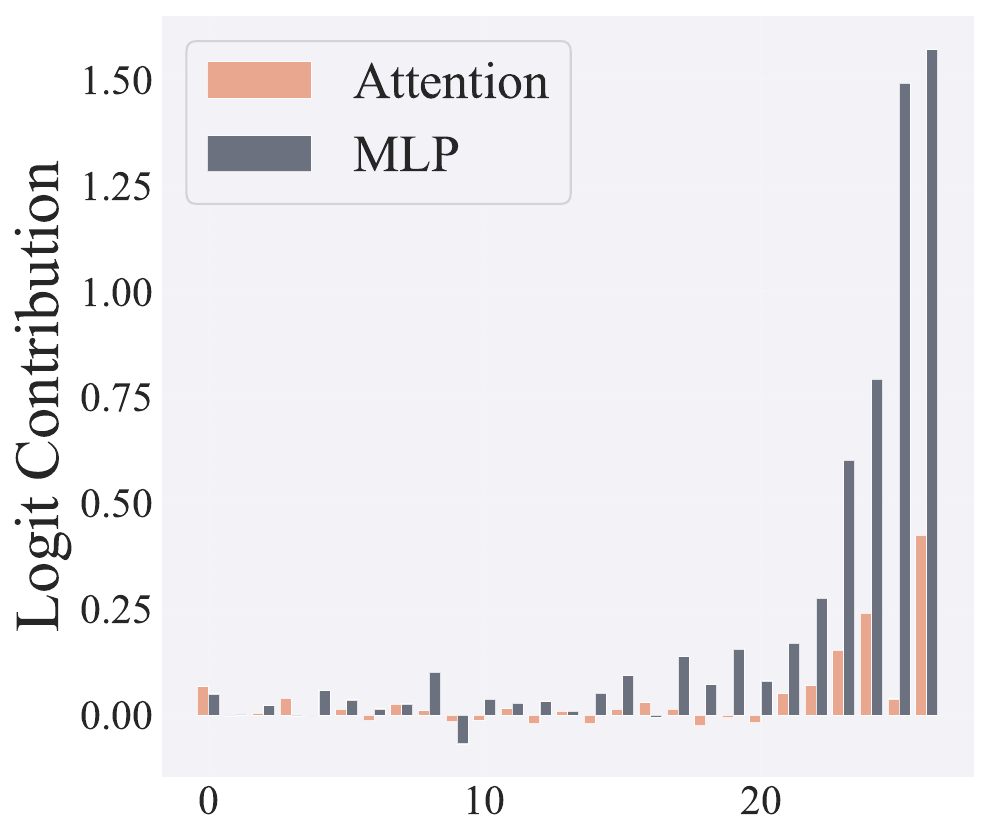}
        \vspace{-0.6cm}
        \caption{}
    \end{subfigure}
    \hfill 
    \begin{subfigure}[b]{0.28\textwidth}
        \centering
        \includegraphics[page=1, width=\textwidth]{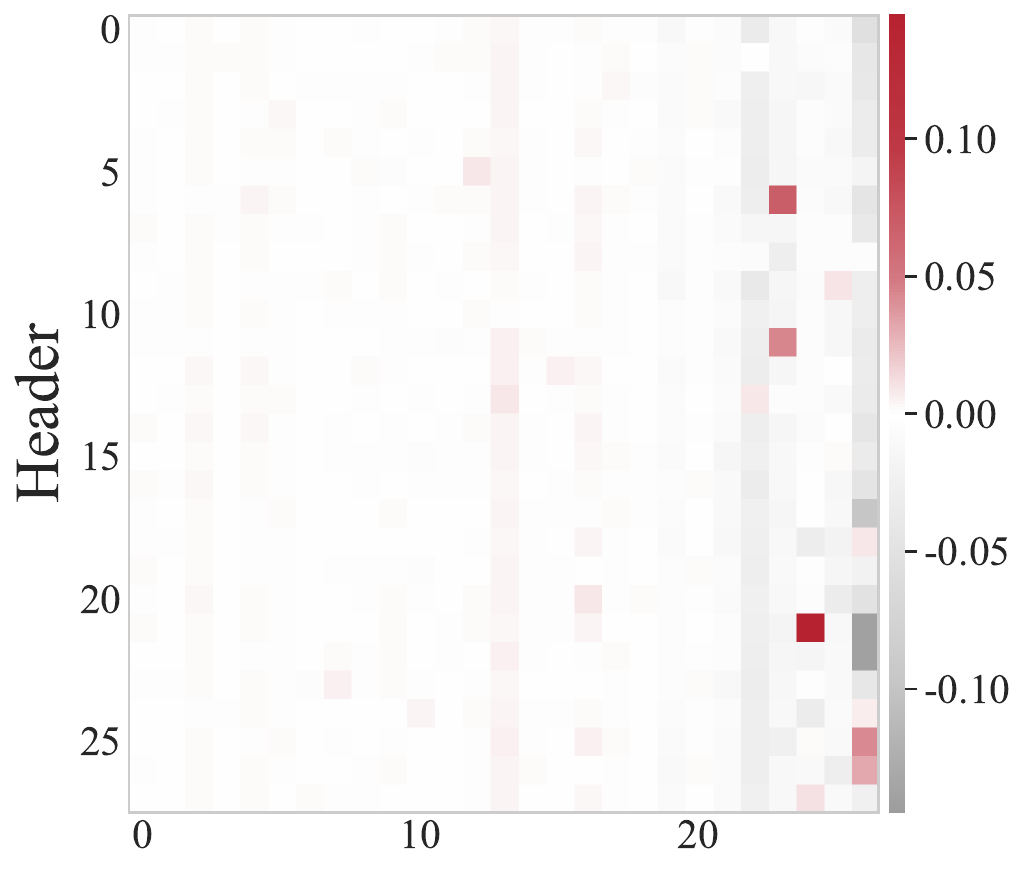}
        \vspace{-0.6cm}
        \caption{}  
    \end{subfigure}
    \caption{\textbf{Functional analysis of attention and validation of functional specialization.}
(a) Layer-wise attention contributions to the instruction-compliant and competing subspaces for successful vision-following samples.
(b) Layer-wise contributions of individual attention heads to the modality arbitration margin for successful samples.
    }
    \label{fig:vision_following_qwenvl_further}
\vspace{-5mm}
\end{wrapfigure}
Besides, we also observe that while vision-following and text-following tasks exhibit largely distinct head activation patterns, a small subset 
of high-contribution heads overlaps across tasks in Fig.~\ref{fig:header_qwenvl_analysis} (b) and Fig.~\ref{fig:vision_following_qwenvl_further} (b).  
Specifically, among the Top-40 heads ranked by their contribution to the Modality Arbitration Margin, five heads overlap between vision-following and text-following, all within the Top-10. These constitute the modality-shared heads.  
The remaining heads are largely modality-specific, indicating that most attention heads specialize for a single modality.

\subsection{Generalization for Mechanistic Dissection of Modality Arbitration}
In this section, we verify that the findings for mechanistic dissection of modality arbitration observed in \S\ref{sec:tripartite} can generalize to other MLLM and open-ended, multi-token generation tasks.

\label{supp:Dissection_generalization}
\begin{figure*}[h] %
    \centering
    \begin{subfigure}[b]{0.325\textwidth}
        \centering
        \includegraphics[page=1, width=\textwidth]{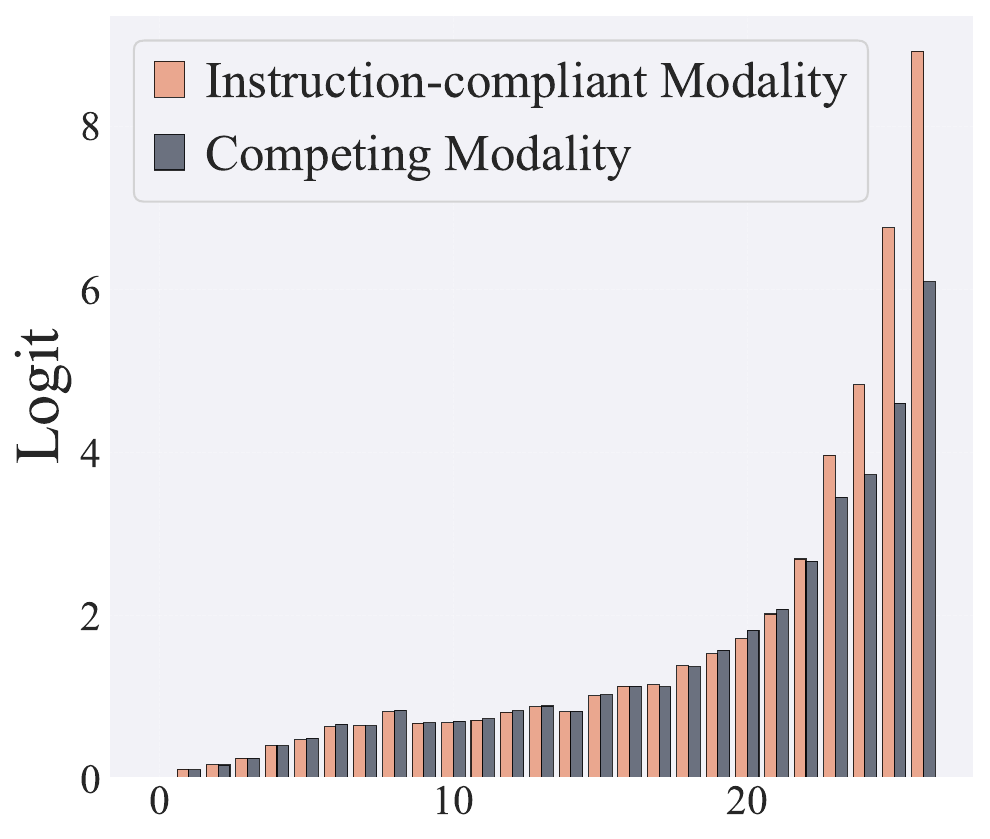}
        \vspace{-0.6cm}
        \caption{}
    \end{subfigure}
    \hfill 
    \begin{subfigure}[b]{0.325\textwidth}
        \centering
        \includegraphics[page=1, width=\textwidth]{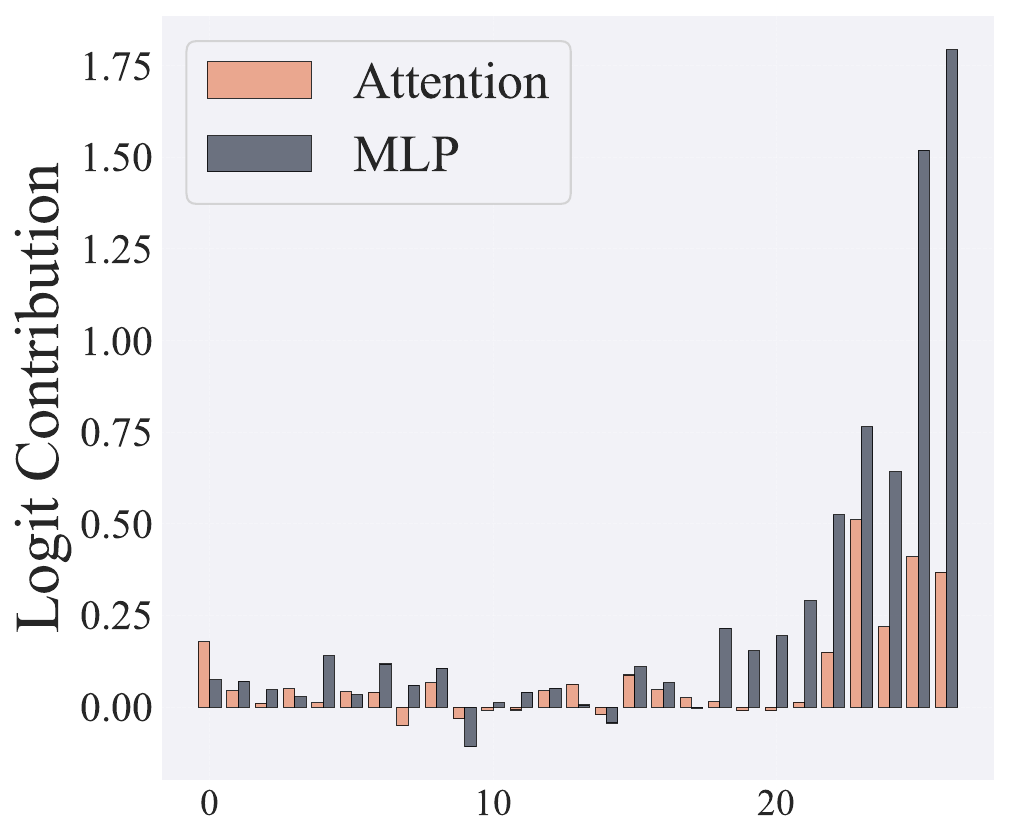}
        \vspace{-0.6cm}
        \caption{}
    \end{subfigure}
    \begin{subfigure}[b]{0.325\textwidth}
        \centering
        \includegraphics[page=1, width=\textwidth]{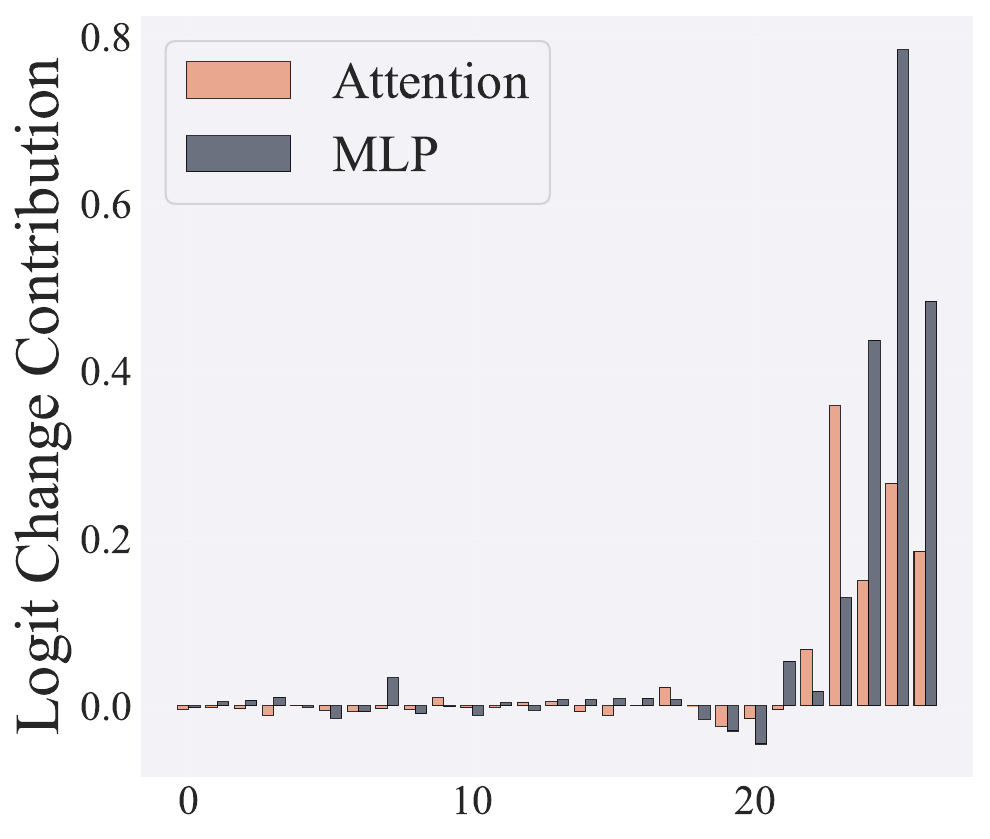}
        \vspace{-0.6cm}
        \caption{}
    \end{subfigure}

  \caption{
\textbf{Layer-wise diagnostic analysis for successful text following for InternVL3-8B.}  
(a) Layer-wise evolution of the instruction-compliant and competing subspace readouts for successful and failed text-following samples, respectively.  
(b) Layer-wise attention contributions to the instruction-compliant and competing subspaces for successful text-following samples.
(c) Layer-wise contributions of attention and MLP to the modality arbitration margin for successful and failed text-following samples, respectively. 
A positive contribution indicates that the corresponding component shifts the arbitration margin toward the instruction-specified intent.
}
  \label{fig:header_internvl_analysis_text}
\end{figure*}

\begin{figure*}[h] %
    \centering
    \begin{subfigure}[b]{0.325\textwidth}
        \centering
        \includegraphics[page=1, width=\textwidth]{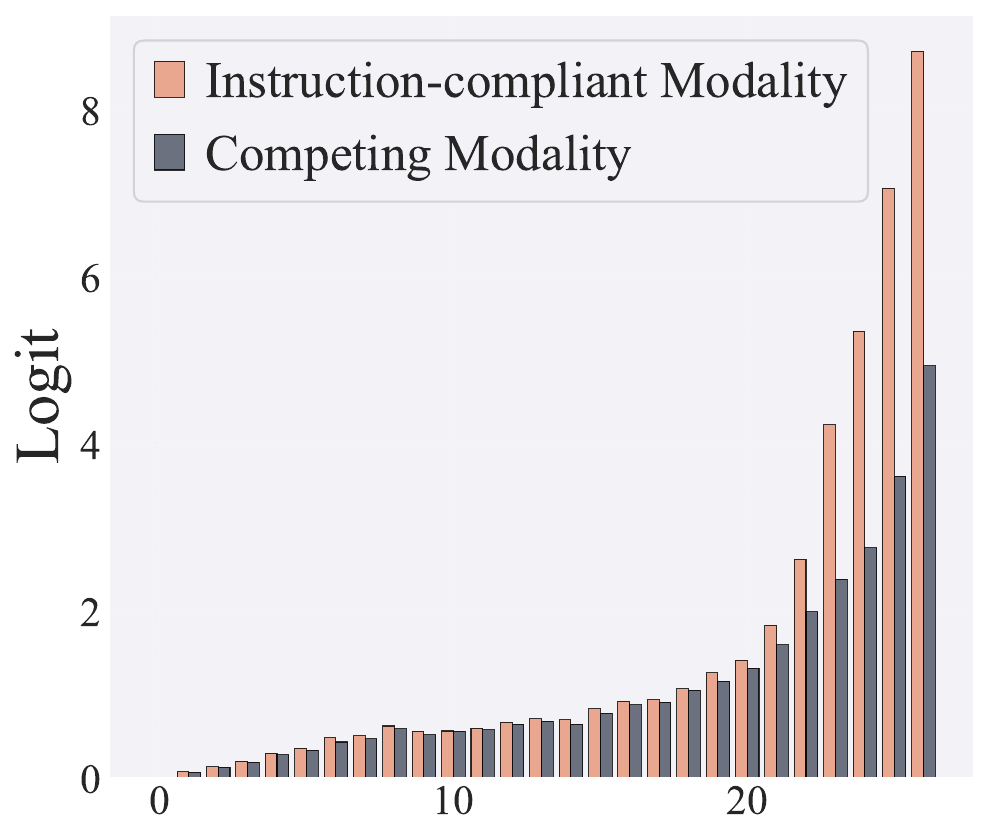}
        \vspace{-0.6cm}
        \caption{}
    \end{subfigure}
    \hfill 
    \begin{subfigure}[b]{0.325\textwidth}
        \centering
        \includegraphics[page=1, width=\textwidth]{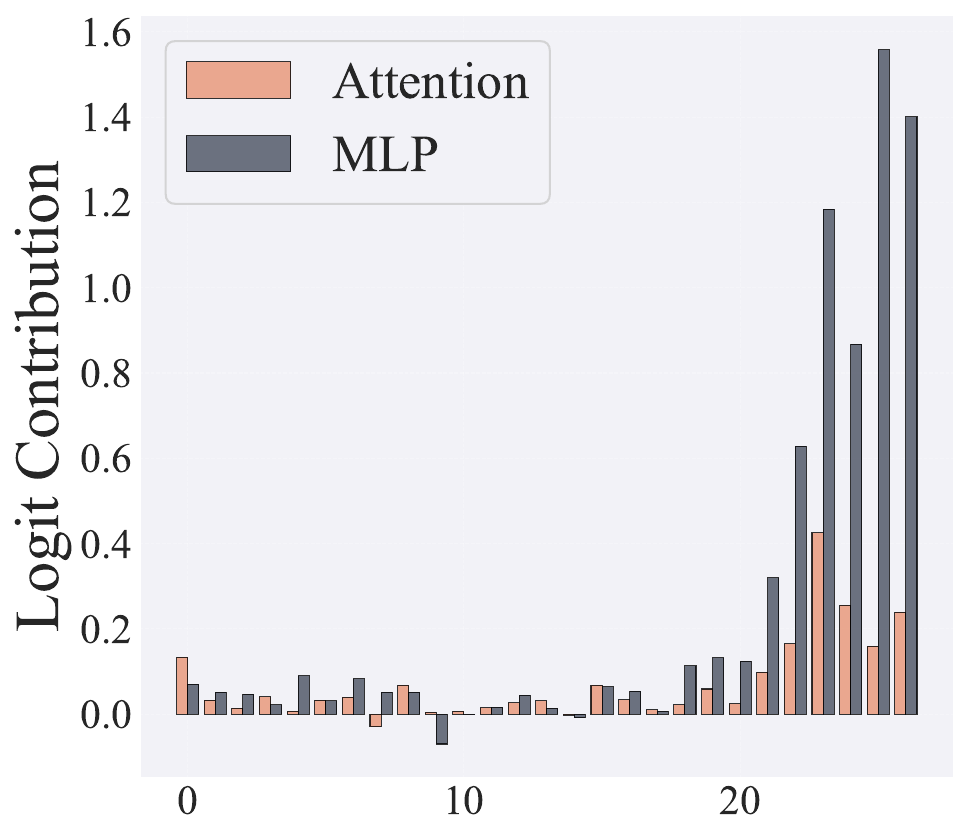}
        \vspace{-0.6cm}
        \caption{}
    \end{subfigure}
    \begin{subfigure}[b]{0.325\textwidth}
        \centering
        \includegraphics[page=1, width=\textwidth]{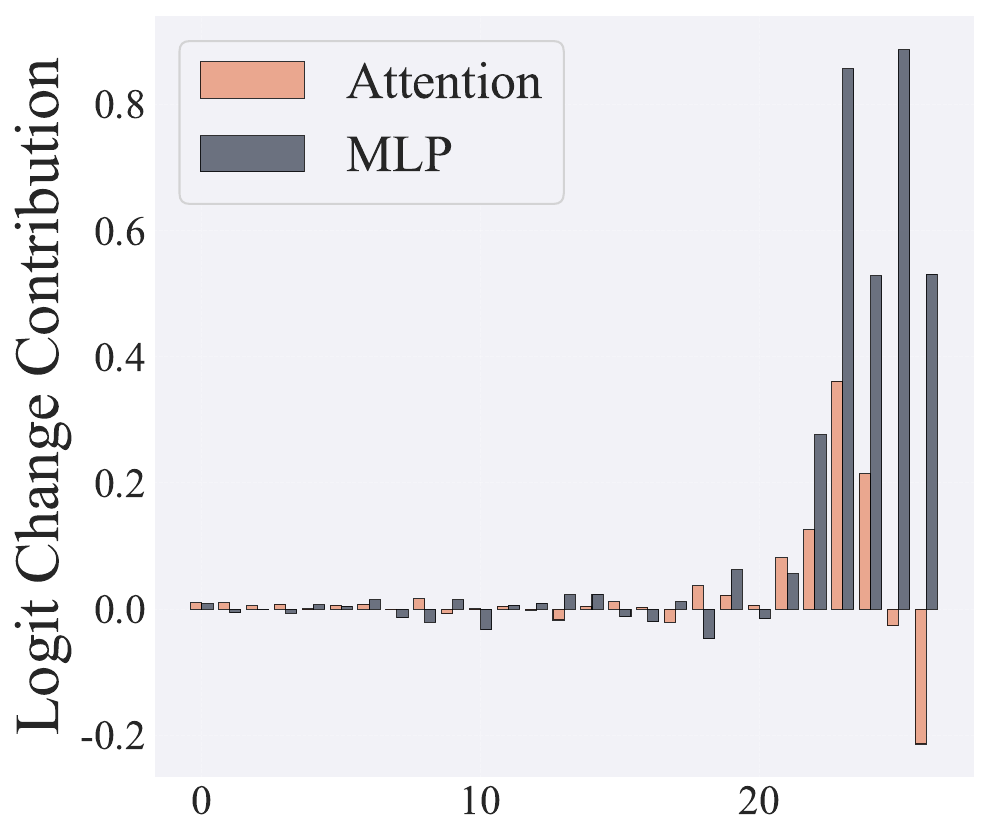}
        \vspace{-0.6cm}
        \caption{}
    \end{subfigure}

  \caption{\textbf{Layer-wise diagnostic analysis for successful vision following for InternVL3-8B.}  
(a) Layer-wise evolution of the instruction-compliant and competing subspace readouts for successful and failed text-following samples, respectively.  
(b) Layer-wise attention contributions to the instruction-compliant and competing subspaces for successful text-following samples.
(c) Layer-wise contributions of attention and MLP to the modality arbitration margin for successful and failed text-following samples, respectively. 
A positive contribution indicates that the corresponding component shifts the arbitration margin toward the instruction-specified intent.
}
  \label{fig:header_internvl_analysis_vision}
\end{figure*}

\paragraph{Generalization to other MLLMs.}
In \S\ref{sec:tripartite}, we provide the results of mechanistic dissection of modality arbitration for Qwen2.5-VL-7B. In this section, we demonstrate that these findings can generalize to InternVL3-8B.
As shown in Fig.~\ref{fig:header_internvl_analysis_text} and Fig.~\ref{fig:header_internvl_analysis_vision}, we can observe that similar patterns for the evolution of subspace, and the contribution of attention and MLP to the logit intensity of the instruction-compliant modality and the modality arbitration margin with Qwen2.5VL-7B.
These results show that the findings for mechanistic dissection of modality arbitration can generalize to other MLLM.

\begin{table}[h]
\centering
\small
\caption{Results on Flickr30k under text interference by amplifying the identified attention heads or random heads. The performance is measured by METEOR, CIDEr and SPICE (Higher is better).}
\label{tab:dissection_caption_generalization}
\begin{tabular}{llccc}
\toprule
Type&Method & METEOR & CIDEr & SPICE \\
\midrule
-&Qwen2.5-VL-7B & 20.0 & 18.6 & 13.5 \\
\midrule
\multirow{3}{*}{\textbf{Random}}
&Amplify\#10 & 19.8  & 18.6 & 13.6 \\
&Amplify-\#20 & 19.8 & 18.7 & 13.5 \\
&Amplify-\#30 & 19.7 & 18.8 & 13.3 \\
\midrule
\multirow{3}{*}{\textbf{Ours}}
&Amplify-\#10 & 18.4 & 19.9 & 13.3 \\
&Amplify-\#20 & 21.1 & 24.3 & 13.5 \\
&Amplify-\#30 & 20.3 & 23.5 & 13.8 \\
\bottomrule
\end{tabular}
\end{table}
\paragraph{Generalization to open-ended, multi-token scenarios.}
In this section, we verify that the mechanistic dissection of modality arbitration in \S\ref{sec:tripartite} can generalize to open-ended, multi-token generation tasks. 
To this end, we investigate the roles of these findings in modality interference tasks~\cite{cai2025diagnosing}. Specifically, we conduct experiments on an image captioning task paired with a misleading text context.
Following the attention amplifying methods in~\S\ref{sec:attention_head_verification}, we validate the functional roles of the identified attention heads through vision following. The performance is measured by METEOR, CIDEr and SPICE. 

As summarized in Table~\ref{tab:dissection_caption_generalization}, progressively amplifying the targeted attention heads improves the performances, especially for METEOR and CIDEr, whereas amplifying on an equal number of randomly selected heads has negligible effect.
These results demonstrate that the mechanistic dissection of modality arbitration maybe extend to open-ended, multi-token generation tasks.

\section{Ablation Studies and Robustness Analysis}
\label{supp:method_ablation}

To verify the validity and robustness of our diagnostic framework, we conduct a series of ablation experiments on the key readout function $S_m$ in Eq.~(\ref{eq:s_l_m}) in Apdx.~\ref{supp:readout_function} and the the amplifying coefficient in Apdx.~\ref{supp:amplifying_cofficient}.

\noindent\textbf{Maximum logit strategy.} $S_m$ leverages the layer-wise subspace probing $\mathrm{Logit}(Y_m \mid \mathbf{h}_i^l)$ with the 
\begin{figure}

\centering
 \includegraphics[page=1, width=0.58\textwidth]{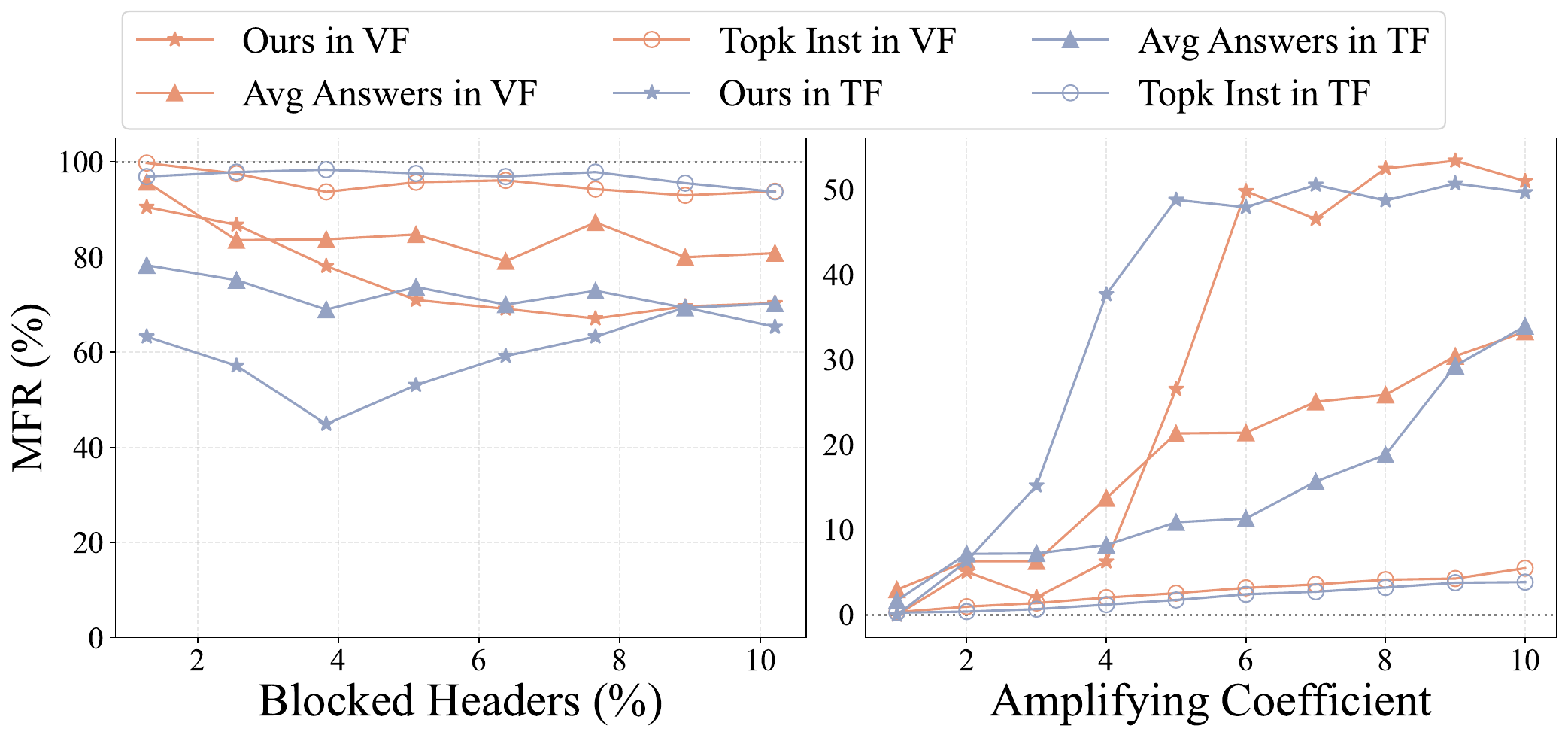}
    \caption{Interventional validation via head intervention via head intervention.
We compare the Modality Following Ratio (MFR) for Text Following (TF) and Vision Following (VF) across three settings:
(1) TopK Inst: utilizing Top-$K$ instruction aggregation ($K>1$); we report results for $K=2$ as a representative instance, given the similar performance trends observed across $K>1$.
(2) Avg Answer: employing an averaging strategy that aggregates logits across all semantically equivalent answer tokens;
(3) Ours: the default max-pooling strategy used in the main text.
Left: Impact of the number of blocked heads on MFR. Right: Impact of the amplification coefficient $\alpha$ on MFR.
    }
    \label{fig:header_ver_ablation}
\end{figure}
\begin{table}[t]
\centering
\caption{Modality Following Ratio (MFR, \%) for Text Following and Vision Following under progressively larger amplification coefficients. Performance remains stable within a moderate range, but collapses sharply once the intervention becomes excessively strong.}
\vspace{0.2cm}
\begin{tabular}{lcccccc}
\toprule
Coefficient & 8 & 10 & 12 & 14 & 16 & 18 \\
\midrule
Text Following & 48.8 & 49.7 & 47.5 & 34.1 & 21.0 & 2.0 \\
Vision Following & 52.5 & 51.0 & 48.5 & 30.6 & 21.9 & 3.3 \\
\bottomrule
\end{tabular}

\label{tab:collapse}
\end{table}
\subsection{Ablation Studies on the Readout Function $S_m$}
\label{supp:readout_function}
Specifically, we adopt different probing strategies to identify the critical attention heads.
We then compare the effects on modality following by blocking or amplifying attention using different attention heads identified via various strategies, in order to evaluate the efficacy of the design choices.

We compare this to an average-logit strategy. As shown in Fig.~\ref{fig:header_ver_ablation}, while average pooling provides moderate localization—yielding approximately a 20\% shift (Blocking or Amplifying) in modality following ratio (MFR) under intervention—it remains inferior to the max-pooling strategy.
maximum logit across all candidate entities for each subspace to measure the model belief in Eq.~(\ref{eq:max_logit}).

\noindent\textbf{Instruction aggregation strategy.} 
$S_m$ utilizes a Top-$K$ strategy to quantify the intensity of a specific subspace within instruction tokens in Eq.~(\ref{eq:max_logit}), with $K=1$ as the default. 
We vary $K$ to evaluate this design choice and observe that $K>1$ results in only minor variations in the modality following ratio (MFR) for both attention blocking and amplification interventions, as shown in Fig.~\ref{fig:header_ver_ablation}.
This observation provides empirical validation for the strategy adopted in our main text.

\subsection{Ablation Studies on the Amplifying coefficient}
\label{supp:amplifying_cofficient}
To examine the stability boundary of amplification intervention, we further evaluate the effect of progressively increasing the amplification coefficient and analyze the resulting collapse behavior.

As shown in Table~\ref{tab:collapse}, the model collapses once the amplification coefficient becomes too large. At a threshold of 18, modality-following performance drops to nearly zero, indicating that excessive intervention destabilizes the arbitration process.







\end{document}